%% file: main.tex
\documentclass{article}

\PassOptionsToPackage{numbers}{natbib}

     \usepackage[final]{neurips_2025}

\usepackage[utf8]{inputenc} 
\usepackage[T1]{fontenc}    
\usepackage[backref]{hyperref}       
\usepackage{url}            
\usepackage{booktabs}       
\usepackage{amsfonts}       
\usepackage{nicefrac}       
\usepackage{microtype}      
\usepackage{xcolor}         

\usepackage{amsmath}
\usepackage[capitalize,noabbrev]{cleveref}
\usepackage{multirow}
\usepackage{bm}
\usepackage{bbm}
\usepackage{soul}
\usepackage{xcolor}
\usepackage{mathtools}
\usepackage{pifont}
\usepackage{subcaption}
\usepackage{wrapfig}
\usepackage{amssymb, algorithm, algorithmic}

\usepackage{amsthm}
\newtheorem*{remark}{Remark}

\DeclareMathOperator*{\argmin}{arg\,min}

\definecolor{BrickRed}{rgb}{0.6,0,0}
\definecolor{RoyalBlue}{rgb}{0,0,0.8}
\definecolor{Tdgreen}{rgb}{0,0.4,0.7}
\definecolor{cadmiumgreen}{rgb}{0.0, 0.42, 0.24}
\hypersetup{colorlinks, citecolor=RoyalBlue, linkcolor=RoyalBlue}

\newcommand{\tabnum}[2]{{#1{\footnotesize±}{\scriptsize#2}}}
\newcommand{\besttabnum}[2]{{\textbf{#1}{\footnotesize\textbf{±}}{\scriptsize\textbf{#2}}}}

\newcommand{\set}[1]{\mathcal{#1}}
\renewcommand{\vec}[1]{\boldsymbol{#1}}
\newcommand{\mat}[1]{\mathbf{#1}}

\newcommand{\nrep}[1]{h_{v}^{\mathcal{G}, #1}}
\newcommand{\nrepfull}[3]{h_{#1}^{#2, #3}}
\newcommand{\gme}{\set{G}^{-\frac{1}{N}}}

\newcommand{\nrepgme}[1]{h_{v}^{\gme, #1}}

\newcommand{\editedgraph}{\set{G}^{\epsilon}}
\newcommand{\radj}{A^{\epsilon}}
\newcommand{\ruv}{A^{\epsilon}_{uv}}
\newcommand{\rvu}{A^{\epsilon}_{vu}}
\newcommand{\oq}{f_{\text{OQ}}}
\newcommand{\gpbrf}{\theta^*_{\epsilon}}

\newcommand{\hth}{h^{\theta}_x}
\newcommand{\hts}{h^{\theta_s}_x}
\newcommand{\htl}{h^{\text{lin},\theta}_x}
\newcommand{\ghth}{h^{\set{G},\theta}_v}
\newcommand{\ghts}{h^{\set{G},\theta_s}_v}
\newcommand{\ghtl}{h^{\set{G},\theta,\text{lin}}_v}
\newcommand{\lpbrfr}{\theta^*_{\text{lin},x',y',\epsilon}}
\newcommand{\lgpbrfr}{\theta^*_{\text{lin},\epsilon}}

\newcommand{\ours}{Ours}

\newcommand{\todo}[1]{{\color{red}{#1}}}

\title{Influence Functions for Edge Edits \\ in Non-Convex Graph Neural Networks}

\author{%
  Jaeseung Heo$^{1}$, Kyeongheung Yun$^{2}$, Seokwon Yoon$^{2}$, \\
  \textbf{MoonJeong Park$^{1}$, Jungseul Ok$^{1,2}$, Dongwoo Kim$^{1,2}$} \\
  $^{1}$Graduate School of Artificial Intelligence \\
  $^{2}$Department of Computer Science \& Engineering \\
  POSTECH, South Korea \\
  \texttt{\{jsheo12304,yuonsinsa,swyoon,mjeongp,jungseul,dongwookim\}@postech.ac.kr}
}

\begin{document}

\maketitle

\input{tex/0.abstract}

\input{tex/1.introduction}

\input{tex/3.preliminary}

\input{tex/4.method}

\input{tex/5.experiments}

\input{tex/6.applications}

\input{tex/2.related_work}
\input{tex/7.conclusion}

\section*{Acknowledgements}
This work was supported by the National Research Foundation of Korea (NRF) grant funded by the Korea government(MSIT) (RS-2024-00337955; RS-2023-00217286) and Institute of Information \& communications Technology Planning \& Evaluation (IITP) grant funded by the Korea government(MSIT) (RS-2024-00457882, National AI Research Lab Project; RS-2019-II191906, Artificial Intelligence Graduate School Program(POSTECH)).

\bibliography{bib}
\bibliographystyle{plainnat}

\newpage
\input{tex/appendix}

\end{document}

%% file: tex/0.abstract.tex
\begin{abstract}
Understanding how individual edges influence the behavior of graph neural networks (GNNs) is essential for improving their interpretability and robustness. Graph influence functions have emerged as promising tools to efficiently estimate the effects of edge deletions without retraining. However, existing influence prediction methods rely on strict convexity assumptions, exclusively consider the influence of edge deletions while disregarding edge insertions, and fail to capture changes in message propagation caused by these modifications. In this work, we propose a proximal Bregman response function specifically tailored for GNNs, relaxing the convexity requirement and enabling accurate influence prediction for standard neural network architectures. Furthermore, our method explicitly accounts for message propagation effects and extends influence prediction to both edge deletions and insertions in a principled way. Experiments with real-world datasets demonstrate accurate influence predictions for different characteristics of GNNs. We further demonstrate that the influence function is versatile in applications such as graph rewiring and adversarial attacks.
\end{abstract}

%% file: tex/1.introduction.tex
\section{Introduction}
Graph neural networks (GNNs) have demonstrated that leveraging structural relationships, often encoded as connectivity between data points, can enhance the predictive performance of neural networks across many tasks. Although the literature clearly identifies the importance of the relationship, the individual contribution of each connectivity, i.e., an edge, remains poorly understood. 

Several recent studies have explored edge importance from a particular perspective. For example, \citet{nguyen2023revisiting} propose an edge rewiring method to mitigate the problem of over-smoothing~\citep{li2018deeper}, a phenomenon where the learned node representation becomes indistinguishable as the depth of the GNN increases. \citet{alon2020bottleneck} suggest edge rewiring methods to overcome the over-squashing, which occurs when information propagation encounters bottlenecks between distant nodes. 

Despite progress in addressing individual challenges, a unified framework for quantifying edge influence across these perspectives would provide a more comprehensive understanding of their role in graph neural networks. 
For instance, it would allow us to assess how modifying a single edge affects model behavior from both over-smoothing and over-squashing perspectives. 
On the other hand, influence functions have been introduced to quantify the impact on evaluation metrics, such as validation loss, when a training data point is removed~\citep{koh2017understanding}. To do so, the function estimates the \emph{changes in model parameters} when the target data point is excluded from the training.

Applying influence functions~\citep{koh2017understanding} to GNNs introduces a unique challenge, primarily because modifying an edge can alter the \emph{propagation paths} and consequently change the underlying \emph{computational graph} structure. 
Previous attempts to adapt influence functions for GNNs have focused on parameter changes, failing to address the influence of the changes in the computational graph~\citep{chen2022characterizing, wu2023gif}. 
Moreover, their influence functions rely on the strict convexity assumption of the loss function with respect to the model parameters, which limits their applicability to widely used non-convex GNNs.

In this work, we propose an \emph{influence function} tailored specifically for GNNs, enabling precise predictions regarding the effects of edge modifications from multiple perspectives. To do so, we overcome the unique challenges of GNNs by deriving the changes in the evaluation function from \emph{fundamental principles of calculus}. As a result, our function can measure the influence of both edge deletion and insertion, the latter of which has not been explored in previous studies. 
To extend the influence function to non-convex GNNs, we establish the proximal Bregman response function~\citep{bae2022if} for node classification. The influence function derived from this response function relies on weaker assumptions that generally hold for non-convex GNNs.
\Cref{fig:toy_example} shows an example of influence analysis for a barbell graph from two different perspectives. Through the analysis, one can identify that the edges connecting two different clusters have opposite influences on over-squashing and over-smoothing, providing a unifying view on edge importance.

Experiments on real-world datasets demonstrate that our influence function accurately predicts edge influence in non-convex GNNs. We further show that the influence function is a versatile tool for analyzing various properties of GNNs. We present three practical applications, including 1) an analysis of edge rewiring methods suggested to improve the predictive performance of GNNs, 2) identifying adversarial edge edits that could alter node predictions, and 3) an analysis of edges connecting the nodes with the same label or different labels in terms of the node classification performance.

\input{fig/figure/toy_example}

%% file: fig/figure/toy_example.tex
\begin{figure}[t]
    \centering
    \begin{subfigure}[a]{\textwidth}
        \centering
        \includegraphics[width=0.85\linewidth]{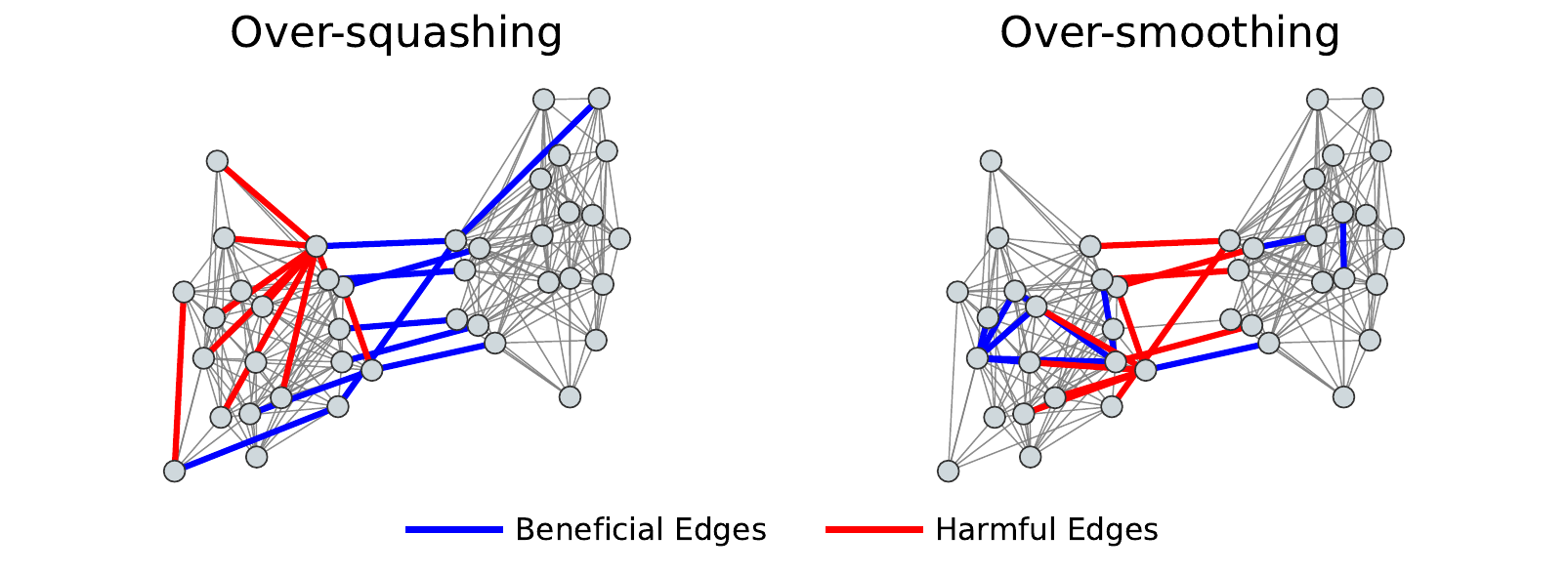}
    \end{subfigure}
    \caption{
    An illustration of beneficial and harmful edges identified by an influence function proposed in this work with respect to two evaluation metrics. 
    A harmful edge is one that either blocks information propagation between nodes (over-squashing \citep{alon2020bottleneck}) or makes node representations indistinguishable (over-smoothing \citep{li2018deeper}).
    The barbell graph consists of two clusters, each with distinct node labels. The influence function can be used to analyze the properties of edges. For example, the edges connecting two different clusters mitigate over-squashing while amplifying over-smoothing. 
    }\label{fig:toy_example}
\end{figure}

%% file: tex/3.preliminary.tex
\section{Preliminary}



\subsection{Influence function for convex models}
Let \(\mathcal{L}(x, y, \theta)\) be a loss function, where \(x\) is the input, \(y\) the label, and \(\theta\) the model parameters. The optimal parameters \(\theta^*\) minimize the empirical loss over the training set \(\mathcal{D}_{\text{train}}\):

\begin{equation}
\label{eqn:empirical_minimization}
\theta^*=\argmin_{\theta}\sum_{(x,y)\in\mathcal{D}_{\text{train}}}\mathcal{L}(x,y,\theta).
\end{equation}

Influence functions quantify how removing a specific training data point \((x',y')\) impacts model parameters. Computing the exact impact would require retraining the model by excluding the data point from the training set, which is computationally expensive. 
Instead, one can model a \emph{response function} that measures the changes in parameters when the data point is upweighted by an amount of $\epsilon \in \mathbb{R}$ by solving the following optimization problem:

\begin{equation}
\label{eqn:loo}
\theta_{x',y',\epsilon}^* \coloneq \argmin_{\theta}\frac{1}{N}\sum_{(x,y)\in\mathcal{D}_{\text{train}}}\mathcal{L}(x,y,\theta)+\epsilon\mathcal{L}(x',y',\theta),
\end{equation}

where $N=\lvert\mathcal{D}_{\text{train}}\rvert$. For example, $\theta_{x',y',\epsilon=-1/N}^*$ corresponds to the optimal parameter obtained from $\mathcal{D}_{\text{train}}\setminus\{x',y'\}$.

The changes in parameter further influence \emph{evaluation function} $f$, such as a validation loss. 
\citet{koh2017understanding} demonstrate that when the loss function \(\mathcal{L}\) is strictly convex with respect to \(\theta\), the derivative of \(f\left(\theta_{x',y',\epsilon}^*\right)\) with respect to \(\epsilon\) evaluated at \(\epsilon=0\) is:

\begin{equation}
\label{eqn:influence_function_gradient}
\left.\frac{df\left(\theta_{x',y',\epsilon}^*\right)}{d\epsilon}\right|_{\epsilon=0}
=\nabla_{\theta}f(\theta^*)^{\top}\left.\frac{d\theta_{x',y',\epsilon}^*}{d\epsilon}\right|_{\epsilon=0}
=-\nabla_\theta f(\theta^*)^{\top} \mathbf{H}_{\theta^*}^{-1}\nabla_{\theta}\mathcal{L}(x',y',\theta^*),
\end{equation}

where \(\mathbf{H}_{\theta^*}=\frac{1}{N}\sum_{(x,y)\in\mathcal{D}_{\text{train}}}\nabla^2_{\theta}\mathcal{L}(x,y,\theta^*)\) is the Hessian matrix evaluated at \(\theta^*\). 
Finally, one can approximate the influence of data point $(x',y')$ on the evaluation function $f$ through the linearization around the optimal parameter $\theta^*$ without retraining:

\begin{equation}
\label{eqn:influence_function}
f\left(\theta_{x',y',-\frac{1}{N}}^*\right)-f\left(\theta^*\right)\approx \frac{1}{N}\nabla_\theta f(\theta^*)^{\top} \mathbf{H}_{\theta^*}^{-1}\nabla_{\theta}\mathcal{L}(x',y',\theta^*).
\end{equation}

When the objective is non-convex, the Hessian can be added with a damping term $\lambda \in \mathbb{R}$, i.e., $\mathbf{H}_{\theta^*} + \lambda I$, leading to a positive definite matrix. We provide a complete derivation in \Cref{apx:if_derivation}.

\subsection{Influence function for neural networks}

Although the influence function in \cref{eqn:influence_function} reliably measures the influence of an input for a linear model with a convex objective, the computation is unreliable in practice with deep neural networks, c.f.~\citep{basu2020influence}. \citet{bae2022if} identify the three main sources of unreliability related to the standard practices of neural network training and fine-tuning: 
    1) In non-convex models, the response function is affected more by parameter initialization than by the influence of the data, as gradient methods approximate the solution.
    2) The addition of the damping term $\lambda$ works as an $\ell_2$ regularizer of \cref{eqn:empirical_minimization} leading to a different response function.
    3) The influence function is measured on fully converged parameter $\theta^*$, which is not true in practice due to many reasons, such as early stopping and over-fitting mitigation.

\citet{bae2022if} propose a new response function, named the proximal Bregman response function (PBRF), to address the three practices:

\begin{align}
\label{eqn:pbrf}
\theta_{\theta_s, x',y',\epsilon}^* \coloneq \argmin_{\theta}\frac{1}{N}\sum_{(x,y)\in\set{D}_{\text{train}}}D_{\mathcal{L}}\left(g_\theta(x), g_{\theta_s}(x), y\right)
+\frac{\lambda}{2}\left\|\theta-\theta_s\right\|^2
+\epsilon\mathcal{L}(x',y',\theta),
\end{align}

where \(g_\theta\) is a model parameterized by \(\theta\), \(\theta_s\) is a reference parameter from which the fine-tuning starts, and \(D_{\mathcal{L}}\) is the Bregman divergence defined as:

\begin{equation}
D_{\mathcal{L}}\left(h,h',y\right) \coloneq \mathcal{L}\left(h, y\right)-\mathcal{L}\left(h', y\right)-\nabla_{h} \mathcal{L}\left(h', y\right)^{\top}\left(h-h'\right).
\end{equation}

For detailed explanations of each term in \cref{eqn:pbrf}, please refer to \citet{bae2022if}.

The objective can be further linearized around the model output to simulate the local approximations made in \cref{eqn:influence_function}, leading to the following influence function\footnote{\cref{eqn:influence_function_pbrf} is originally proposed by \citet{teso2021interactive} as an approximation of \cref{eqn:influence_function}. \citet{bae2022if} provide a corresponding response function to the approximation.}: 

\begin{equation}\label{eqn:influence_function_pbrf}
f\left(\theta_{\theta_s,x',y',-\frac{1}{N}}^*\right) - f\left(\theta_s\right)\approx \frac{1}{N}\nabla_\theta f(\theta_s)^{\top} \left(\mathbf{J}_{h\theta_s}^{\top} \mathbf{H}_{h_s} \mathbf{J}_{h\theta_s}+\lambda\mathbf{I}\right)^{-1}\nabla_{\theta}\mathcal{L}(x',y',\theta_s),
\end{equation}

where \(h\) and \(h_s\) denote the model outputs parameterized by \(\theta\) and \(\theta_s\), respectively. \(\mathbf{J}_{h\theta_s}\) is the Jacobian of the model outputs with respect to the parameters, and \(\mathbf{H}_{h_s}\) is the Hessian of \(\mathcal{L}\) with respect to the outputs. 
When the loss function is convex with respect to the model outputs, the matrix \(\mathbf{J}_{h\theta_s}^{\top} \mathbf{H}_{h_s} \mathbf{J}_{h\theta_s}\) is positive semi-definite. 
This convexity condition is satisfied by commonly used loss functions, such as cross-entropy and mean squared error. 
Consequently, for \(\lambda > 0\), the matrix \(\mathbf{J}_{h\theta_s}^{\top} \mathbf{H}_{h_s} \mathbf{J}_{h\theta_s} + \lambda \mathbf{I}\) becomes positive definite, ensuring its invertibility.
We provide the complete derivation in \Cref{apx:if_derivation}.

\if\else
\subsection{Influence Function for Transductive Node Classification}

In this paper, we focus on influence functions for transductive node classification tasks in graph neural networks. We consider an undirected graph \(\mathcal{G} = \{\mathcal{V}, \mathcal{E}, \mathbf{X}\}\), where \(\mathcal{V}\) denotes the set of nodes, \(\mathcal{E}\) is the set of edges, and \(\mathbf{X} \in \mathbb{R}^{|\mathcal{V}|\times d}\) is the node feature matrix with \(d\)-dimensional features. 
Existing methods approximate the influence of removing a single edge \( e \) by formulating an objective that slightly increases the weight of edge \( e \) by a small factor \(\epsilon\)~\citep{chen2022characterizing, wu2023gif}:

\begin{equation}
\label{eqn:graph_loo}
\theta_{e,\epsilon}^*=\arg\min_{\theta}\frac{1}{N}\sum_{v\in\mathcal{V}_{\text{train}}}\mathcal{L}(h_v,y_v)+\epsilon\sum_{v \in \mathcal{V}_{\text{train}}}\left(\mathcal{L}(h_v,y_v)-\mathcal{L}(h^{-e}_v,y_v)\right),
\end{equation}
where \(h=\text{GNN}(\mathcal{G},\theta)\) and $h^{-e}=\text{GNN}(\{\mathcal{V},\mathcal{E}\setminus\{e\},\mathbf{X}\},\theta)$ represents node embeddings of the original graph and the edge-removed graph computed by the GNN. When \(\epsilon = -1/N\), the loss term computed from the original node representations \(h\) is replaced with the loss from the edge-removed representations \(h^{-e}\), simulating the removal of edge \(e\). Under this objective, the influence function is expressed as:

\begin{equation}
\label{eqn:graph_influence}
f\left(\theta_{e,-\frac{1}{N}}^*\right) - f\left(\theta^*\right)\approx\frac{1}{N}\nabla_{\theta}f(\theta^*)^{\top}\mathbf{H}_{\theta^*}^{-1}\sum_{v\in\mathcal{V}_{\text{train}}}\left(\nabla_{\theta}\mathcal{L}(h_v,y_v)-\nabla_{\theta}\mathcal{L}(h^{-e}_v,y_v)\right).
\end{equation}
\fi

%% file: tex/4.method.tex
\section{Quantifying the influence of edge edits on GNNs}
\label{sec:method}

\paragraph{Problem setup and notation} 
We propose an influence function tailored for non-convex GNNs, designed to quantify how edge deletions and insertions perturb model predictions and evaluation functions. We consider node classification on an undirected graph \(\mathcal{G} = (\mathcal{V}, \mathcal{E}, \mathbf{X})\), where \(\mathcal{V}\) is the set of nodes, \(\mathcal{E} \subseteq \mathcal{V} \times \mathcal{V} \) forms the set of edges, and \(\mathbf{X} \in \mathbb{R}^{|\mathcal{V}| \times d}\) represents the matrix of node feature \(\mathbf{X}_v \in \mathbb{R}^d\) for all node $v \in \mathcal{V}$. We also represent the graph structure using a binary symmetric adjacency matrix \(A \in \{0, 1\}^{|\mathcal{V}| \times |\mathcal{V}|}\), where \(A_{uv} = 1\) indicates an edge between nodes \(u\) and \(v\).
The goal of the node classification is to predict the ground truth label $y_v \in \mathcal{Y}$ of node $v \in \mathcal{V}$. 
The notation $\nrep{\theta}$ refers to the representation of node $v$ obtained by a GNN parameterized by $\theta$ on graph $\mathcal{G}$. A standard supervised training involves a minimization of the average prediction loss $\mathcal{L}(\nrep{\theta}, y_v)$ over the entire training set $\mathcal{V}_\text{train} \subseteq \mathcal{V}$ with respect to the model parameter $\theta$.


We focus on analyzing the effect of inserting or deleting a single undirected edge \(\{u,v\}\). 
The derivation naturally generalizes to multiple edge edits, which are provided in \Cref{apx:if_multiple}.
We first define the reweighted adjacency matrix \(\ruv{}\) such that only the \((u,v)\) and \((v,u)\) entries are updated as \(\ruv{} = A_{uv} + (2\mathbb{I}[\{u,v\} \in \mathcal{E}] - 1)N\epsilon\), while all other entries remain unchanged. Setting \(\epsilon = -1/N\) corresponds to deleting the edge if it exists, or inserting it otherwise. 
We denote the edge-reweighted graph as \(\editedgraph = \{\mathcal{V}, \mathcal{E}, \radj{}\}\), where we omit the target edge $\{u, v\}$ from the adjacency matrix when it is clear from context for notational simplicity. We collectively refer to edge deletions and insertions as \emph{edge edits}.

\paragraph{Decomposition of influence function}
Let $f(\theta, \mathcal{G})$ be an evaluation function, to which we want to measure the influence of an edge edit. Note that unlike standard parameterization of the evaluation function, i.e., $f(\theta)$, the evaluation function needs to be parameterized by both the model parameter $\theta$ and the graph structure $\mathcal{G}$, because an edge edit not only changes the model parameter but also changes the message propagation paths in GNNs.
Due to the dependency of the evaluation function on the graph structure, the derivative of the evaluation function with respect to the change in the weighting of a target edge is decomposed as follows via the chain rule:

\begin{equation}
\label{eqn:ours_grad}
\left.\frac{d f(\gpbrf{}, \editedgraph)}{d\epsilon}\right|_{\epsilon=0}
= \underbrace{
\nabla_{\theta}f(\theta^*_0,\set{G})^\top
\left.\frac{\partial \gpbrf{}}{\partial\epsilon}\right|_{\epsilon=0}
}_{\text{parameter shift}}
+\underbrace{
\left.\frac{\partial f(\theta, \editedgraph)}{\partial \radj{}} \frac{\partial \radj{}}{\partial \epsilon} \right|_{\theta=\theta^*_0,\;\epsilon=0}
}_{\text{message propagation}},
\end{equation}

where \(\gpbrf{}\) represents the response function of an edge edit; a formal definition is provided in the following paragraph.

\begin{remark} The influence functions proposed for GNNs in \citet{chen2022characterizing,wu2023gif} only consider the effect of parameter shift while missing the changes in the message propagation path.
\end{remark}

\paragraph{Parameter shift}
To quantify the change in model parameters for non-convex GNNs, we propose a graph-adapted version of the PBRF. The original PBRF,  introduced in \Cref{eqn:pbrf}, only quantifies the changes in parameters when the weight of a single data point is modified, inapplicable to our scenario where the weight of an edge changes. 
To address this challenge, we introduce an edge-edit PBRF that explicitly accounts for changes in node representations caused by an edge edit.

We define the edge-edit PBRF as follows\footnote{For notational simplicity, we omit the label \(y_v\) of node \(v\) in expressions involving the loss function or the Bregman divergence, as it is clear from context.}:

\begin{equation}
\label{eqn:graph_pbrf}
\gpbrf{} := \argmin_{\theta}\frac{1}{N}\sum_{v\in\mathcal{V}_{\text{train}}}D_{\mathcal{L}}\left(\nrep{\theta},\nrep{\theta_s}\right)
+\frac{\lambda}{2}\left\|\theta-\theta_s\right\|^2
+\sum_{v\in\mathcal{V}_{\text{train}}}\epsilon\left(\mathcal{L}\left(\nrep{\theta}\right)-\mathcal{L}\left(\nrepgme{\theta}\right)\right).
\end{equation}

The first two terms regularize \(\theta\) to stay close to the reference parameter \(\theta_s\), both in terms of the output space and the parameter space. The final term, with \(\epsilon < 0\), encourages \(\theta\) to increase the loss on the original graph while decreasing the loss on the edge-edited graph. Thus, edge-edit PBRF can be interpreted as identifying parameters near \(\theta_s\) that fail to predict correctly on the original graph but succeed on the edge-edited graph, thereby responding to the edge edit. The scalar \(\epsilon\) controls the magnitude of this response to the edge edit. Note that \(\theta^*_0 = \theta_s\), since the Bregman divergence \(D_{\mathcal{L}}(h, h', y)\) is minimized when \(h = h'\).

Based on the edge-edit PBRF objective, the  changes in the evaluation function caused by the parameter shift are then given by:

\begin{equation}
\label{eqn:retraining_effect}
\nabla_{\theta}f(\theta^*_0,\set{G})^{\top}\left.\frac{\partial \gpbrf{}}{\partial\epsilon}\right|_{\epsilon=0}
=-\nabla_{\theta}f(\theta_s,\set{G})^{\top}\mathbf{G}^{-1}\sum_{v\in\mathcal{V}_{\text{train}}}\left(\nabla_{\theta}\mathcal{L}\left(\nrep{\theta_s}\right)-\nabla_{\theta}\mathcal{L}\left(\nrepgme{\theta_s}\right)\right).
\end{equation}

where \(\mathbf{G} = \mathbf{J}_{h\theta_s}^{\top} \mathbf{H}_{h_s} \mathbf{J}_{h\theta_s} + \lambda \mathbf{I}\) denotes the generalized Gauss–Newton Hessian with a damping term, and \(h\) and \(h_s\) denote the node representations obtained using parameters \(\theta\) and \(\theta_s\), respectively.
A detailed derivation is provided in \Cref{apx:pbrf_derivation}.

\paragraph{Message propagation}
To quantify the changes in the evaluation function caused by the modification in the message propagation path, we can further expand the message propagation term in \cref{eqn:ours_grad}.
Since \(\partial A^{\epsilon}_{ij} / \partial \epsilon = 0\) for all \(\{i,j\} \ne \{u,v\}\), the message propagation term simplifies as follows:

\begin{align}
\label{eqn:propagation_effect}
\left.\frac{\partial f(\theta, \editedgraph)}{\partial \radj{}} \frac{\partial \radj{}}{\partial \epsilon} \right|_{\theta=\theta^*_0,\;\epsilon=0}
&= \left.\frac{\partial f(\theta,\editedgraph)}{\partial \ruv{}}\frac{\partial \ruv{}}{\partial \epsilon}\right|_{\theta=\theta_s,\,\epsilon=0}
+ \left.\frac{\partial f(\theta,\editedgraph)}{\partial \rvu{}}\frac{\partial \rvu{}}{\partial \epsilon}\right|_{\theta=\theta_s,\,\epsilon=0} \notag\\
&= (2\mathbb{I}[\{u,v\}\in\mathcal{E}]-1)\,N\left(\frac{\partial f(\theta_s,\set{G})}{\partial A_{uv}}+\frac{\partial f(\theta_s,\set{G})}{\partial A_{vu}}\right).
\end{align}

\paragraph{Unified influence function under edge edits}

By substituting \Cref{eqn:propagation_effect} and \Cref{eqn:retraining_effect} into \Cref{eqn:ours_grad} and linearizing around \(\epsilon = 0\), we obtain a first-order approximation of the influence function that captures both parameter shift and message propagation effects. The resulting influence of an edge edit is given by:

\begin{align}
\label{eqn:influence_edge_removal}
f\left(\theta_{-\frac{1}{N}}^*,\set{G}^{-\frac{1}{N}}\right) - f(\theta_s,\set{G}) 
&\approx \frac{1}{N}\nabla_{\theta}f(\theta_s,\set{G})^\top\mathbf{G}^{-1}\sum_{v\in\mathcal{V}_{\text{train}}}\left(\nabla_{\theta}\mathcal{L}\left(\nrep{\theta_s}\right)-\nabla_{\theta}\mathcal{L}\left(\nrepgme{\theta_s}\right)\right) \notag \\[6pt]
&\quad -(2\mathbb{I}[\{u,v\}\in\mathcal{E}]-1)\left(\frac{\partial f(\theta_s,\set{G})}{\partial A_{uv}}+\frac{\partial f(\theta_s,\set{G})}{\partial A_{vu}}\right).
\end{align}

Directly computing \(\mathbf{G}^{-1}\) is computationally infeasible for large models. To address this, we approximate the inverse Hessian-vector product \(\mathbf{G}^{-1}\nabla_{\theta}f(\theta_s,\set{G})\) using the LiSSA algorithm~\citep{agarwal2016second}, a stochastic iterative method. A detailed description is provided in \Cref{apx:lissa}.


%% file: tex/5.experiments.tex
\section{Validation of influence function}
\label{sec:experiments}
We measure the correctness of the proposed influence function on three different evaluation metrics: over-squashing and over-smoothing measures, and a validation loss. Our goal is to precisely predict how much over-squashing and over-smoothing measures, and validation loss change when an existing edge is deleted from the graph or when a potential edge is added between two nodes.

\subsection{Evaluation functions}
We explain the three evaluation functions considered in detail.

\paragraph{Over-squashing}
Over-squashing~\citep{alon2020bottleneck} is the phenomenon in which information from distant nodes is overly compressed during message passing, preventing it from effectively influencing node representations.
\citet{topping2021understanding} propose a gradient-based metric \(\partial h_v / \partial \mathbf{X}_u\) to quantify the influence of the initial node feature $\mathbf{X}_u$ on the node representation $h_v$. One way to measure over-squashing in a graph is by averaging the gradients between all pairs of distant nodes. However, computing these gradients for every such pair is computationally expensive. Moreover, estimating the influence function requires taking the derivative of this measurement, which is even more costly due to the complexity of the measurement itself.
To address this issue, we propose an alternative over-squashing measure similar to the gradient-based one but without derivation. 
Let $\mathcal{N}_L(v)$ be the set of nodes that can be reached from node $v$ in exactly $L$ hops. To measure the influence of node $u$ to node $v$ in an $L$-layer GNN, we first define modified graph \(\mathcal{G}'(v)=\{\mathcal{V},\mathcal{E},\mathbf{X}'\}\), where

\begin{equation}
\mathbf{X}'_u=\begin{cases}
\vec{0}, & \text{if } u \in \mathcal{N}_L(v), \\
\mathbf{X}_u, & \text{otherwise,}
\end{cases}
\end{equation}

for all $u$ in $\mathcal{V}$. With the modified graph, we propose a new over-squashing measure as:

\begin{equation}
\label{eqn:eval_osq}
\oq(\theta, \set{G}) = \sum_{v \in \mathcal{V}} 
\left\| \nrep{\theta} - \nrepfull{v}{\set{G}'(v)}{\theta}  \right\|_{2}.
\end{equation}

The measure computes the average difference in node representations with and without $L$-hop neighborhood node features. It is similar to the gradient-based metric in that both quantify how a node’s representation changes when input features of other nodes are modified, although one does so through gradient computation and the other through direct input masking.

\paragraph{Over-smoothing}
We use the Dirichlet energy that quantifies the over-smoothing phenomenon in GNNs \citep{li2018deeper}. Dirichlet energy is defined as the average squared $\ell_2$ distance between the embeddings of adjacent nodes \citep{cai2020note}, and serves as a proxy for the representational diversity across the graph. A lower Dirichlet energy indicates more severe over-smoothing, as node representations become increasingly indistinguishable due to excessive message passing. Conversely, higher Dirichlet energy suggests that node embeddings remain more discriminative.


\input{fig/figure/vs_ours}

\paragraph{Validation loss} We use a standard mean cross-entropy loss on the validation set as an evaluation function.



\subsection{Actual vs. predicted influence}
\paragraph{Datasets and experimental setup}

We conduct experiments on five datasets: the citation graphs Cora, Citeseer, and Pubmed~\citep{sen2008collective, yang2016revisiting}, where the task is to predict each paper's research area based on citation relationships; and the Wikipedia graphs Chameleon and Squirrel~\citep{rozemberczki2021multi}, where the task is to estimate page traffic based on hyperlink relationships. For the citation graphs, we follow the data splits provided by~\citet{yang2016revisiting}, and for the Wikipedia graphs, we use the splits from~\citet{pei2020geom}.

We evaluate the prediction performance of our method on three representative graph neural networks: GCN~\citep{kipf2016semi}, GAT~\citep{velivckovic2017graph}, and ChebNet~\citep{defferrard2016convolutional}. 
We compare our approach with GIF, an existing graph influence function~\citep{chen2022characterizing, wu2023gif}. 
To measure the actual influence, we first train the model on the original graph. We then retrain the GNN on the edge-edited graph by optimizing the minimization objective defined by each response function. Specifically, for our influence function, we fine-tuned the model by minimizing the objective in \Cref{eqn:graph_pbrf}, using the original model parameters as \(\theta_s\) and setting \(\epsilon = -1/N\). For GIF, we retrain the model by minimizing the loss on the edge-edited graph, using the same parameter initialization as the original model. The actual influence is computed as the difference between the evaluation values of the original model and the retrained model.

\paragraph{Results}
\Cref{fig:vs_ours} presents scatter plots for the Cora dataset, where each point shows the actual and predicted influence of a single edge edit. The x-axis represents the predicted influence computed by each method, and the y-axis represents the actual influence. The first row illustrates results from GIF~\citep{chen2022characterizing, wu2023gif}, while the second row shows results from our proposed method for edge deletions and insertions, respectively. Additional results for other GNNs and datasets are provided in \Cref{apx:other_gnns}.

GIF~\citep{chen2022characterizing, wu2023gif} fails to reliably estimate actual influences on non-convex GNNs, achieving correlations of only $0.09$ and $0.14$ for over-squashing and Dirichlet energy, respectively. In contrast, our proposed method significantly improves prediction, achieving correlations up to $0.95$, closely aligning with the ideal predictions shown by the grey dotted line. Moreover, our method exhibits strong predictive capability for both edge deletions and insertions.

\input{fig/table/rewiring_result}

We further evaluate how editing the graph to improve each measurement affects prediction on the test nodes. To perform the edge edits, we first compute the influence function with respect to each measurement and then edit the edges that are predicted to improve the corresponding measurement. For validation loss, we edit the edges with the $k$ smallest influence values, as negative influence indicates that removing the edge is expected to reduce the measurement. For the other measurements, we edit the edges with the $k$ largest influence values. The value of \(k\) is selected to maximize validation accuracy. We compare the edge edits produced by our influence function to two baselines: random edge deletion (Random) and edge edits using GIF~\citep{chen2022characterizing,wu2023gif} in place of our influence function.

\Cref{tab:test_acc} shows the test accuracy when the model is trained and evaluated on edge-edited graphs. Dirichlet energy (DE), over-squashing ($\oq$), and validation loss (VL) are used as shorthand in the table. 
Our edits based on validation loss achieve the best performance across three datasets. 
This result suggests that editing edges to reduce validation loss can be helpful for improving test accuracy. In contrast, editing edges to improve Dirichlet energy or the over-squashing measurement does not consistently improve test accuracy. 
This indicates that optimizing these intermediate metrics does not necessarily translate to better predictive performance. Finally, editing edges to reduce validation loss using GIF fails to improve test accuracy, 
which we attribute to its inaccurate influence estimation on non-convex GNNs.

\paragraph{Analysis on the influence of message propagation}

\input{fig/figure/retraining_vs_perturbing}

We analyze the importance of explicitly incorporating the influence of message propagation when predicting the influence of edge edits. If the influence of the message propagation is negligible or highly correlated with the influence of the parameter shift, predicting the influence of the message propagation may not be necessary. To validate the importance of the message propagation influence, we measure its correlation with the influence of parameter shift. 

\Cref{fig:retraining_vs_perturbing} shows scatter plots comparing these two influences on the Cora dataset. We observe a low correlation between the message propagation and parameter shift influence across all metrics, while their magnitudes remain comparable. These results underscore the necessity of explicitly measuring the influence of message propagation in estimating the influence of edge edits.

\paragraph{Influence estimation for multiple edge edits}
\input{fig/figure/multiple_edge_edit}

We evaluate the accuracy of the proposed influence function in predicting the actual influence under multiple edge edits. \Cref{fig:multiple_edges} presents scatter plots of predicted versus actual influence when inserting different numbers of edges. The experiments are conducted on the Cora dataset, with validation loss used as the evaluation metric. Specifically, we report results for simultaneous insertions of 10, 20, and 100 edges.

Our influence function maintains a high correlation with the actual influence, achieving 0.84 even under 100 simultaneous edge edits. Nonetheless, \Cref{fig:multiple_edges} shows that the accuracy of influence estimation diminishes as the number of simultaneous edits increases. We attribute this degradation to first-order approximation errors induced by substantial parameter shifts, a phenomenon that has also been consistently observed in prior studies on group influence estimation~\citep{koh2019accuracy,basu2020second,song2023rge}.

%% file: fig/figure/vs_ours.tex
\begin{figure}[t]
    \centering

    \begin{subfigure}[b]{\textwidth}
        \centering
        \includegraphics[width=0.32\linewidth]{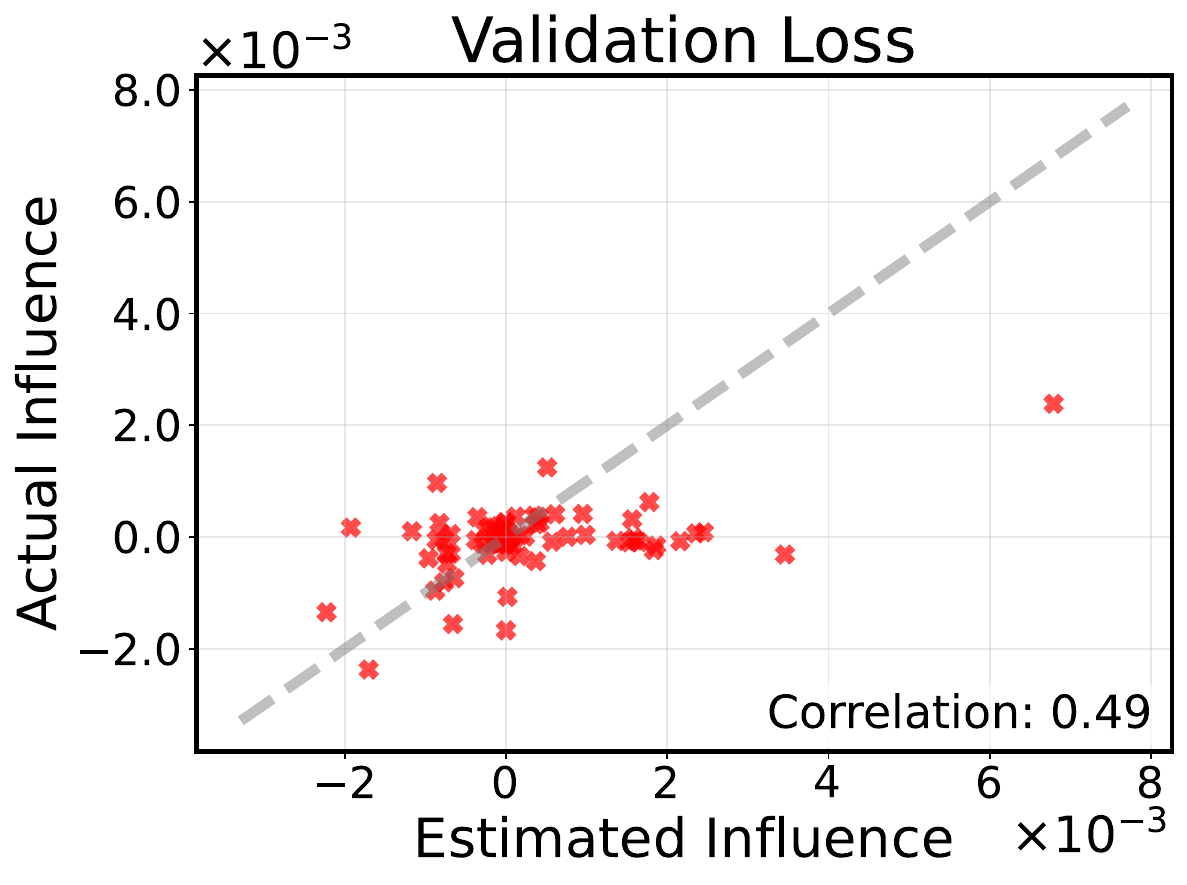}
        \includegraphics[width=0.32\linewidth]{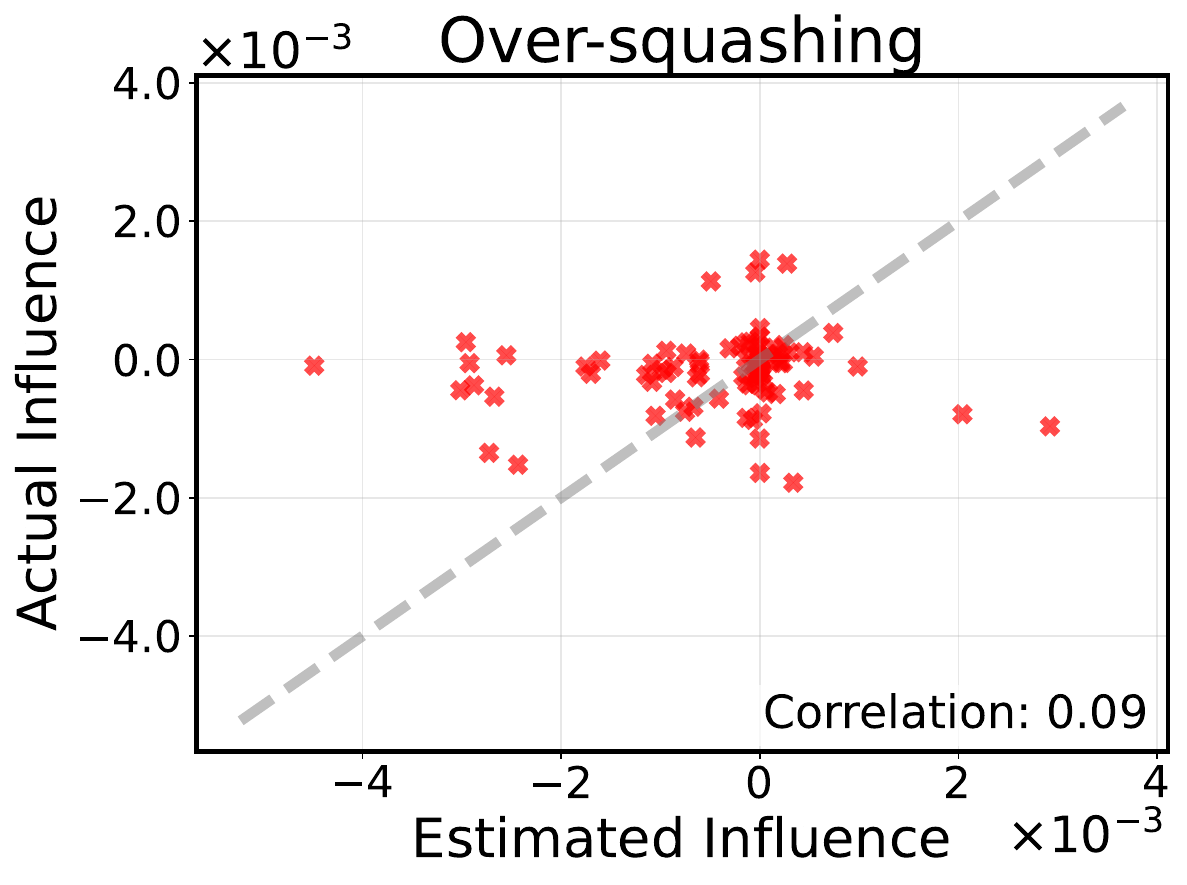}
        \includegraphics[width=0.32\linewidth]{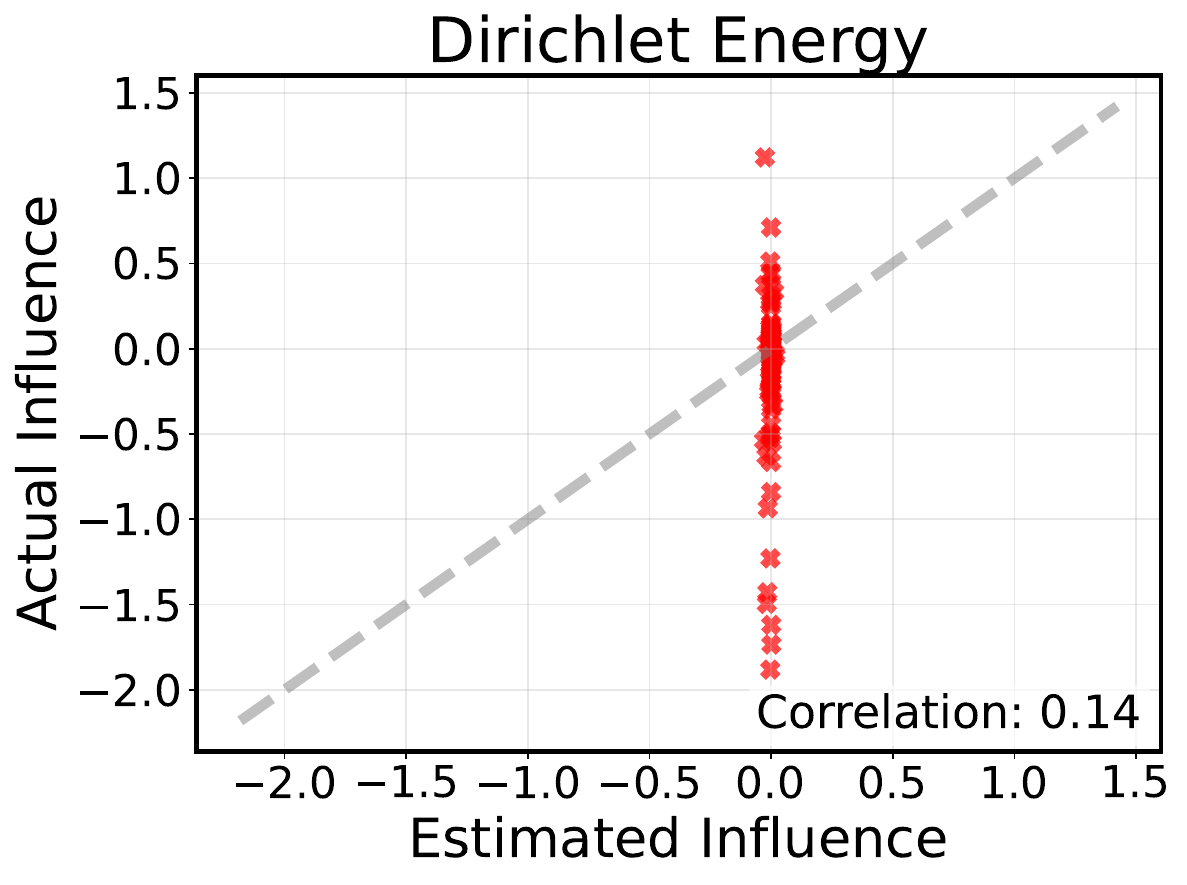}
        \caption{GIF~\citep{chen2022characterizing, wu2023gif}}
    \end{subfigure}

    \begin{subfigure}[b]{\textwidth}
        \centering
        \includegraphics[width=0.32\linewidth]{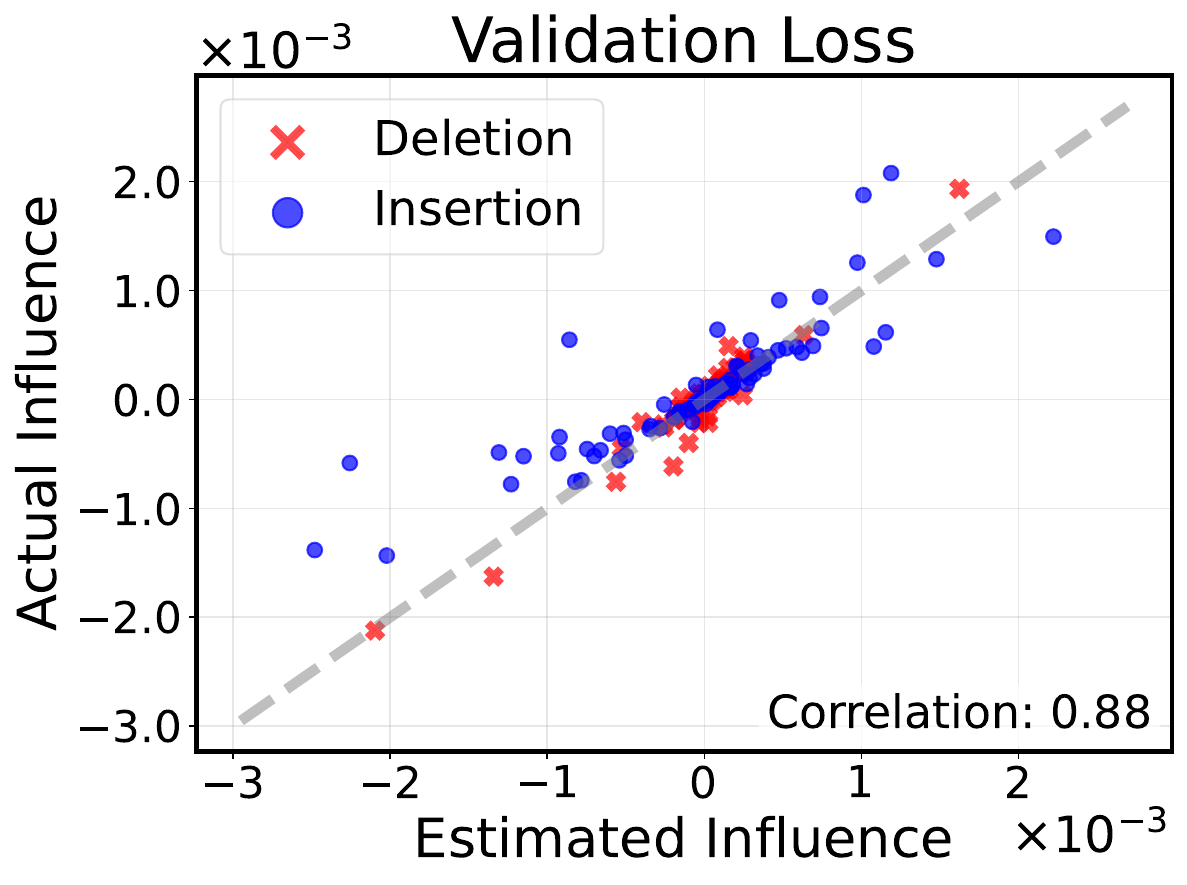}
        \includegraphics[width=0.32\linewidth]{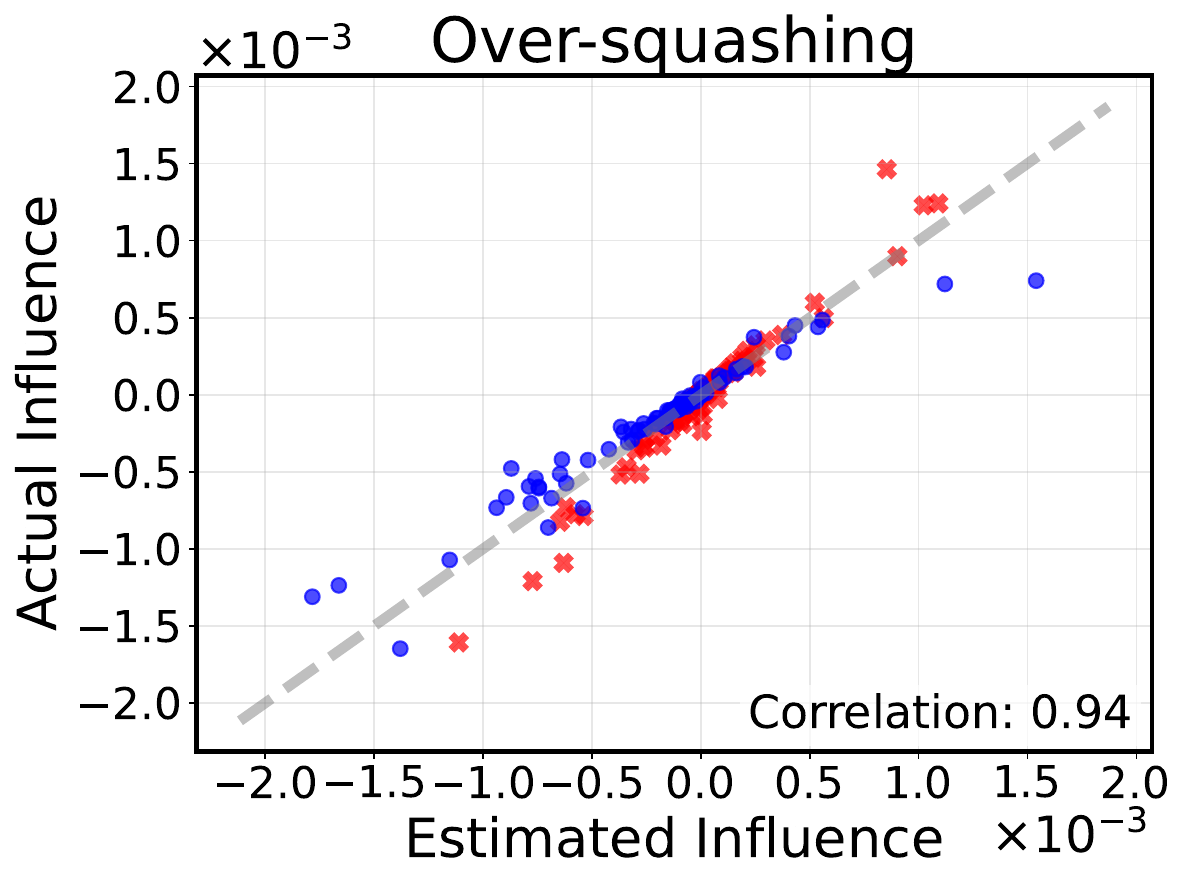}
        \includegraphics[width=0.32\linewidth]{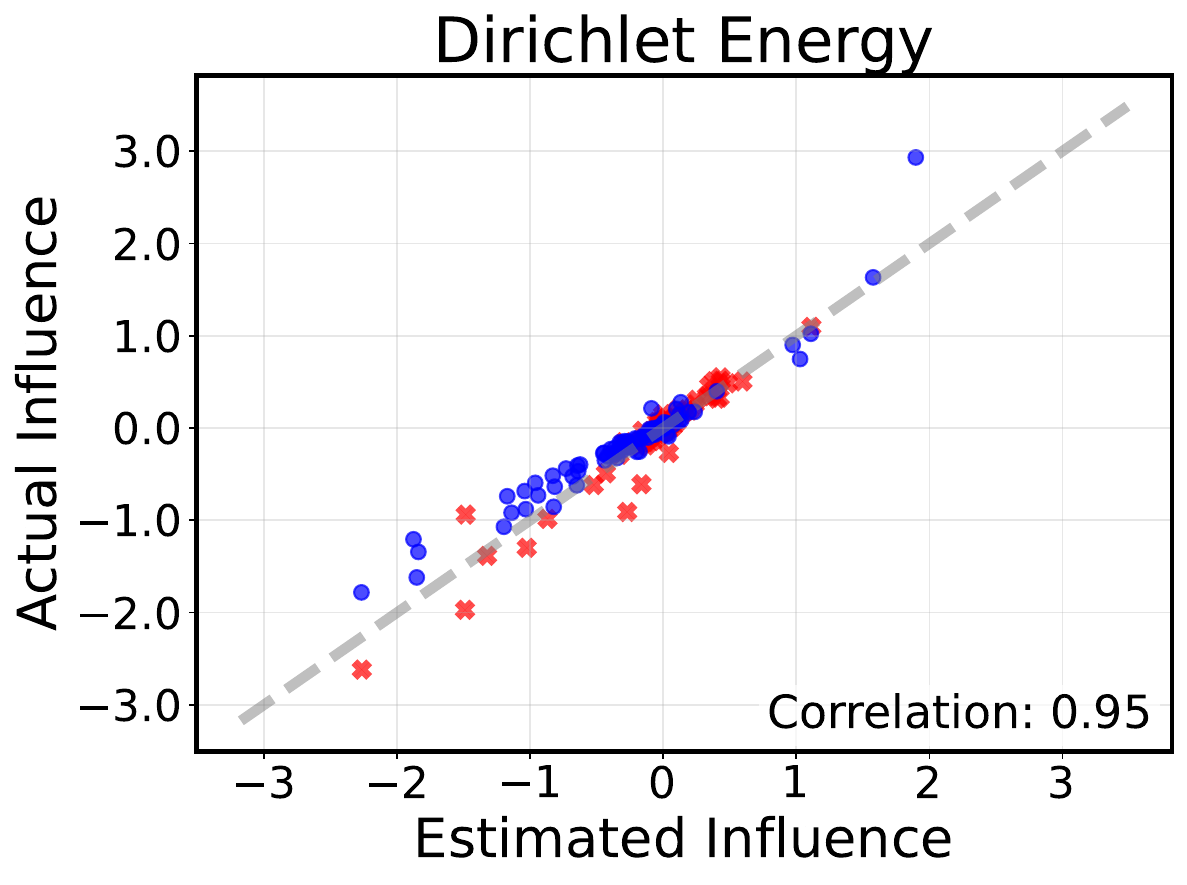}
        \caption{\ours{}}
    \end{subfigure}
    
    \caption{Predicted influence versus actual influence on a four-layer GCN. The x-axis represents the predicted influence, the y-axis represents the actual influence, and the dotted line represents the perfect alignment. 
    }
    \label{fig:vs_ours}
\end{figure}

%% file: fig/table/rewiring_result.tex
\begin{wraptable}{r}{0.45\linewidth}
    \caption{Test accuracy on edge-edited graphs. The best result is highlighted in bold.}
    \centering
    \resizebox{\linewidth}{!}{
        \begin{tabular}{lccc} 
            \toprule 
            & Cora & CiteSeer & PubMed \\
            \midrule
            GCN & \tabnum{81.0}{0.3} & \tabnum{69.3}{0.5} & \tabnum{75.6}{1.0} \\
            Random & \tabnum{81.1}{0.4} & \tabnum{69.2}{0.4} & \tabnum{75.7}{0.8} \\
            GIF & \tabnum{80.9}{0.5} & \tabnum{69.2}{0.5} & \tabnum{75.6}{0.9} \\
            \ours{} (DE) & \tabnum{80.8}{0.4} & \tabnum{69.5}{0.5} & \tabnum{75.4}{1.2} \\
            \ours{} ($\oq$) & \tabnum{81.1}{0.4} & \tabnum{69.3}{0.5} & \tabnum{75.4}{1.0} \\
            \ours{} (VL) &  \besttabnum{82.1}{0.5} & \besttabnum{69.6}{0.7} & \besttabnum{76.4}{1.3}\\
            \bottomrule
        \end{tabular}
    }
    \label{tab:test_acc}
\end{wraptable}

\if\else
\begin{table*}[t!]
    \caption{Test accuracy on edge-edited graphs. We report the performance of edge selection strategies, including random deletions (Random), GIF~\citep{chen2022characterizing, wu2023gif}, and our proposed influence-based methods targeting Dirichlet energy (DE), over-squashing ($\oq$), and validation loss. The best result for each dataset is highlighted in bold.}
    \centering
    \resizebox{0.6\linewidth}{!}{
        \begin{tabular}{lcccccc} 
             \toprule 
             & \multirow{2}{*}{GCN} & \multirow{2}{*}{Random} & \multirow{2}{*}{GIF} & \multicolumn{3}{c}{\ours} \\
             \cmidrule(lr){5-7}
             & & &  & DE & $\oq$ & Val loss \\
             \midrule
             Cora & \tabnum{81.0}{0.3} & \tabnum{81.1}{0.4} & \tabnum{80.9}{0.5} & \tabnum{80.8}{0.4} &\tabnum{81.1}{0.4} & \besttabnum{82.1}{0.5} \\
             CiteSeer & \tabnum{69.3}{0.5} & \tabnum{69.2}{0.4} & \tabnum{69.2}{0.5} & \tabnum{69.5}{0.5} &\tabnum{69.3}{0.5} & \besttabnum{69.6}{0.7} \\
             PubMed & \tabnum{75.6}{1.0} & \tabnum{75.7}{0.8} & \tabnum{75.6}{0.9} & \tabnum{75.4}{1.2} &\tabnum{75.4}{1.0} & \besttabnum{76.4}{1.3} \\
             \bottomrule
        \end{tabular}
    }
    \label{tab:test_acc}
\end{table*}
\fi

%% file: fig/figure/retraining_vs_perturbing.tex
\begin{figure}[t]
    \centering
    \begin{subfigure}[b]{0.32\textwidth}
        \centering
        \includegraphics[width=\linewidth]{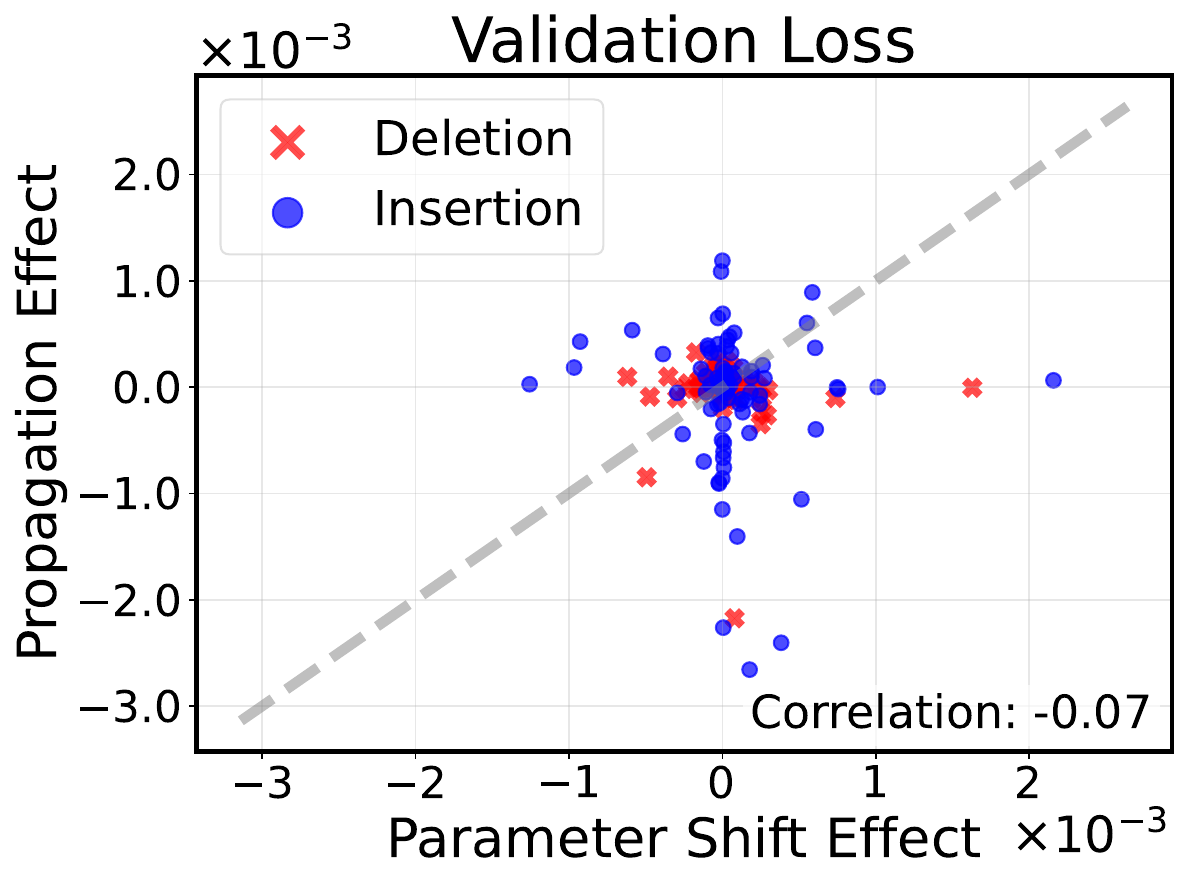}
    \end{subfigure}
    \begin{subfigure}[b]{0.32\textwidth}
        \centering
        \includegraphics[width=\linewidth]{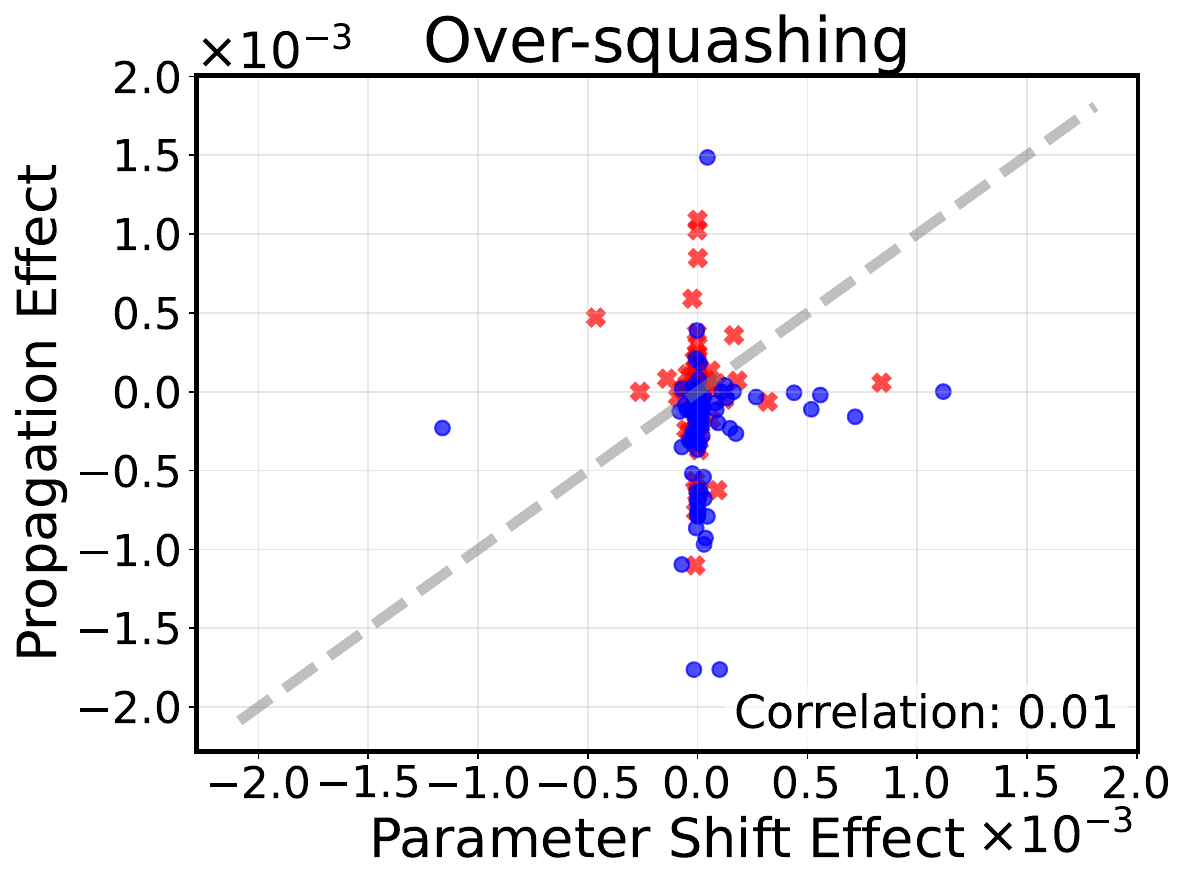}
    \end{subfigure}
    \begin{subfigure}[b]{0.32\textwidth}
        \centering
        \includegraphics[width=\linewidth]{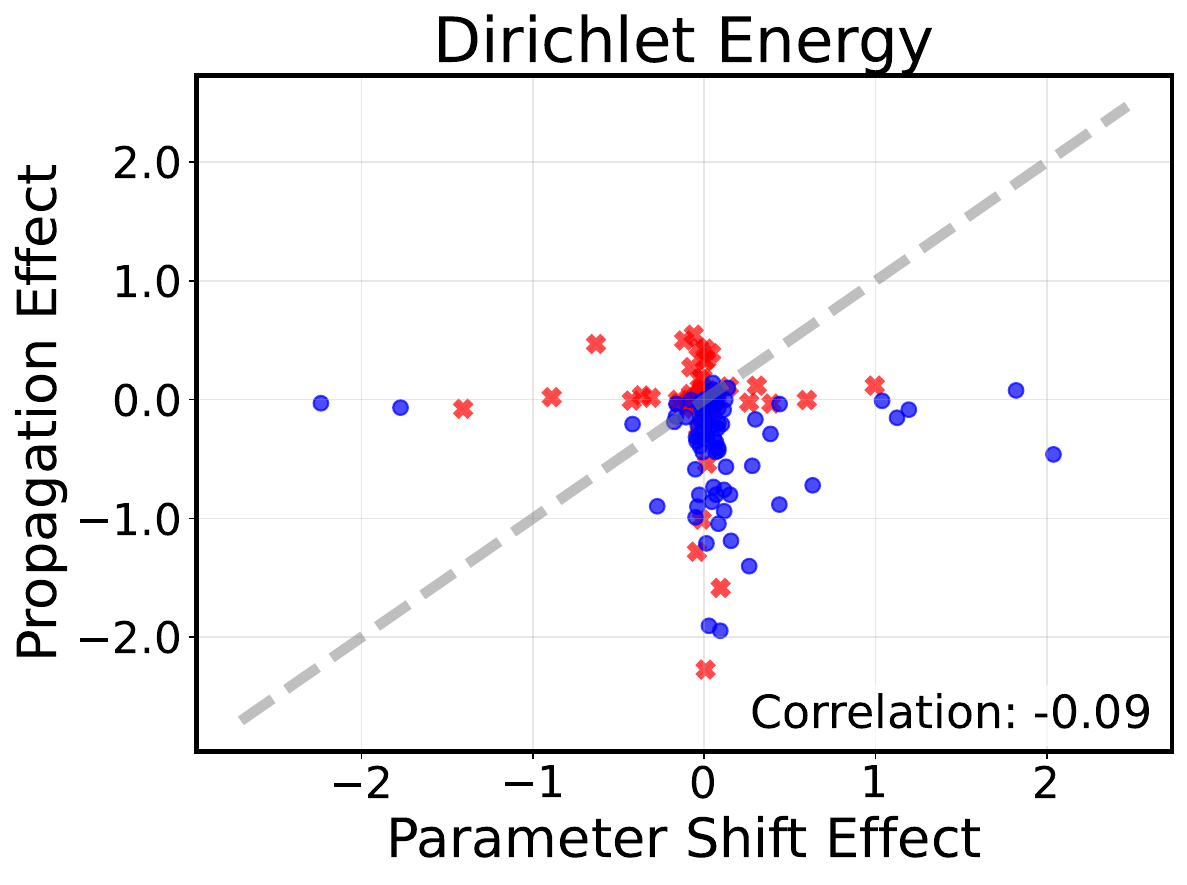}
    \end{subfigure}

    \caption{The relationship between parameter shift effect and message propagation effect defined in \Cref{eqn:ours_grad}. The x-axis denotes the parameter shift effect, and the y-axis denotes the message propagation effect.}
    \label{fig:retraining_vs_perturbing}
\end{figure}

%% file: fig/figure/multiple_edge_edit.tex
\begin{figure}[t]
    \centering
    \begin{subfigure}[b]{0.32\textwidth}
        \centering
        \includegraphics[width=\linewidth]{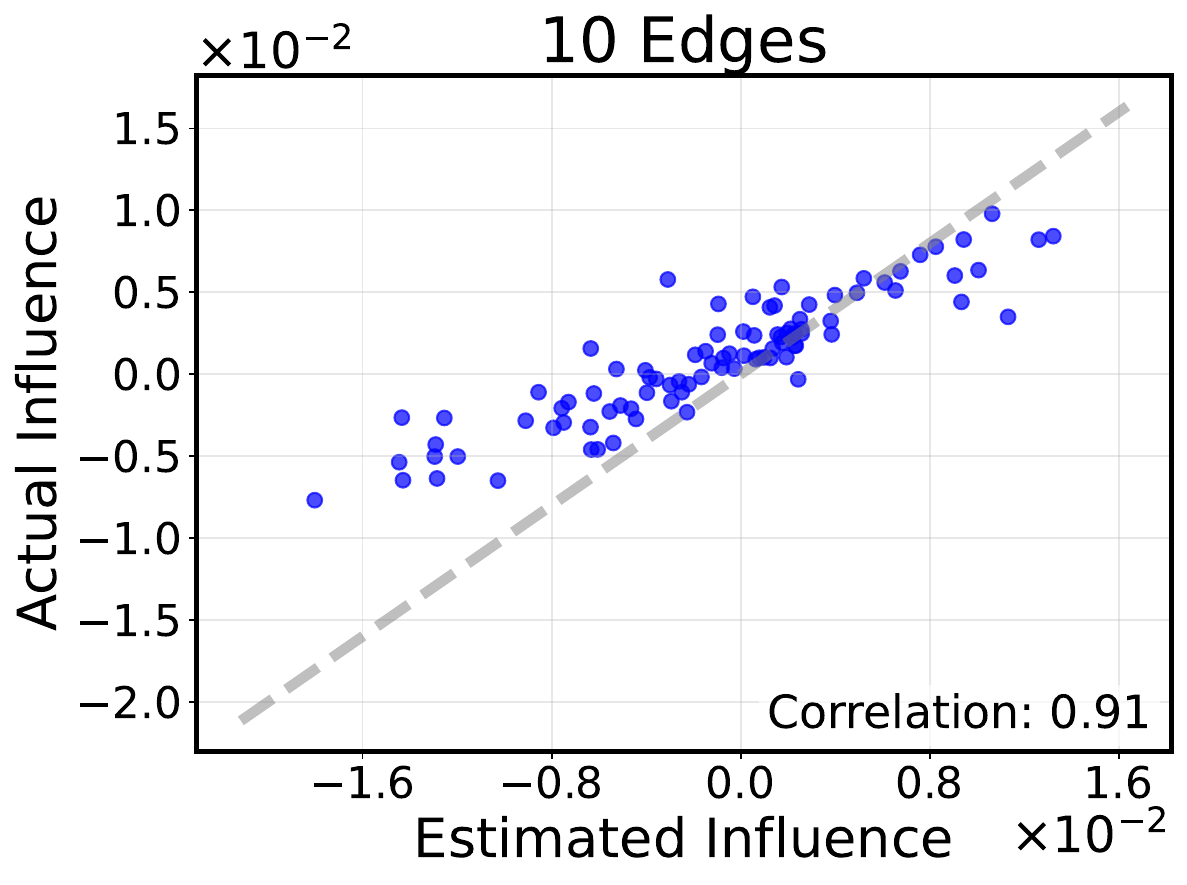}
    \end{subfigure}
    \begin{subfigure}[b]{0.32\textwidth}
        \centering
        \includegraphics[width=\linewidth]{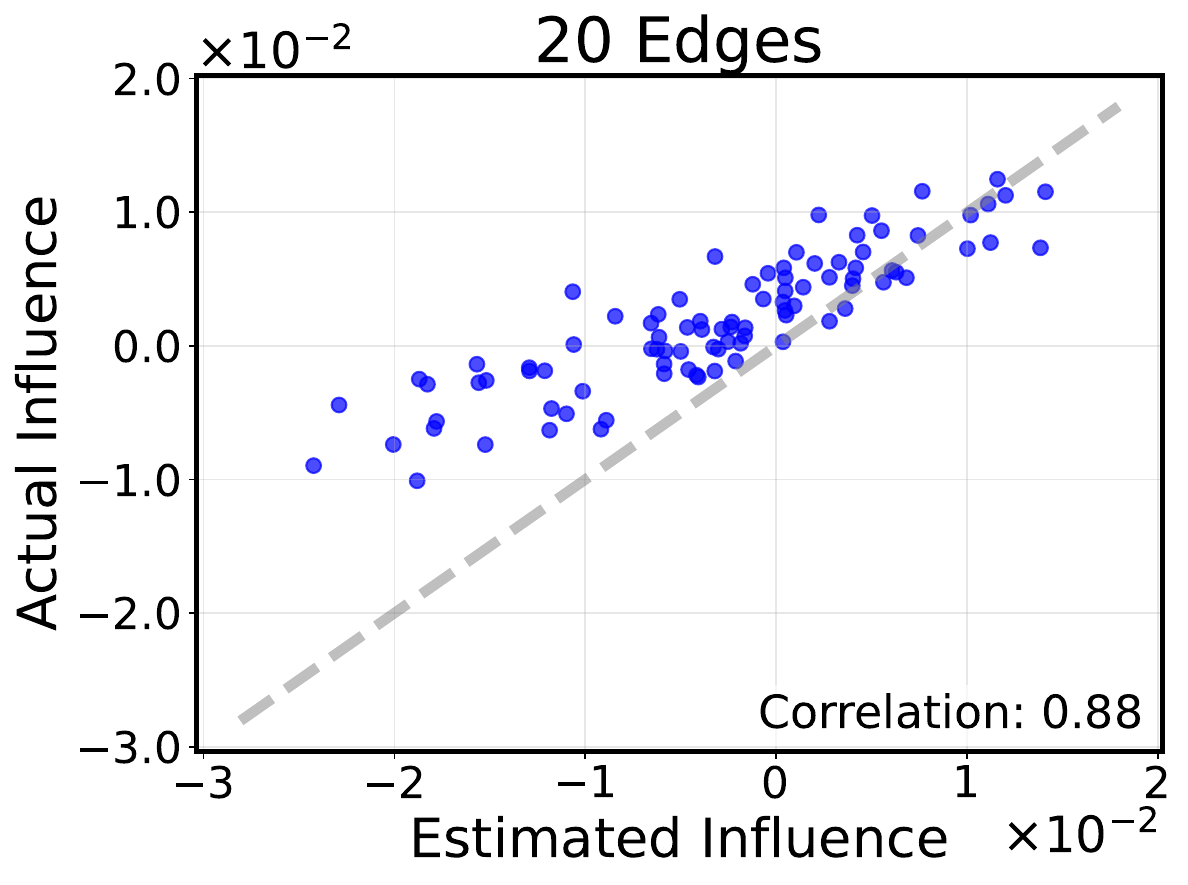}
    \end{subfigure}
    \begin{subfigure}[b]{0.32\textwidth}
        \centering
        \includegraphics[width=\linewidth]{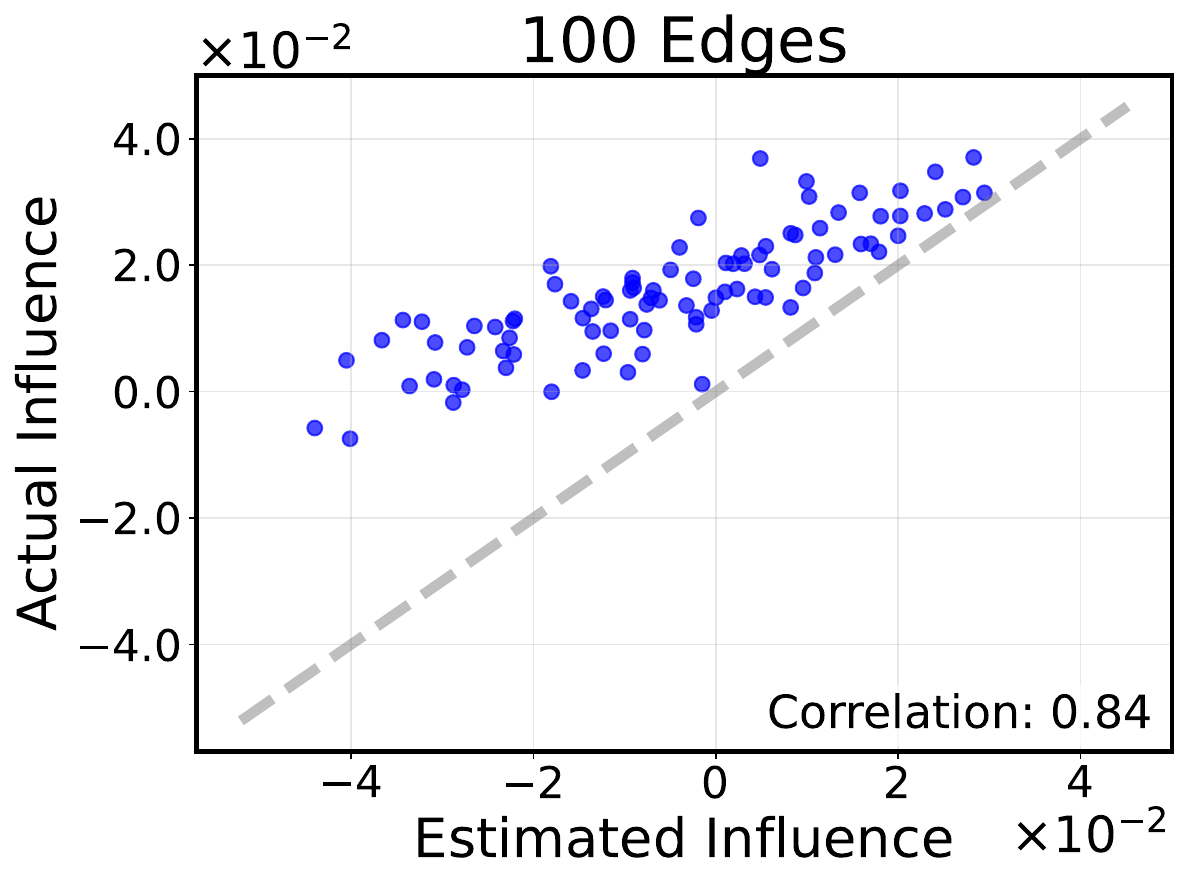}
    \end{subfigure}

    \caption{Predicted influence versus actual influence on a four-layer GCN under varying numbers of inserted edges.}
    \label{fig:multiple_edges}
\end{figure}

%% file: tex/6.applications.tex
\section{Applications}
\label{sec:applications}

\paragraph{Influence function as a tool for adversarial attack}
\input{fig/table/attack_result}
Once we identify the most influential edges in the entire input graph, we can significantly affect the model's output by editing them.
This process represents a form of global adversarial attack for the graph, aiming to reduce the performance of the entire graph.

\cref{tab:attack} presents the adversarial attack performance of our methods, comparing it to previous influence-function-based methods and other adversarial attack methods, DICE~\citep{waniek2018hiding} and PRBCD~\citep{NEURIPS2021_3ea2db50}, in the white-box scenario.
As a result, the GIF shows poor performance, while our method with a validation loss metric outperforms the others.
An interesting observation is that, among the three metrics we tested, the validation loss-based attack is more effective than the tailored attack methods.

\paragraph{Explanation of the characteristics of beneficial edges}

\input{fig/figure/avg_influences}

We demonstrate the influence of edges connecting the node with the same label (\emph{homophilic edge}) and different labels (\emph{heterophilic edge}) on GCN.
\Cref{fig:homophily_analysis} presents a dumbbell plot showing the mean influence of homophilic and heterophilic edges across six datasets, which include three homophilic graphs (Cora, CiteSeer, and PubMed), where edges tend to connect nodes with the same labels, and three heterophilic graphs (Chameleon, Actor, and Squirrel), where edges tend to connect nodes with different labels.

For both types of graphs, adding homophilic edges is more beneficial than adding heterophilic ones. The opposite effect is observed for edge deletion. These findings suggest that increasing a graph's homophily is beneficial, which aligns with the homophilic nature of the tested GNNs.

\paragraph{Analysis of the effect of edge rewiring}

\input{fig/figure/histogram_rewiring}

Our approach provides multiple analytical perspectives on existing rewiring strategies through various evaluation metrics. 
Specifically, we analyze BORF~\citep{nguyen2023revisiting} and FoSR~\citep{karhadkar2023fosr}. BORF alleviates over-squashing by inserting edges between nodes with negative curvature and mitigates over-smoothing by removing edges with positive curvature, while FoSR alleviates over-squashing by inserting edges that enlarge the spectral gap.

\Cref{fig:histogram_rewiring} presents the estimated influence of edges selected by BORF and FoSR from three measurement perspectives. Edge insertions chosen by both methods generally increase the over-squashing measurement, while edge deletions in BORF tend to increase the over-smoothing measurement, confirming that each method effectively targets its intended GNN challenge. Nevertheless, we also observe unintended side effects: edge insertions often exacerbate over-smoothing, and neither insertions nor deletions consistently reduce validation loss. Mitigating these side effects could substantially enhance the overall effectiveness and reliability of edge rewiring methods.


\if\else
Our approach provides multiple analytical perspectives on existing rewiring strategies through various evaluation metrics. 
We analyze two representative rewiring methods. BORF~\citep{nguyen2023revisiting} alleviates over-squashing by inserting edges between nodes with negative curvature and reduces over-smoothing by removing edges with positive curvature. FoSR~\citep{karhadkar2023fosr} addresses over-squashing by inserting edges that increase the spectral gap of the graph.

\Cref{fig:histogram_rewiring} shows the estimated influence of edges selected by these methods from three different measurement perspectives. Edge insertions selected by both methods tend to increase the over-squashing measurement, while edge deletions by BORF tend to increase the over-smoothing measurement. These results indicate that both methods effectively target their intended GNN challenges. However, we also observe unintended side effects: edge insertions from both methods tend to worsen the over-smoothing measurement, and their operations do not consistently reduce validation loss. Addressing these unintended effects could significantly improve the overall effectiveness and reliability of these rewiring strategies.
\fi




\if\else
\subsection{Improving Target Measurements via Edge Rewiring}

We have proposed a method to quantify the impact of edge insertion or removal on specific target measurements, enabling systematic selection of edges expected to improve these metrics. We demonstrate the efficacy of our approach by editing the top-$k\%$ edges predicted to have the highest helpful influence. Specifically, for validation loss, we select edges with the most negative influence, as negative influence suggests that editing these edges will reduce validation loss. Conversely, for over-squashing and Dirichlet energy, we select edges with the highest positive influence, as positive influence indicates that editing these edges will enhance these metrics.

\Cref{fig:naive_re_wiring} illustrates how target measurements change after edge editing. The upper figure presents results from edge removal, and the lower figure shows results from edge insertion. The blue line indicates performance after editing the top-$k\%$ positively influential edges identified by our method, with a ratio of $1.0$ indicating that all positively influential edges are edited. The orange and green lines show the performance after random edge editing and edge editing using the existing influence function, respectively, matching the number of edges edited by our approach at each corresponding ratio.

\input{fig/figure/naive_re_wiring}

Random edge editing degrades performance in almost all cases, likely due to unintended disruption of informative connections and essential node paths. An exception is the improvement in Dirichlet energy observed for random edge removal, aligning with prior findings that random edge removal can alleviate over-smoothing~\citep{rong2019dropedge}. Additionally, since the existing influence function inaccurately estimates actual influences, edge editing guided by it often fails to consistently enhance the target metrics. In contrast, selectively removing or inserting the top-$k\%$ positively influential edges identified by our method substantially improves these target measurements. These results demonstrate that our method reliably identifies influential edges, enabling targeted edge editing that substantially improves GNN performance.

\todo{In progress...}

We further evaluate whether graph rewiring based on the proposed influence function can improve test accuracy. We use the mean validation loss as the evaluation metric. Since certain edge edits that improved validation loss did not necessarily improve test loss, we first divide the validation set into five subsets and then filter out edge edits that are predicted to consistently improve validation accuracy across all subsets. After filtering, we select the top-$k$ beneficial edge edits to perform graph rewiring. \Cref{tab:rewiring_result} reports the classification accuracy for a two-layer GCN. We compare performance on the original graph, after random edge removal, and after rewiring using the baseline method by~\citet{chen2022characterizing}.

Our method improves test accuracy across all three datasets, indicating that using validation loss as a proxy for test loss is effective to some extent. Meanwhile, random edge removal or edge rewiring based on inaccurate influence predictions negatively affects accuracy.
\fi

\if\else
\input{fig/figure/kfold_result}

However, in many cases, the target measurement of interest is the performance on test nodes (e.g., test accuracy), which cannot be evaluated beforehand; thus, the influence function cannot be directly computed. \citet{} propose improving test performance indirectly by improving validation loss. While this approach is commonly used, assuming strong generalization from validation to test sets, our experiments reveal that this assumption does not necessarily hold in practice.

The blue line in \Cref{fig:kfold_result} shows the mean test loss when edges are removed or inserted based on our method, using mean validation loss as the evaluation function. Despite achieving a significant reduction in validation loss, test loss increases, indicating that improvements in validation loss do not necessarily translate to enhanced test performance.

To mitigate this issue, we propose a simple yet effective method that identifies edges consistently exerting positive influence across multiple validation subsets rather than on a single validation set. Specifically, we partition the validation nodes into $k$ subsets and remove only those edges exhibiting positive influence across all subsets. \Cref{fig:kfold_result} illustrates the mean test loss for varying numbers of subsets $k$. By constraining edge editing in this manner, improvements observed in validation loss more reliably generalize to the test loss.
\fi

%% file: fig/table/attack_result.tex
\begin{table*}[t!]
    \caption{Test accuracy under adversarial edge edits. The attack ratio denotes the percentage of edges added or removed with respect to the total number of edges in the original graph. The values in parentheses next to each dataset name indicate the test accuracy on the original graph before any edits.}
    \centering
    \resizebox{0.9\linewidth}{!}{
        \begin{tabular}{lccccccccc} 
             \toprule 
             & \multicolumn{3}{c}{Cora (81.0)} & \multicolumn{3}{c}{CiteSeer (69.3)} & \multicolumn{3}{c}{PubMed (75.6)} \\
             \cmidrule(lr){2-4}\cmidrule(lr){5-7}\cmidrule(lr){8-10}
             Attack ratio & 1\% & 3\% & 5\% & 1\% & 3\% & 5\% & 1\% & 3\% & 5\% \\ 
             \midrule
             DICE   & \tabnum{80.8}{0.4} & \tabnum{80.3}{0.5} & \tabnum{80.0}{0.5}
                    & \tabnum{69.0}{0.7} & \tabnum{68.5}{0.5} & \tabnum{68.1}{0.5}
                    & \tabnum{75.1}{0.9} & \tabnum{74.5}{1.4} & \tabnum{73.5}{1.1} \\
             PRBCD  & \tabnum{80.6}{0.5} & \tabnum{79.4}{0.5} & \tabnum{78.7}{0.9}
                    & \tabnum{68.5}{0.5} & \tabnum{67.4}{0.8} & \tabnum{66.4}{0.5}
                    & \tabnum{74.6}{1.3} & \tabnum{72.9}{1.8} & \tabnum{70.9}{1.7} \\
             GIF & \tabnum{80.8}{0.4} & \tabnum{80.8}{0.6} & \tabnum{80.6}{0.4} & \tabnum{69.8}{0.7} & \tabnum{69.8}{0.7} & \tabnum{69.3}{0.5} & \tabnum{75.4}{1.1} & \tabnum{75.3}{1.0} & \tabnum{75.2}{0.9} \\
             \ours{} (DE) & \tabnum{81.0}{0.3} & \tabnum{80.7}{0.3} & \tabnum{80.8}{0.3} & \tabnum{69.1}{0.5} & \tabnum{69.2}{0.7} & \tabnum{68.8}{0.8} & \tabnum{75.6}{1.0} & \tabnum{75.2}{1.0} & \tabnum{75.0}{1.2} \\
             \ours{} ($\oq$) & \tabnum{80.9}{0.4} & \tabnum{80.9}{0.6} & \tabnum{80.6}{0.7} & \tabnum{68.9}{0.6} & \tabnum{69.0}{0.5} & \tabnum{68.6}{0.8} & \tabnum{75.4}{0.9} & \tabnum{74.9}{1.0} & \tabnum{74.4}{1.1} \\
             \ours{} (VL) & \besttabnum{80.2}{0.7} & \besttabnum{79.1}{0.7} & \besttabnum{78.4}{0.8} & \besttabnum{68.2}{0.5} & \besttabnum{66.7}{0.6} & \besttabnum{65.1}{0.9} & \besttabnum{73.2}{1.4} & \besttabnum{71.2}{1.5} & \besttabnum{69.9}{1.1} \\
             \bottomrule
        \end{tabular}
    }
    \label{tab:attack}
\end{table*}

%% file: fig/figure/avg_influences.tex
\begin{wrapfigure}[20]{r}{0.5\textwidth}
    \centering
    \includegraphics[width=\linewidth]{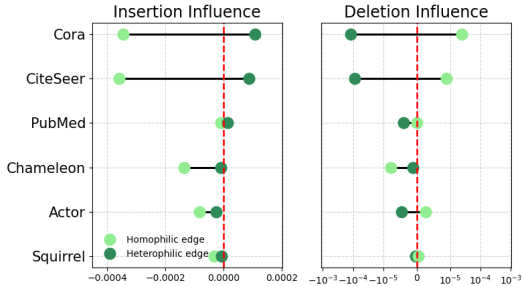}
    \caption{Mean influence of homophilic and heterophilic edges on validation loss for edge insertion (left) and edge deletion (right). Each dumbbell connects the average influence of homophilic (light green) and heterophilic (dark green) edges across six datasets. A negative value indicates that the edge edit decreases the validation loss, thus improving the performance.}
    \label{fig:homophily_analysis}
\end{wrapfigure}

%% file: fig/figure/histogram_rewiring.tex
\begin{figure}[t]
    \centering
    \begin{subfigure}[a]{\textwidth}
        \centering
        \includegraphics[width=1\linewidth]{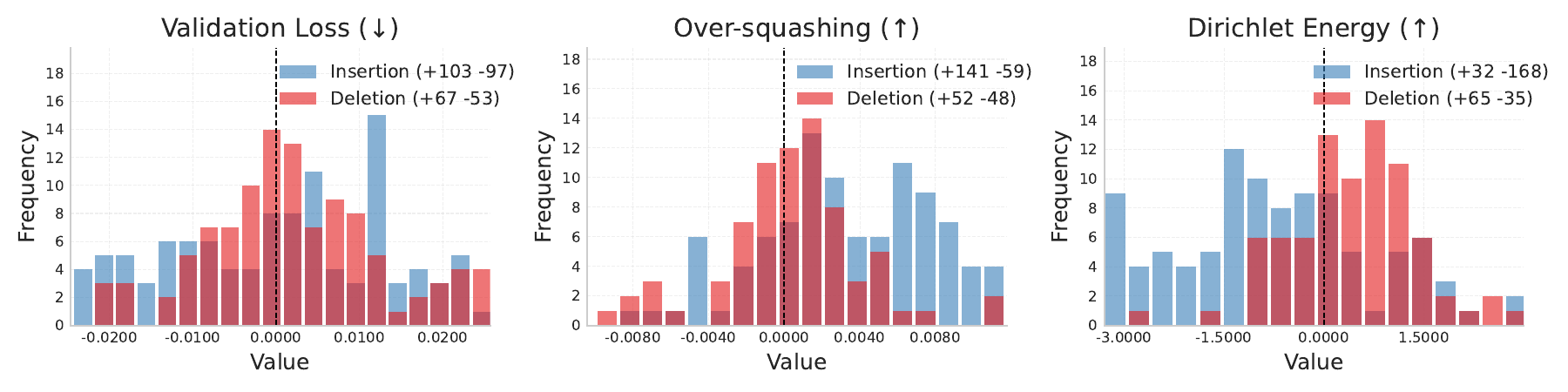}
        \caption{BORF~\citep{nguyen2023revisiting}}
    \end{subfigure}

    \begin{subfigure}[b]{\textwidth}
        \centering
        \includegraphics[width=1\linewidth]{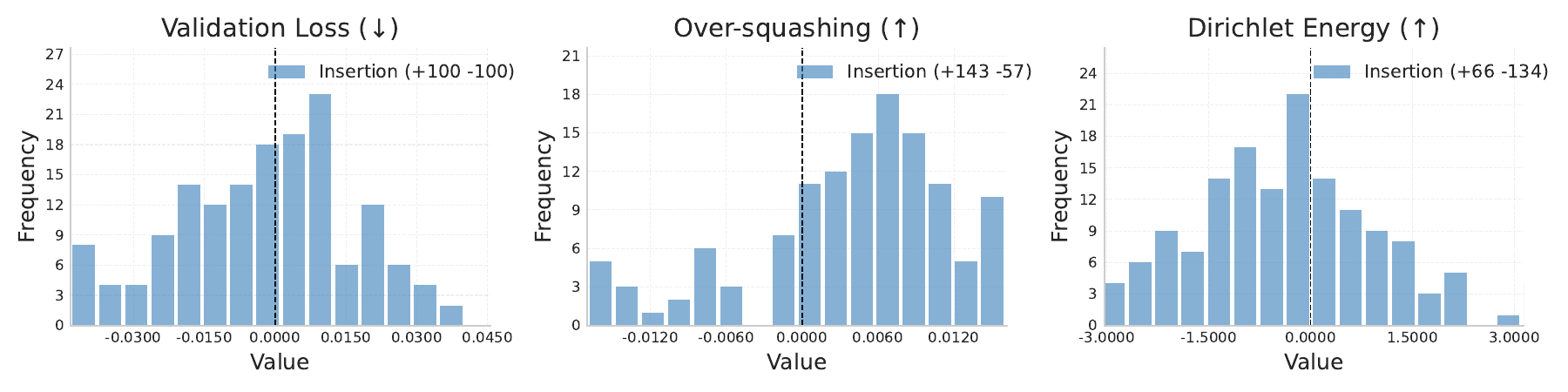}
        \caption{FoSR~\citep{karhadkar2023fosr}}
    \end{subfigure}

    \caption{Histograms of the estimated influence of edge insertions (blue) and deletions (red) selected by BORF~\citep{nguyen2023revisiting} and FoSR~\citep{karhadkar2023fosr}, evaluated on a four-layer GCN trained on the Texas dataset. Influences are measured across validation loss, over-squashing, and over-smoothing. Arrows ($\downarrow/\uparrow$) indicate the desired direction of each measurement (decrease/increase).}
    \label{fig:histogram_rewiring}
\end{figure}

%% file: fig/figure/naive_re_wiring.tex
\begin{figure}[t]
    \centering
    \begin{subfigure}[b]{\textwidth}
        \centering
        \includegraphics[width=0.32\linewidth]{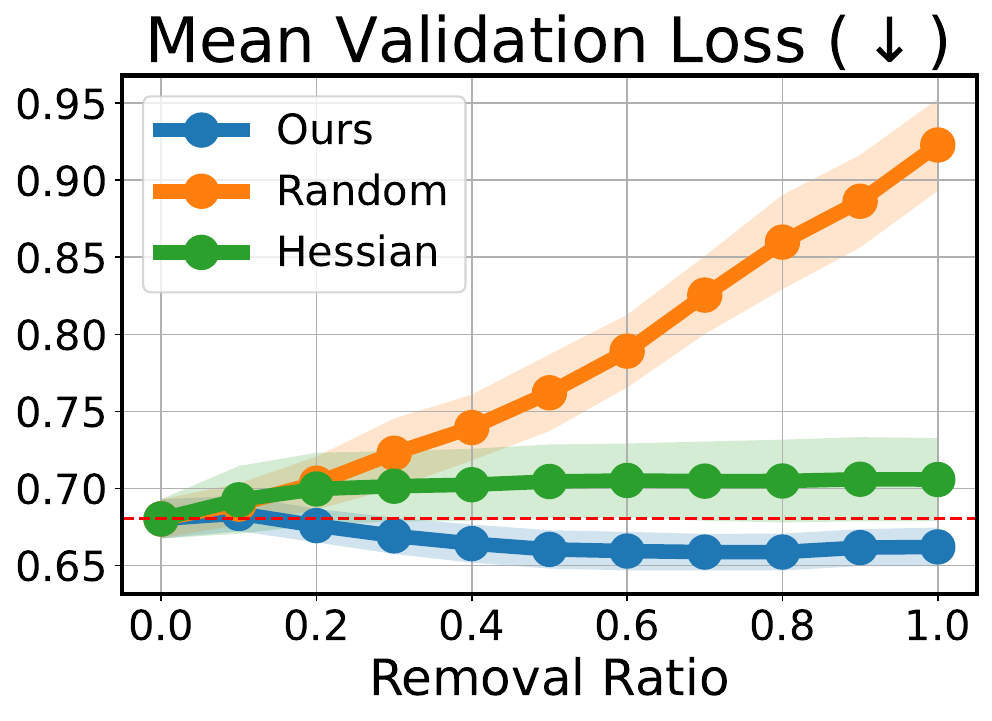}
        \includegraphics[width=0.32\linewidth]{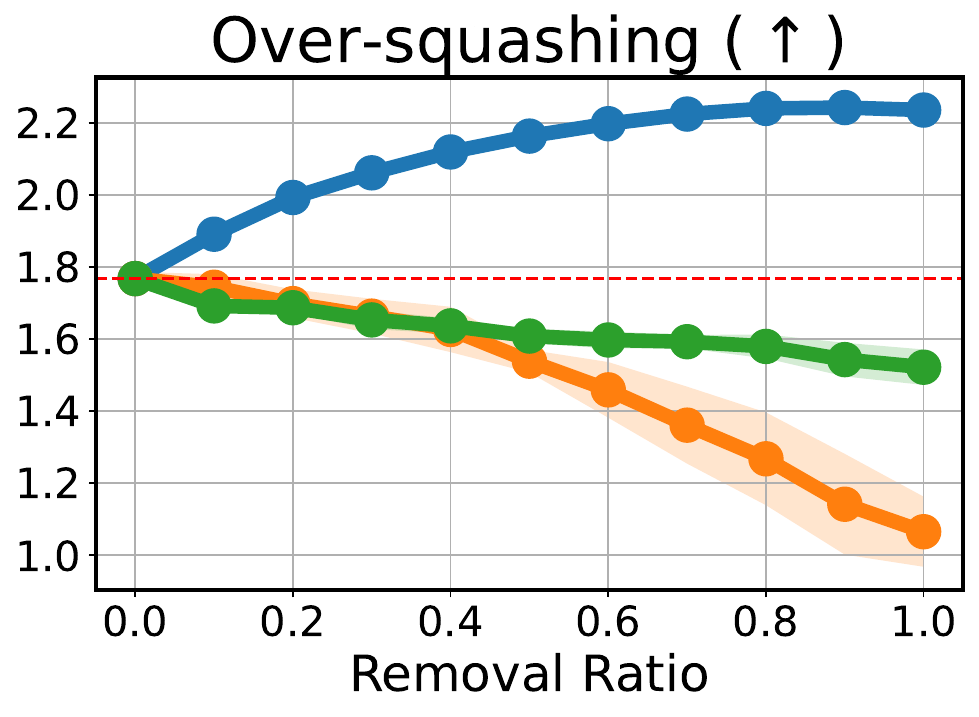}
        \includegraphics[width=0.32\linewidth]{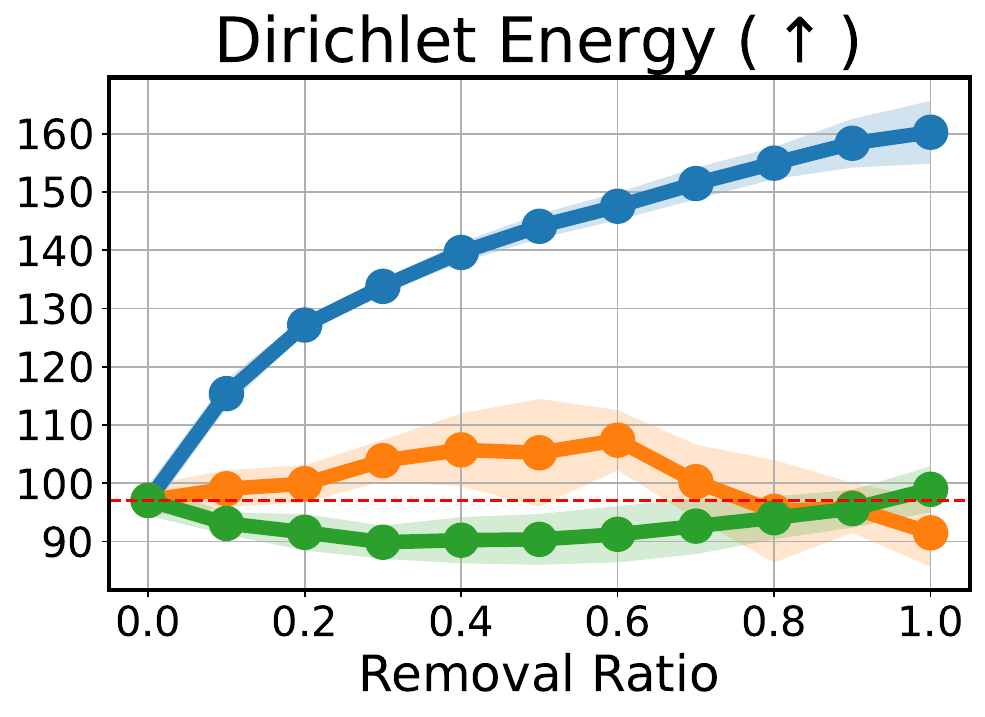}
        \caption{Edge removal}
    \end{subfigure}

    \begin{subfigure}[b]{\textwidth}
        \centering
        \includegraphics[width=0.32\linewidth]{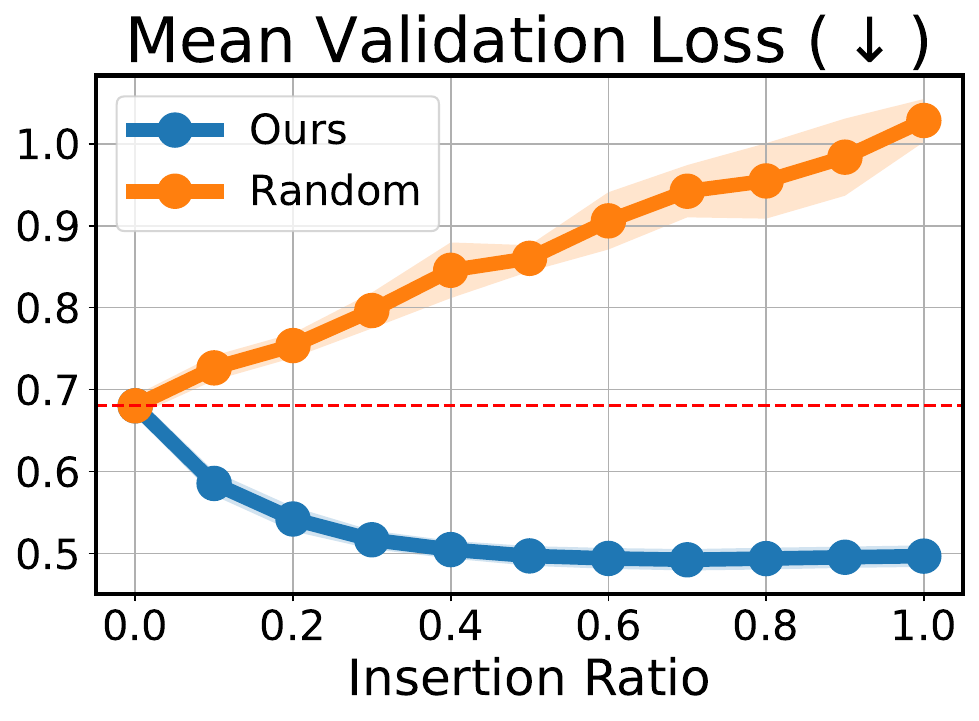}
        \includegraphics[width=0.32\linewidth]{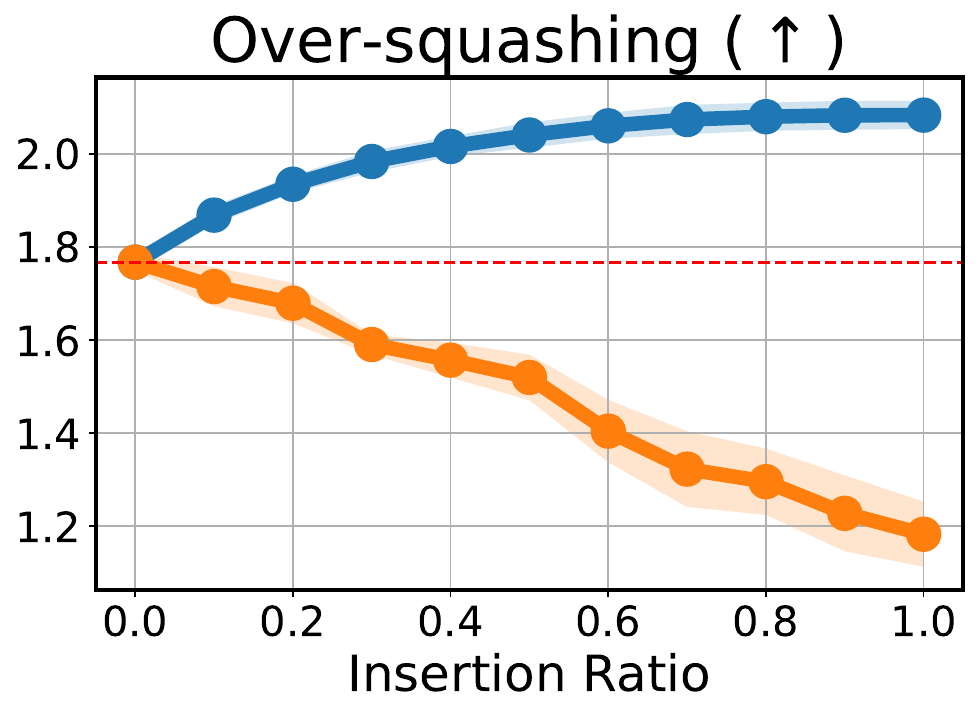}
        \includegraphics[width=0.32\linewidth]{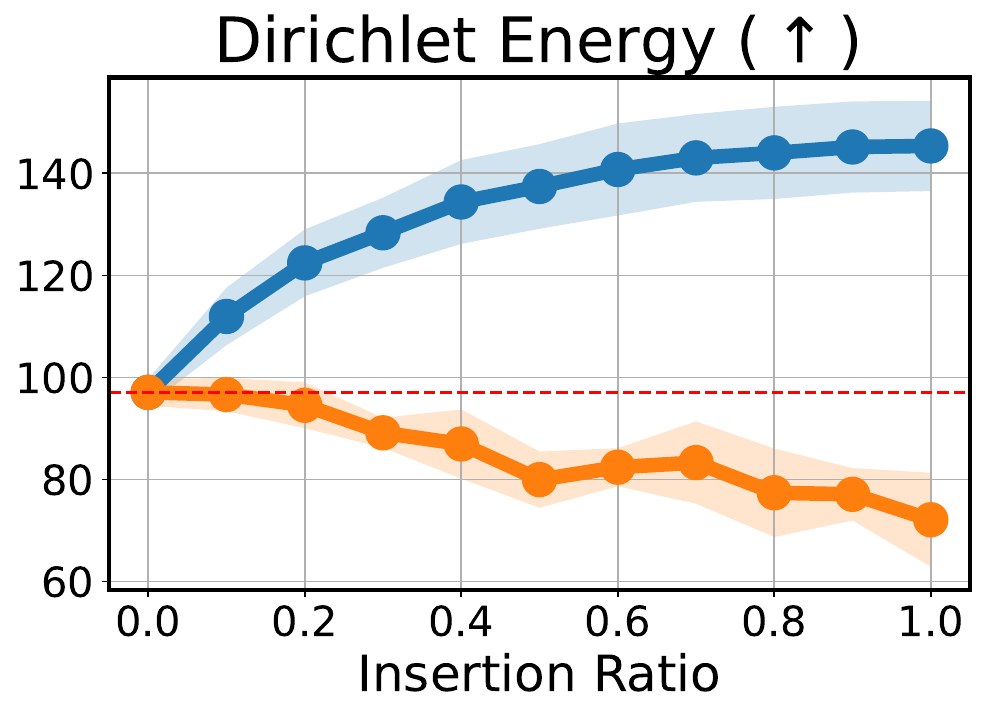}
        \caption{Edge insertion}
    \end{subfigure}

    \caption{The changes in mean validation loss, over-squashing, and Dirichlet energy after removing or inserting the top-$k$ positively influential edges. Arrows indicate whether each measurement is targeted to be decreased ($\downarrow$) or increased ($\uparrow$).}
    \label{fig:naive_re_wiring}
\end{figure}

%% file: fig/figure/kfold_result.tex
\begin{figure}[t]
    \centering
    \begin{subfigure}[b]{0.49\textwidth}
        \centering
        \includegraphics[width=0.49\linewidth]{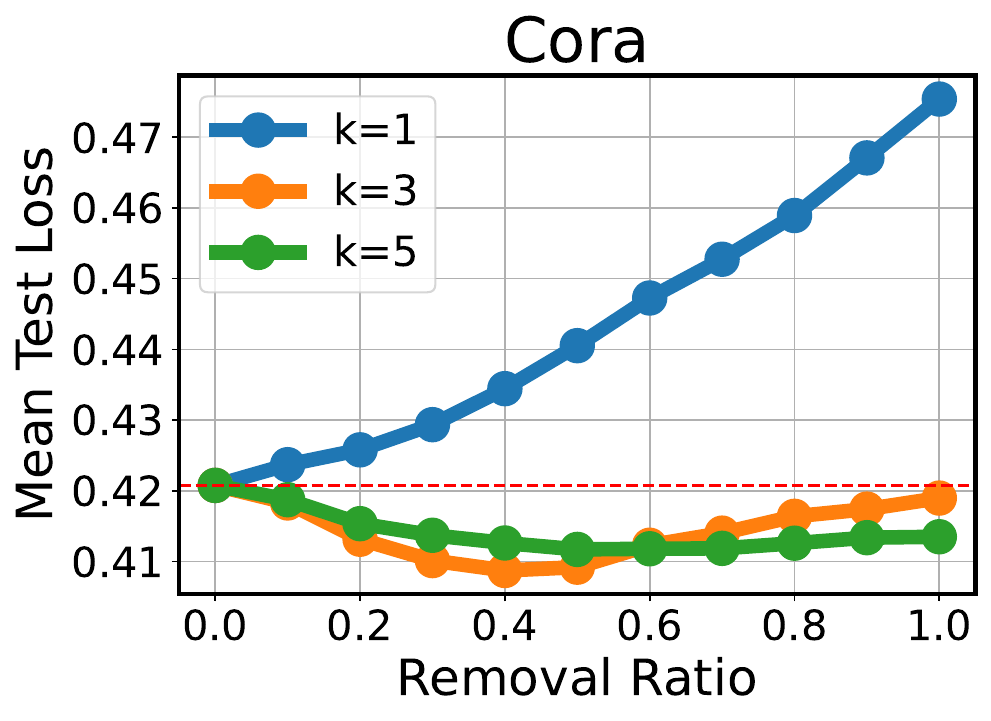}
        \includegraphics[width=0.49\linewidth]{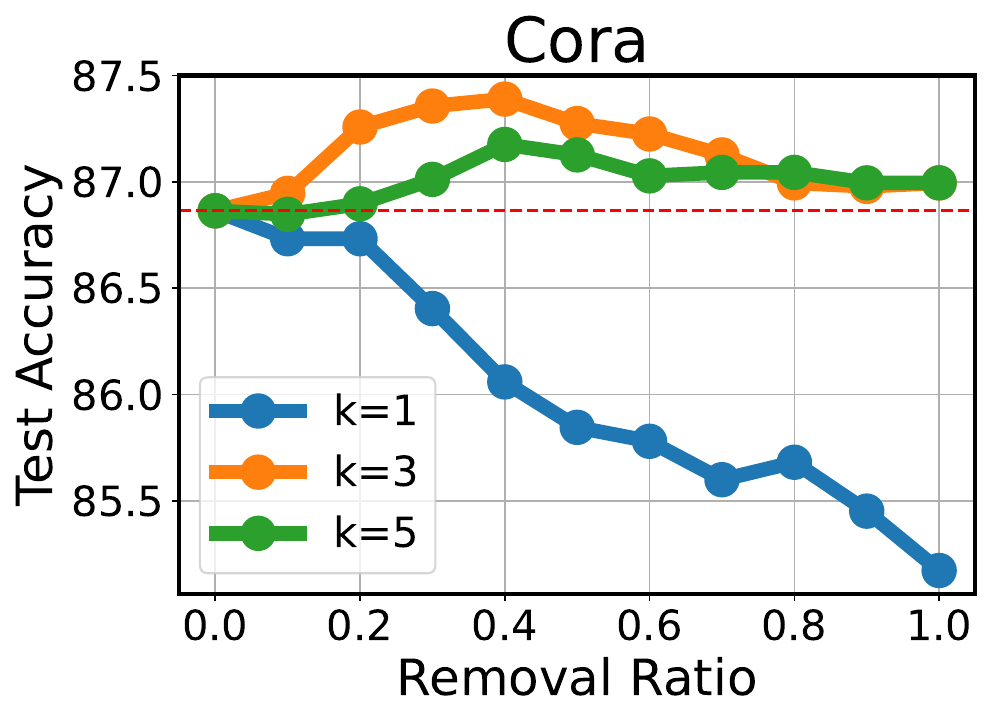}
        \caption{Edge Removal}
    \end{subfigure}
    \begin{subfigure}[b]{0.49\textwidth}
        \centering
        \includegraphics[width=0.49\linewidth]{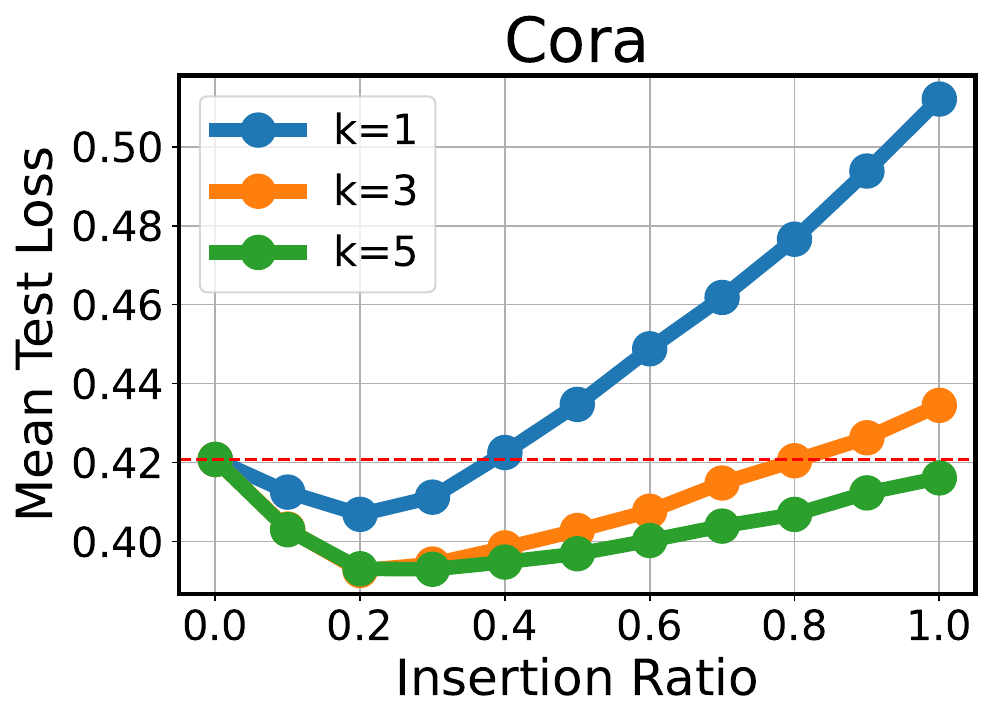}
        \includegraphics[width=0.49\linewidth]{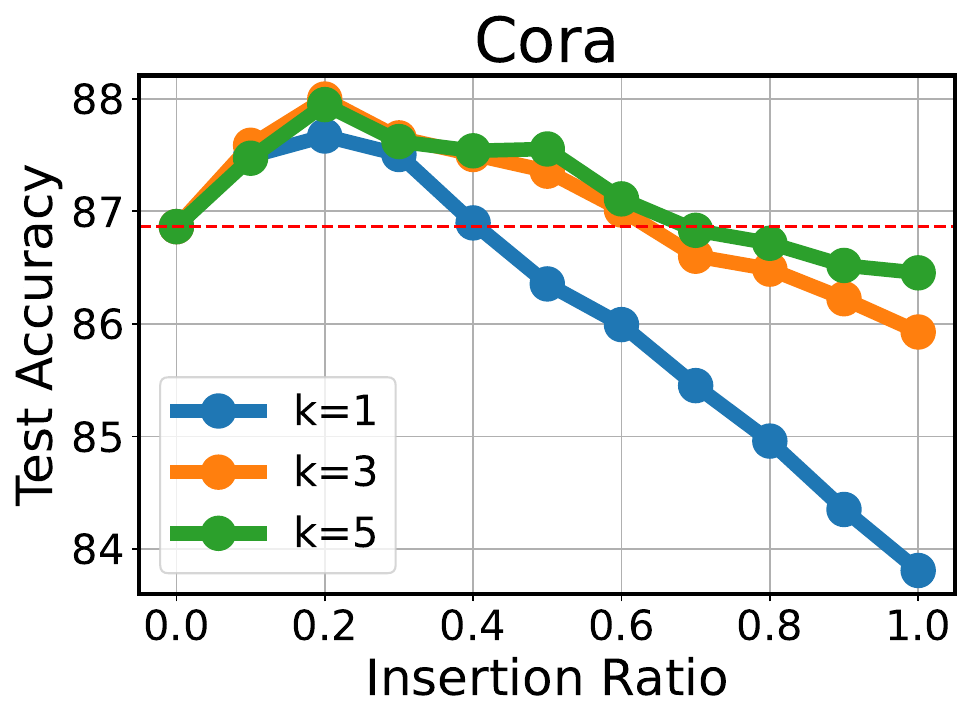}
        \caption{Edge Insertion}
    \end{subfigure}
    \caption{Mean test loss when removing or inserting edges identified by our method with varying numbers of validation subsets ($k$).}
    \label{fig:kfold_result}
\end{figure}

%% file: tex/2.related_work.tex
\section{Related work}


Originally introduced by \citet{cook1982residuals}, the influence function quantifies how removing a single training point would affect model parameters after retraining. \citet{koh2017understanding} adapted this concept to modern machine learning, providing efficient approximations of the influence on model predictions without the need for retraining. Influence functions have since been widely employed for model interpretability~\citep{han2020explaining, guo2020fastif}, data valuation~\citep{chhabra2024data, hu2024most}, adversarial analysis~\citep{cohen2020detecting, cohen2024membership}, and unlearning~\citep{zhang2024recommendation, yao2024large}.

A significant limitation of influence functions is the assumption of strict convexity in the loss function~\citep{basu2020influence}, which does not hold for many deep learning models. To overcome this, \citet{teso2021interactive} approximate the Hessian using the Fisher information matrix, while \citet{bae2022if} introduce a proximal Bregman response function objective that relaxes this convexity requirement. Recent studies have further extended influence functions to diffusion models~\citep{mlodozeniec2024influence} and large language models~\citep{grosse2023studying, choe2024your, zhang2025correcting}. In graph settings, \citet{chen2022characterizing} and \citet{wu2023gif} apply influence functions to transductive node classification, and \citet{song2023rge} analyze group-level influence. 

\if\else
\subsection{Edge-related challenges in GNNs}

Effective message passing strongly depends on the graph structure, leading to challenges such as over-smoothing and over-squashing. Over-smoothing~\citep{li2018deeper} occurs when node embeddings become indistinguishable due to excessive message passing, particularly in deeper GNN layers. A common approach to mitigate over-smoothing is random edge deletion~\citep{rong2019dropedge}. Extending this idea, \citet{spinelli2021fairdrop} propose dropping edges between nodes sharing sensitive attributes, while \citet{han2023structure} introduce methods based on node similarity and GNN layer structures. More recently, \citet{fang2023dropmessage} unified various random edge-dropping methods into a generalized framework.

Over-squashing~\citep{alon2020bottleneck} occurs when information from distant nodes becomes excessively compressed due to limited communication paths, impeding effective message propagation. To alleviate over-squashing, curvature-based approaches guide rewiring by identifying edges using curvature metrics: Ollivier-Ricci curvature~\citep{nguyen2023revisiting}, Jost and Liu curvature~\citep{10.1145/3583780.3614997}, and augmented Forman-Ricci curvature~\citep{pmlr-v231-fesser24a}. In contrast, spectral-based methods, such as that proposed by \citet{karhadkar2023fosr}, strategically insert edges to increase the spectral gap, thereby mitigating over-squashing.
\fi

%% file: tex/7.conclusion.tex
\section{Conclusion}
\label{sec:conclusion}
We introduce an enhanced graph influence function that estimates the impact of edge perturbations on model predictions in Graph Neural Networks. Unlike existing graph influence functions, our approach explicitly incorporates message propagation effects and relaxes the convexity assumption, enabling it to be applied to commonly used non-convex GNNs. Additionally, we extend the framework to handle both edge deletions and insertions, broadening its applicability to real-world graph rewiring tasks. 
Experimental results on various real-world datasets demonstrate that our method provides significantly more accurate estimates of influence than previous methods and enables effective improvements in key target measurements such as validation loss, over-squashing, and over-smoothing. 
Despite these advances, our method has limitations. Although it achieves high accuracy under multiple edge edits, performance gradually declines as the number of simultaneous edits grows, and scalability to deep GNNs remains challenging. Future work could address these issues to enhance the applicability of the method to complex graph learning tasks.


%% file: tex/appendix.tex
\appendix


\input{tex/appendix/1.influence_function_derivation}
\input{tex/appendix/2.graph_influence_function_derivation}
\input{tex/appendix/2.5.IF_multiple_edges}
\input{tex/appendix/3.lissa}
\input{tex/appendix/4.additional_results}
\input{tex/appendix/6.experimental_settings}

%% file: tex/appendix/1.influence_function_derivation.tex
\section{Derivation of the influence function}
\label{apx:if_derivation}

In this section, we derive the influence function for linear models~\citep{koh2017understanding}, as shown in \Cref{eqn:influence_function_gradient}, and for neural networks~\citep{bae2022if}, as shown in \Cref{eqn:influence_function_pbrf}. In particular, we reformulate the derivation by \citet{bae2022if} to fit our setting.

\subsection{Influence function for linear models}

We begin by considering the re-weighted objective introduced in \Cref{eqn:loo}:
\begin{equation}
\mathcal{J}(\theta, \epsilon) = \frac{1}{N} \sum_{(x,y)\in\mathcal{D}_{\text{train}}} \mathcal{L}(x, y, \theta) + \epsilon\, \mathcal{L}(x', y', \theta).
\end{equation}

Assuming that the loss function \(\mathcal{L}\) is twice continuously differentiable and that \(\mathcal{J}(\theta, \epsilon)\) is strictly convex in \(\theta\), the optimal parameter \(\theta_{x',y',\epsilon}^*\) minimizing the objective satisfies the first-order optimality condition:
\begin{equation}
\label{eqn:stationary}
\quad \nabla_{\theta} \mathcal{J}(\theta_{x',y',\epsilon}^*, \epsilon) = 0.
\end{equation}

To justify that the response function \(\theta_{x', y', \epsilon}^*\) is differentiable with respect to \(\epsilon\), we apply the Implicit Function Theorem to the optimality condition. The following conditions must be satisfied for the theorem to apply:

\begin{itemize}
    \item The function \(\nabla_{\theta} \mathcal{J}(\theta, \epsilon)\) is continuously differentiable in both \(\theta\) and \(\epsilon\). This holds because \(\mathcal{J}(\theta, \epsilon)\) is constructed as a linear combination of smooth loss functions, and its dependence on \(\epsilon\) is linear.
    
    \item The optimality condition \(\nabla_{\theta} \mathcal{J}(\theta_{x',y',\epsilon}^*, \epsilon) = 0\) holds by definition, since \(\theta_{x',y',\epsilon}^*\) minimizes the objective \(\mathcal{J}(\theta, \epsilon)\).
    
    \item The Hessian \(\nabla^2_{\theta} \mathcal{J}(\theta, \epsilon)\), taken with respect to \(\theta\), is non-singular in a neighborhood of \(\epsilon = 0\), as \(\mathcal{J}(\theta, \epsilon)\) is assumed to be strictly convex in \(\theta\).
\end{itemize}

Under these conditions, the Implicit Function Theorem ensures that \(\theta_{x',y',\epsilon}^*\) is continuously differentiable with respect to \(\epsilon\), and we differentiate \Cref{eqn:stationary} using the chain rule:
\begin{equation}
\frac{d}{d\epsilon} \left( \nabla_{\theta} \mathcal{J}(\theta_{x',y',\epsilon}^*, \epsilon) \right)
= \nabla^2_{\theta} \mathcal{J}(\theta_{x',y',\epsilon}^*, \epsilon) \cdot \frac{d\theta_{x',y',\epsilon}^*}{d\epsilon}
+ \nabla_{\theta} \mathcal{L}(x', y', \theta_{x',y',\epsilon}^*) = 0.
\end{equation}

Solving for the derivative yields:
\begin{equation}
\frac{d\theta_{x',y',\epsilon}^*}{d\epsilon}
= -\left( \nabla^2_{\theta} \mathcal{J}(\theta_{x',y',\epsilon}^*, \epsilon) \right)^{-1}
\nabla_{\theta} \mathcal{L}(x', y', \theta_{x',y',\epsilon}^*).
\end{equation}

Evaluating at \(\epsilon = 0\), we obtain the influence function:
\begin{equation}
\left. \frac{d\theta_{x',y',\epsilon}^*}{d\epsilon} \right|_{\epsilon=0}
= -\left( \nabla^2_{\theta} \mathcal{J}(\theta^*, 0) \right)^{-1}
\nabla_{\theta} \mathcal{L}(x', y', \theta^*),
\end{equation}
where \(\theta^* := \theta_{x',y',\epsilon=0}^*\) is the minimizer of the original objective without perturbation.

\newpage
\subsection{Influence function for neural networks}
\label{apx:subsec:pbrf}

\citet{bae2022if} demonstrated that the influence function using the generalized Gauss-Newton Hessian corresponds to that of the linearized form of the proximal Bregman response function objective\footnote{For notational simplicity, we omit the label \(y\) in expressions involving the loss function or the Bregman divergence, as it is clear from context.}:
\begin{align}
\label{eqn:linearized_pbrf}
\lpbrfr 
\coloneqq \argmin_{\theta}
\frac{1}{N} \sum_{(x, y) \in \mathcal{D}_{\text{train}}}
D_{\mathcal{L}_{\text{quad}}} \left(\htl, \hts\right)
+ \frac{\lambda}{2} \left\| \theta - \theta_s \right\|^2
+ \epsilon\, \nabla_{\theta} \mathcal{L}(x', \theta_s)^\top \theta,
\end{align}
where \(\hth = g_\theta(x)\), and \(\mathcal{L}_{\text{quad}}\) and \(\htl\) are the quadratic and linear approximations of the loss and model output, respectively:
\begin{align}
\mathcal{L}_{\text{quad}}(\hth)
&= \mathcal{L}(\hts)
+ \nabla_{\hth} \mathcal{L}(\hts)^\top (\hth - \hts)
+ \frac{1}{2}(\hth - \hts)^\top \nabla_{\hth}^2 \mathcal{L}(\hts)(\hth - \hts), \notag \\
\htl
&= \hts + \mathbf{J}_{\hth \theta_s}(\theta - \theta_s),
\end{align}
where \(\mathbf{J}_{\hth \theta_s} := \frac{\partial \hth}{\partial \theta}\big|_{\theta=\theta_s}\) is the Jacobian of the model output with respect to the parameters.

We now expand the Bregman divergence term. First, using \(\mathcal{L}_{\text{quad}}(\hts)=\set{L}(\hts)\):
\begin{align}
D_{\mathcal{L}_{\text{quad}}}(\htl, \hts)
&= \mathcal{L}_{\text{quad}}(\htl)
- \mathcal{L}_{\text{quad}}(\hts)
- \nabla_{\hth} \mathcal{L}_{\text{quad}}(\hts)^\top (\htl - \hts) \notag \\
&= \mathcal{L}(\hts)
+ \nabla_{\hth} \mathcal{L}(\hts)^\top (\hth - \hts)
+ \frac{1}{2}(\hth - \hts)^\top \nabla_{\hth}^2 \mathcal{L}(\hts)(\hth - \hts) \notag \\
&\quad - \mathcal{L}(\hts)
- \nabla_{\hth} \mathcal{L}(\hts)^\top (\hth - \hts).
\end{align}

Next, using \(\htl-\hts=\mathbf{J}_{\hth \theta_s}(\theta - \theta_s)\):
\begin{align}
D_{\mathcal{L}_{\text{quad}}}(\htl, \hts)
&= \nabla_{\hth} \mathcal{L}(\hts)^\top \mathbf{J}_{\hth \theta_s}(\theta - \theta_s)
+ \frac{1}{2}(\theta - \theta_s)^\top \mathbf{J}_{\hth \theta_s}^\top \nabla_{\hth}^2 \mathcal{L}(\hts) \mathbf{J}_{\hth \theta_s}(\theta - \theta_s) \notag \\
&\quad - \nabla_{\hth} \mathcal{L}(\hts)^\top \mathbf{J}_{\hth \theta_s}(\theta - \theta_s) \notag \\
&= \frac{1}{2}(\theta - \theta_s)^\top \mathbf{J}_{\hth \theta_s}^\top \nabla_{\hth}^2 \mathcal{L}(\hts) \mathbf{J}_{\hth \theta_s}(\theta - \theta_s).
\end{align}

Since \(\lpbrfr\) is the optimal solution, the gradient of the objective with respect to \(\theta\) is zero at this point:
\begin{align}
\label{eqn:linearized_pbrf_grad_zero}
0 
&= -\frac{1}{N}\sum_{(x,y)\in\set{D}_{\text{train}}}\mathbf{J}_{\hth \theta_s}^\top \nabla_{\hth}^2 \mathcal{L}(\hts) \mathbf{J}_{\hth \theta_s}(\lpbrfr - \theta_s)
+ \lambda(\lpbrfr - \theta_s)
+ \epsilon\, \nabla_{\theta} \mathcal{L}(x', \theta_s).
\end{align}

Solving for \(\lpbrfr\), we obtain:
\begin{equation}
\lpbrfr = \theta_s + \left(
\mathbf{J}_{h^{\theta} \theta_s}^\top \mat{H}_{h_{s}} \mathbf{J}_{h^{\theta} \theta_s}
+ \lambda \mat{I}
\right)^{-1} \nabla_{\theta} \mathcal{L}(x', \theta_s) \epsilon,
\end{equation}
where \(\mathbf{J}_{h^{\theta} \theta_s}^\top \mat{H}_{h_{s}} \mathbf{J}_{h^{\theta} \theta_s}\coloneq\frac{1}{N}\sum_{(x,y)\in\set{D}_{\text{train}}}\mathbf{J}_{\hth \theta_s}^\top \nabla_{\hth}^2 \mathcal{L}(\hts) \mathbf{J}_{\hth \theta_s}\) is the generalized Gauss-Newton Hessian.

%% file: tex/appendix/2.graph_influence_function_derivation.tex
\newpage
\section{Gradient derivation for edge-edit PBRF}
\label{apx:pbrf_derivation}

Similar to the derivation of the influence function for the PBRF objective in \Cref{apx:subsec:pbrf}, we demonstrate that the influence function in \Cref{eqn:retraining_effect} corresponds to that of the linearized form of the edge-edit PBRF objective:
\begin{align}
\lgpbrfr := \argmin_{\theta}
&\frac{1}{N} \sum_{v \in \set{V}_{\text{train}}}
D_{\set{L}_{\text{quad}}} \left(\ghtl, \nrep{\theta_s}\right)
+ \frac{\lambda}{2} \left\| \theta - \theta_s \right\|^2 \notag \\
&+ \sum_{v \in \set{V}_{\text{train}}} \epsilon
\left(
\nabla_{\theta}\set{L}(\nrep{\theta_s}) - \nabla_{\theta}\set{L}(\nrepgme{\theta_s})
\right)^\top \theta,
\end{align}
where \(\set{L}_{\text{quad}}\) and \(\ghtl\) denote the quadratic and linear approximations of the loss and the model output, respectively:
\begin{align}
\set{L}_{\text{quad}}(\ghth)
&= \set{L}(\ghts)
+ \nabla_{\ghth} \set{L}(\ghts)^\top (\ghth - \ghts) \notag \\
&\quad + \frac{1}{2} (\ghth - \ghts)^\top \nabla_{\ghth}^2 \set{L}(\ghts)(\ghth - \ghts), \notag \\
\ghtl
&= \ghts + \mathbf{J}_{\ghth \theta_s}(\theta - \theta_s),
\end{align}
where \(\mathbf{J}_{\ghth \theta_s} := \frac{\partial \ghth}{\partial \theta} \big|_{\theta = \theta_s}\) is the Jacobian of the model output with respect to the parameters.

We first expand the Bregman divergence term:
\begin{align}
D_{\set{L}_{\text{quad}}}(\ghtl, \ghts)
&= \set{L}_{\text{quad}}(\ghtl)
- \set{L}_{\text{quad}}(\ghts)
- \nabla_{\ghth} \set{L}_{\text{quad}}(\ghts)^\top (\ghtl - \ghts) \notag \\
&= \set{L}(\ghts)
+ \nabla_{\ghth} \set{L}(\ghts)^\top (\ghth - \ghts) \notag \\
&+ \frac{1}{2}(\ghth - \ghts)^\top \nabla_{\ghth}^2 \set{L}(\ghts)(\ghth - \ghts) \notag \\
&\quad - \set{L}(\ghts)
- \nabla_{\ghth} \set{L}(\ghts)^\top (\ghth - \ghts).
\end{align}

Using \(\ghtl - \ghts = \mathbf{J}_{\ghth \theta_s}(\theta - \theta_s)\), we substitute into the expression:
\begin{align}
D_{\set{L}_{\text{quad}}}(\ghtl, \ghts)
&= \nabla_{\ghth} \set{L}(\ghts)^\top \mathbf{J}_{\ghth \theta_s}(\theta - \theta_s) \notag \\
&\quad + \frac{1}{2} (\theta - \theta_s)^\top \mathbf{J}_{\ghth \theta_s}^\top \nabla_{\ghth}^2 \set{L}(\ghts) \mathbf{J}_{\ghth \theta_s}(\theta - \theta_s) \notag \\
&\quad - \nabla_{\ghth} \set{L}(\ghts)^\top \mathbf{J}_{\ghth \theta_s}(\theta - \theta_s) \notag \\
&= \frac{1}{2} (\theta - \theta_s)^\top \mathbf{J}_{\ghth \theta_s}^\top \nabla_{\ghth}^2 \set{L}(\ghts) \mathbf{J}_{\ghth \theta_s}(\theta - \theta_s).
\end{align}

Since \(\lgpbrfr\) is the optimal parameter minimizing the objective, the derivative of the objective with respect to \(\theta\) at \(\theta = \lgpbrfr\) is zero:
\begin{align}
0 
= &-\frac{1}{N} \sum_{v \in \set{V}_{\text{train}}} \mathbf{J}_{\ghth \theta_s}^\top \nabla_{\ghth}^2 \set{L}(\ghts) \mathbf{J}_{\ghth \theta_s}(\lgpbrfr - \theta_s)
+ \lambda(\lgpbrfr - \theta_s) \notag \\
&\quad + \sum_{v \in \set{V}_{\text{train}}} \epsilon \left(
\nabla_{\theta}\set{L}(\nrep{\theta_s}) - \nabla_{\theta}\set{L}(\nrepgme{\theta_s})
\right).
\end{align}

Rearranging the terms, we obtain:
\begin{align}
\lgpbrfr
= \theta_s + \epsilon\, \mat{G}^{-1} \sum_{v \in \set{V}_{\text{train}}} \left(
\nabla_{\theta}\set{L}(\nrep{\theta_s}) - \nabla_{\theta}\set{L}(\nrepgme{\theta_s})
\right),
\end{align}
where \(\mat{G} := \frac{1}{N} \sum_{v \in \set{V}_{\text{train}}} \mathbf{J}_{\ghth \theta_s}^\top \nabla_{\ghth}^2 \set{L}(\ghts) \mathbf{J}_{\ghth \theta_s} + \lambda \mat{I}\) is the generalized Gauss-Newton matrix.

Taking the derivative with respect to \(\epsilon\), we obtain:
\begin{align}
\frac{\partial \lgpbrfr}{\partial \epsilon}
= \lim_{\epsilon \to 0} \frac{\lgpbrfr - \theta_s}{\epsilon}
= \mat{G}^{-1} \sum_{v \in \set{V}_{\text{train}}} \left(
\nabla_{\theta}\set{L}(\nrep{\theta_s}) - \nabla_{\theta}\set{L}(\nrepgme{\theta_s})
\right),
\end{align}
which confirms that the linearized edge-edit PBRF objective yields the same influence function as derived in \Cref{eqn:retraining_effect}.

%% file: tex/appendix/2.5.IF_multiple_edges.tex
\section{Influence function for multiple edge edits}
\label{apx:if_multiple}

\paragraph{Problem setup for multiple edge edits}
In this section, we extend our influence function to handle multiple edge edits. Let \(\mathcal{S}\) denote the set of edges to be edited. We generalize the definition of \(A^{\epsilon}\) in \Cref{sec:method} to the multi-edge case by defining  

\[
\ruv = A_{uv} + \bigl(2\mathbb{I}[\{u,v\}\in\mathcal{S}] - 1\bigr)N\epsilon .
\]  

The edge-reweighted graph is then given by \(\mathcal{G}^{\epsilon} = \{\mathcal{V}, \mathcal{E}, A^{\epsilon}\}\), where edges in \(\mathcal{S}\) are reweighted. Setting \(\epsilon = -1/N\) implies that all edges in \(\mathcal{S}\) are deleted if they exist, and inserted otherwise. 

\paragraph{Parameter shift}
Following \Cref{sec:method}, the edge-edit PBRF is defined as

\begin{equation}
\gpbrf{} := \argmin_{\theta}\frac{1}{N}\sum_{v\in\mathcal{V}_{\text{train}}}D_{\mathcal{L}}\left(\nrep{\theta},\nrep{\theta_s}\right)
+\frac{\lambda}{2}\left\|\theta-\theta_s\right\|^2
+\sum_{v\in\mathcal{V}_{\text{train}}}\epsilon\left(\mathcal{L}\left(\nrep{\theta}\right)-\mathcal{L}\left(\nrepgme{\theta}\right)\right).
\end{equation}

Following the same derivation steps as in \Cref{apx:pbrf_derivation}, the contribution of the parameter shift to the evaluation function is given by
\begin{equation}
\nabla_{\theta}f(\theta^*_0,\set{G})^{\top}\left.\frac{\partial \gpbrf{}}{\partial\epsilon}\right|_{\epsilon=0}
=-\nabla_{\theta}f(\theta_s,\set{G})^{\top}\mathbf{G}^{-1}\sum_{v\in\mathcal{V}_{\text{train}}}\left(\nabla_{\theta}\mathcal{L}\left(\nrep{\theta_s}\right)-\nabla_{\theta}\mathcal{L}\left(\nrepgme{\theta_s}\right)\right),
\end{equation}
where \(\mathbf{G} = \mathbf{J}_{h\theta_s}^{\top} \mathbf{H}_{h_s} \mathbf{J}_{h\theta_s} + \lambda \mathbf{I}\). The only distinction from the single-edge case lies in the redefinition of $\mathcal{G}^{-1/N}$, which now denotes the graph with multiple edge edits.

\paragraph{Message propagation}
Unlike the single-edge case in \Cref{sec:method}, the perturbed adjacency matrix $A^{\epsilon}$ is influenced simultaneously by all edges in the set $\mathcal{S}$. Consequently, the message propagation effect can no longer be attributed to a single edge but must be decomposed into contributions from all edited edges. By applying the chain rule, the overall effect is expressed as a summation over gradients with respect to each reweighted edge:

\begin{align}
\left.\frac{\partial f(\theta, \editedgraph)}{\partial \radj{}} \frac{\partial \radj{}}{\partial \epsilon} \right|_{\theta=\theta^*_0,\;\epsilon=0}
&= \sum_{\{u,v\}\in\set{S}}\left.\frac{\partial f(\theta,\editedgraph)}{\partial \ruv{}}\frac{\partial \ruv{}}{\partial \epsilon}\right|_{\theta=\theta_s,\,\epsilon=0}
+ \left.\frac{\partial f(\theta,\editedgraph)}{\partial \rvu{}}\frac{\partial \rvu{}}{\partial \epsilon}\right|_{\theta=\theta_s,\,\epsilon=0} \notag\\
&= \sum_{\{u,v\}\in\set{S}}(2\mathbb{I}[\{u,v\}\in\mathcal{E}]-1)\,N\left(\frac{\partial f(\theta_s,\set{G})}{\partial A_{uv}}+\frac{\partial f(\theta_s,\set{G})}{\partial A_{vu}}\right).
\end{align}

\paragraph{Unified influence function under multiple edge edits}

By substituting the parameter shift and message propagation terms into \Cref{eqn:ours_grad} and linearizing around \(\epsilon = 0\), the resulting influence under multiple edge edits can be approximated as
\begin{align}
f\left(\theta_{-\frac{1}{N}}^*,\set{G}^{-\frac{1}{N}}\right) - f(\theta_s,\set{G}) 
&\approx \frac{1}{N}\nabla_{\theta}f(\theta_s,\set{G})^\top\mathbf{G}^{-1}\sum_{v\in\mathcal{V}_{\text{train}}}\left(\nabla_{\theta}\mathcal{L}\left(\nrep{\theta_s}\right)-\nabla_{\theta}\mathcal{L}\left(\nrepgme{\theta_s}\right)\right) \notag \\[6pt]
&\quad -\sum_{\{u,v\}\in\set{S}}(2\mathbb{I}[\{u,v\}\in\mathcal{E}]-1)\left(\frac{\partial f(\theta_s,\set{G})}{\partial A_{uv}}+\frac{\partial f(\theta_s,\set{G})}{\partial A_{vu}}\right).
\end{align}

%% file: tex/appendix/3.lissa.tex
\newpage
\section{Description of LiSSA}
\label{apx:lissa}

To approximate the inverse of the generalized Gauss–Newton Hessian-vector product \(\mathbf{G}^{-1}v\), 
we employ the LiSSA algorithm~\citep{agarwal2016second}. LiSSA estimates \(\mathbf{G}^{-1}v\) by iteratively accumulating powers of the residual matrix \((\mathbf{I} - \mathbf{G})\) applied to the vector \(v\). When the spectral radius of \((\mathbf{I} - \mathbf{G})\) is less than 1, the inverse can be expressed using the Neumann series:
\begin{equation}
\mathbf{G}^{-1}v = \sum_{k=0}^{\infty} (\mathbf{I} - \mathbf{G})^k v.
\end{equation}

Letting \(r^{(K)} = \sum_{k=0}^{K} (\mathbf{I} - \mathbf{G})^k v\), the iteration is defined recursively as:
\begin{equation}
\label{eqn:lissa_iteration}
r^{(0)} = v, \quad r^{(k+1)} = v + (\mathbf{I} - \mathbf{G}) r^{(k)}.
\end{equation}

In practice, we perform the update in Equation~\eqref{eqn:lissa_iteration} until convergence. The iteration is terminated early if the update difference \(\|r^{(k+1)} - r^{(k)}\|\) falls below a predefined threshold, or when the number of iterations reaches 10,000.

Since the spectral radius of \((\mathbf{I} - \mathbf{G})\) is not necessarily less than 1, we rescale the matrix to ensure convergence. Specifically, we define a scaled matrix \(\mathbf{G}_s = \frac{1}{s} \mathbf{G}\), where \(s > \lambda_{\max}(\mathbf{G})\), so that the spectral radius of \((\mathbf{I} - \mathbf{G}_s)\) is less than 1. The inverse is then computed via
\[
\mathbf{G}^{-1}v = \frac{1}{s} \mathbf{G}_s^{-1}v,
\]
and LiSSA is applied to approximate \(\mathbf{G}_s^{-1}v\).

To avoid explicitly storing the generalized Gauss–Newton matrix \(\mathbf{G}\), the matrix-vector product \(\mathbf{G} r^{(k)}\) in Equation~\eqref{eqn:lissa_iteration} is computed approximately using a Jacobian-based heuristic. Specifically, we compute the Jacobian-vector product \(\vec{x} = \mathbf{J}_{h\theta_s} r^{(k)}\), apply the Hessian to obtain \(\vec{x} \leftarrow \mathbf{H}_{h_s} \vec{x}\), and finally compute the transposed Jacobian-vector product \(\mathbf{G} r^{(k)} = \mathbf{J}_{h\theta_s}^\top \vec{x}\).

\if\else
\newpage
\section{Description of LiSSA}
\label{apx:lissa}

To approximate the inverse of the generalized Gauss–Newton Hessian-vector product \(\mathbf{G}^{-1}v\), 
the LiSSA algorithm~\citep{agarwal2016second} is employed. LiSSA estimates \(\mathbf{G}^{-1}v\) by iteratively accumulating powers of the residual matrix \((\mathbf{I} - \mathbf{G})\) applied to the vector \(v\). When the spectral norm of \(\mathbf{G}\) is less than 1, the inverse can be expressed as a Neumann series:
\begin{equation}
\mathbf{G}^{-1}v = \sum_{k=0}^{\infty} (\mathbf{I} - \mathbf{G})^k v.
\end{equation}

Letting \(\vec{r}^{(K)} = \sum_{k=0}^{K} (\mathbf{I} - \mathbf{G})^k v\), the iteration is defined recursively as:
\begin{equation}
\label{eqn:lissa_iteration}
r^{(0)} = v, \quad r^{(k+1)} = v + (\mathbf{I} - \mathbf{G}) r^{(k)}.
\end{equation}

The approximation of \(\mathbf{G}^{-1}v\) is obtained by repeating the update in Equation~\eqref{eqn:lissa_iteration} until convergence. In the experimental setup, the maximum number of iterations is set to 10{,}000, and the iteration is terminated early if the update change \(\|r^{(k+1)} - r^{(k)}\|\) falls below a predefined threshold.

Since the matrix \(\mathbf{G}\) does not necessarily satisfy the condition \(\|\mathbf{G}\| < 1\), it is rescaled to ensure convergence. A scaled matrix \(\mathbf{G}_s = \frac{1}{s} \mathbf{G}\) is introduced, where \(s > \lambda_{\max}(\mathbf{G})\). The inverse is then computed as:
\[
\mathbf{G}^{-1}v = \frac{1}{s} \mathbf{G}_s^{-1}v,
\]
and the LiSSA iteration is applied to \(\mathbf{G}_s^{-1}v\), which satisfies the convergence condition. To avoid explicitly storing the generalized Gauss–Newton matrix \(\mathbf{G}\), the matrix-vector product \(\mathbf{G}r^{(k)}\) in Equation~\eqref{eqn:lissa_iteration} is computed using a Jacobian-based heuristic. Specifically, the product is computed as follows:  
(i) compute the Jacobian-vector product \(\vec{x} = \mathbf{J}_{h\theta_s} r^{(k)}\);  
(ii) apply the Hessian: \(\vec{x} \leftarrow \mathbf{H}_{h_s} \vec{x}\);  
(iii) compute the transposed Jacobian-vector product: \(\mathbf{G}r^{(k)} = \mathbf{J}_{h\theta_s}^\top \vec{x}\).
\fi

%% file: tex/appendix/4.additional_results.tex
\section{Experimental result on other datasets and GNNs}
\label{apx:other_gnns}

In this section, we present scatter plots for additional non-convex GNNs and datasets. \Cref{fig:other_datasets} shows the results for additional datasets, while \Cref{fig:chebnet} and \Cref{fig:gat} show the results for ChebNet~\citep{defferrard2016convolutional} and GAT~\citep{velivckovic2017graph}, respectively.  
Under these settings, our influence function accurately predicts the actual influence, consistently showing a correlation above \(0.8\).

\input{fig/figure/other_datasets}
\input{fig/figure/chebnet}
\input{fig/figure/gat}

%% file: fig/figure/other_datasets.tex
\begin{figure}[h]
    \centering
    \begin{subfigure}[b]{\textwidth}
        \centering
        \includegraphics[width=0.24\linewidth]{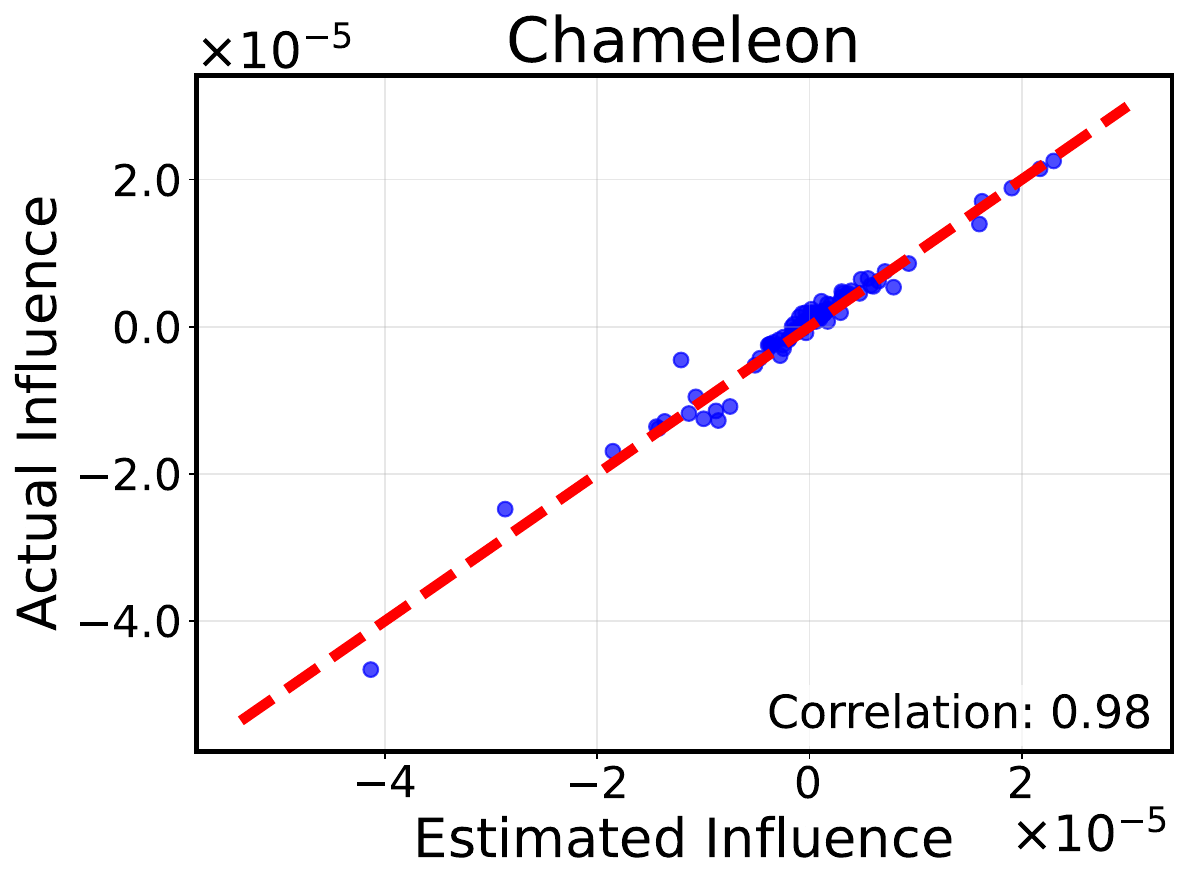}
        \includegraphics[width=0.24\linewidth]{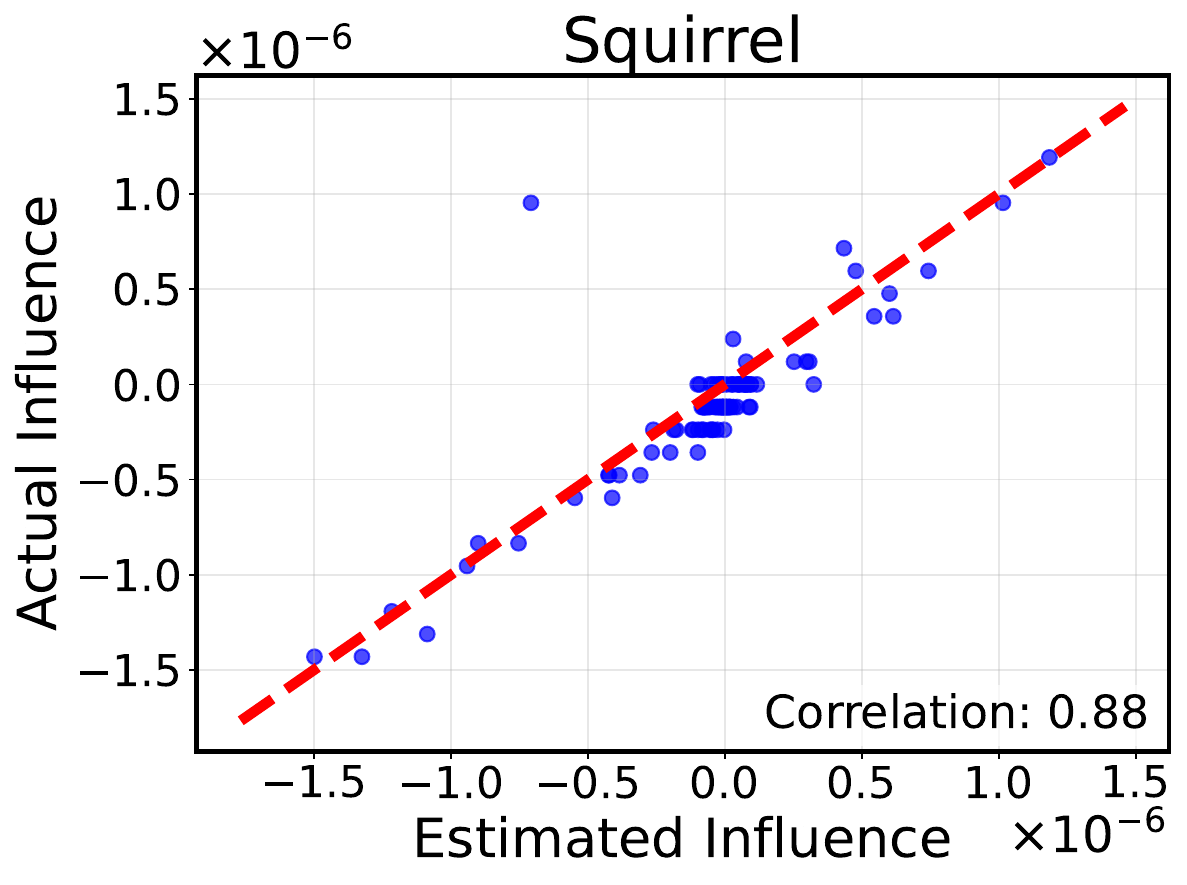}
        \includegraphics[width=0.24\linewidth]{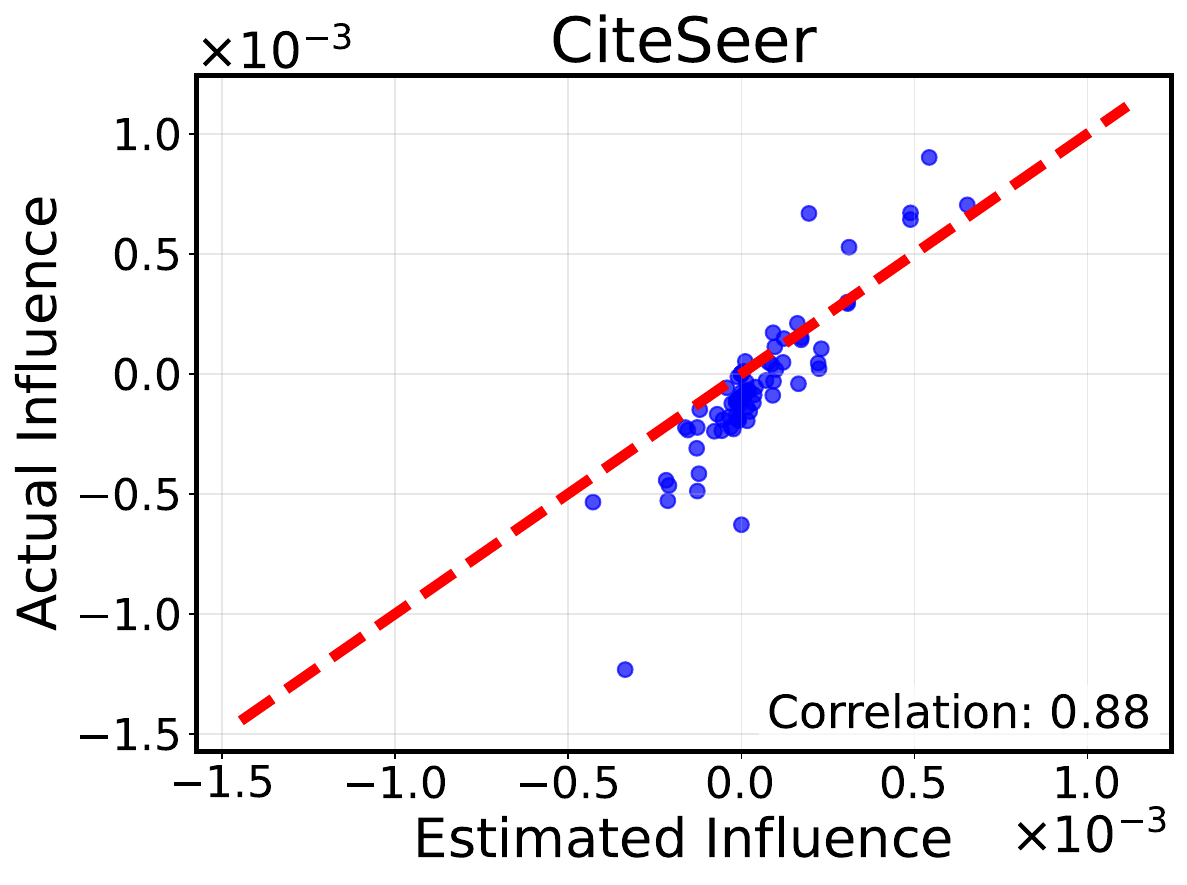}
        \includegraphics[width=0.24\linewidth]{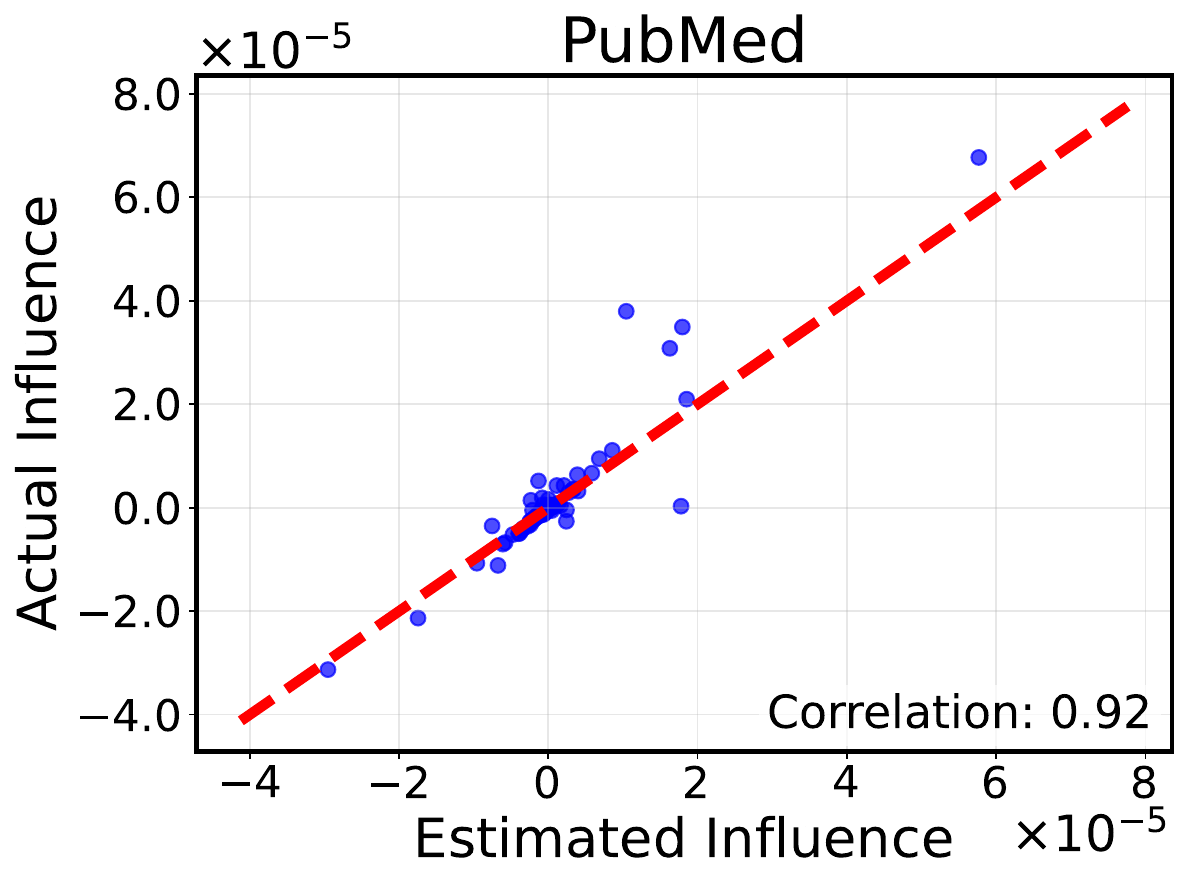}
        \caption{Deletion}
    \end{subfigure}

    \begin{subfigure}[b]{\textwidth}
        \centering
        \includegraphics[width=0.24\linewidth]{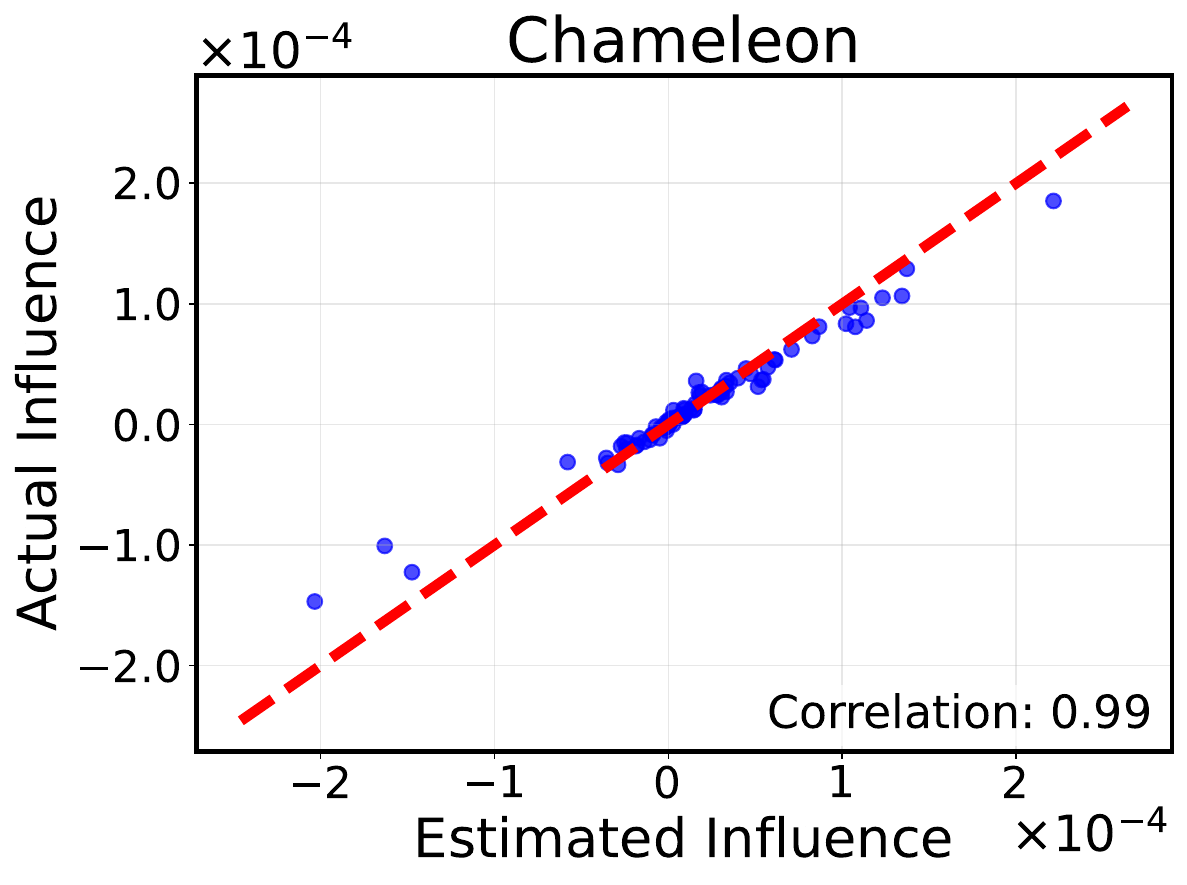}
        \includegraphics[width=0.24\linewidth]{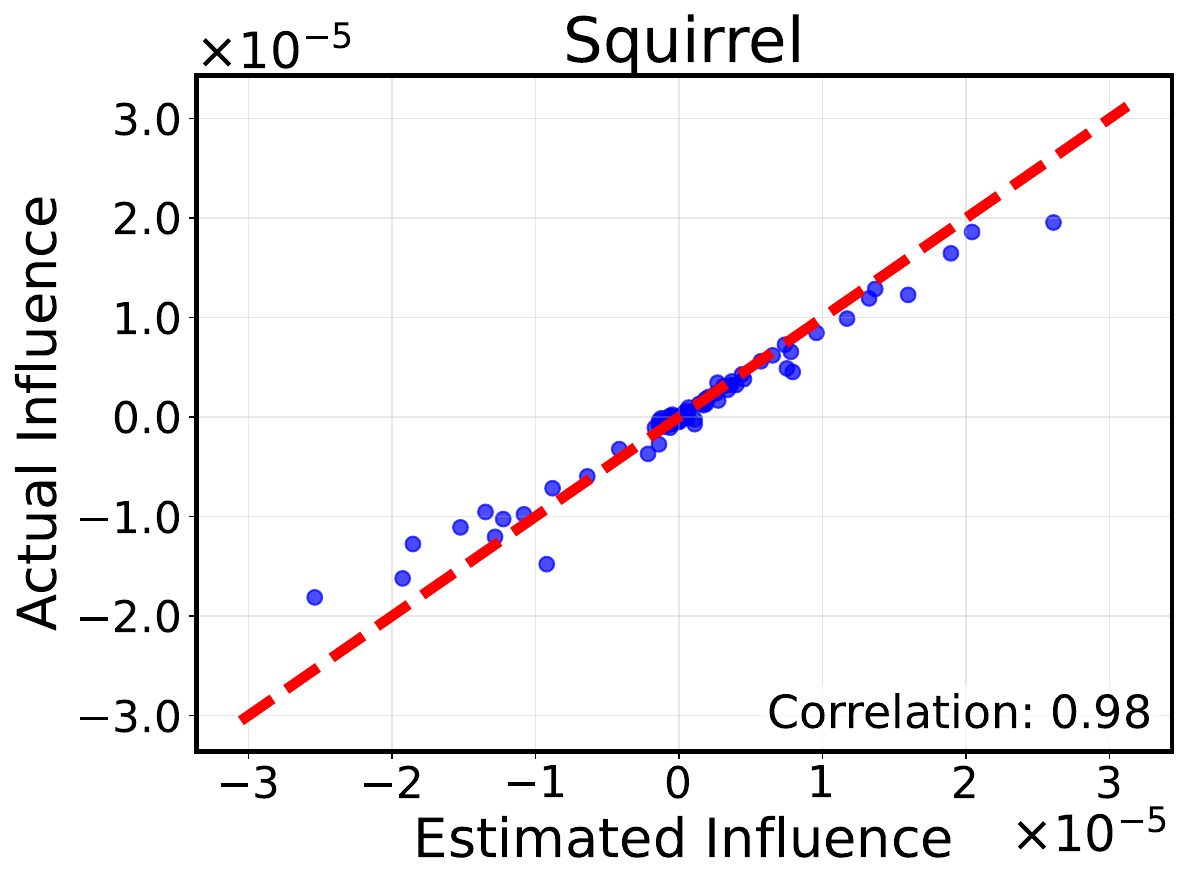}
        \includegraphics[width=0.24\linewidth]{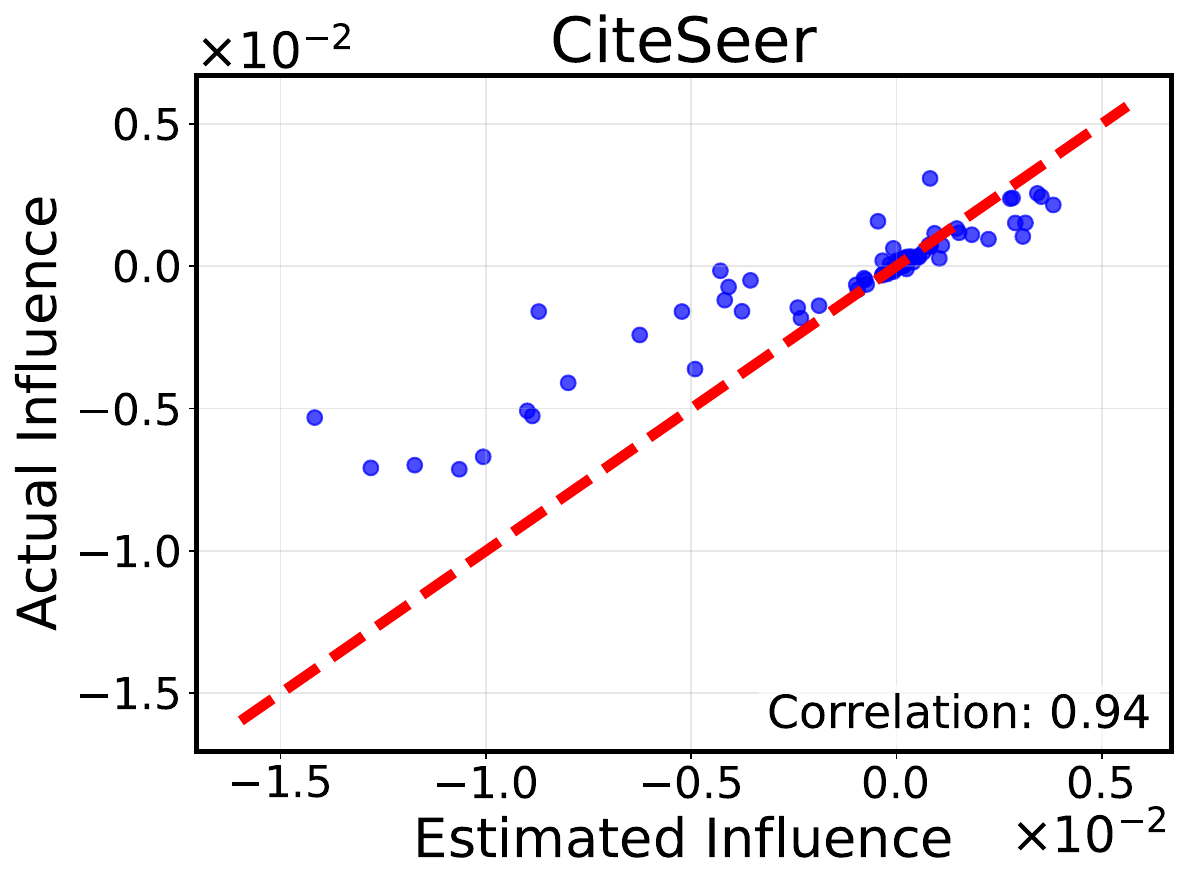}
        \includegraphics[width=0.24\linewidth]{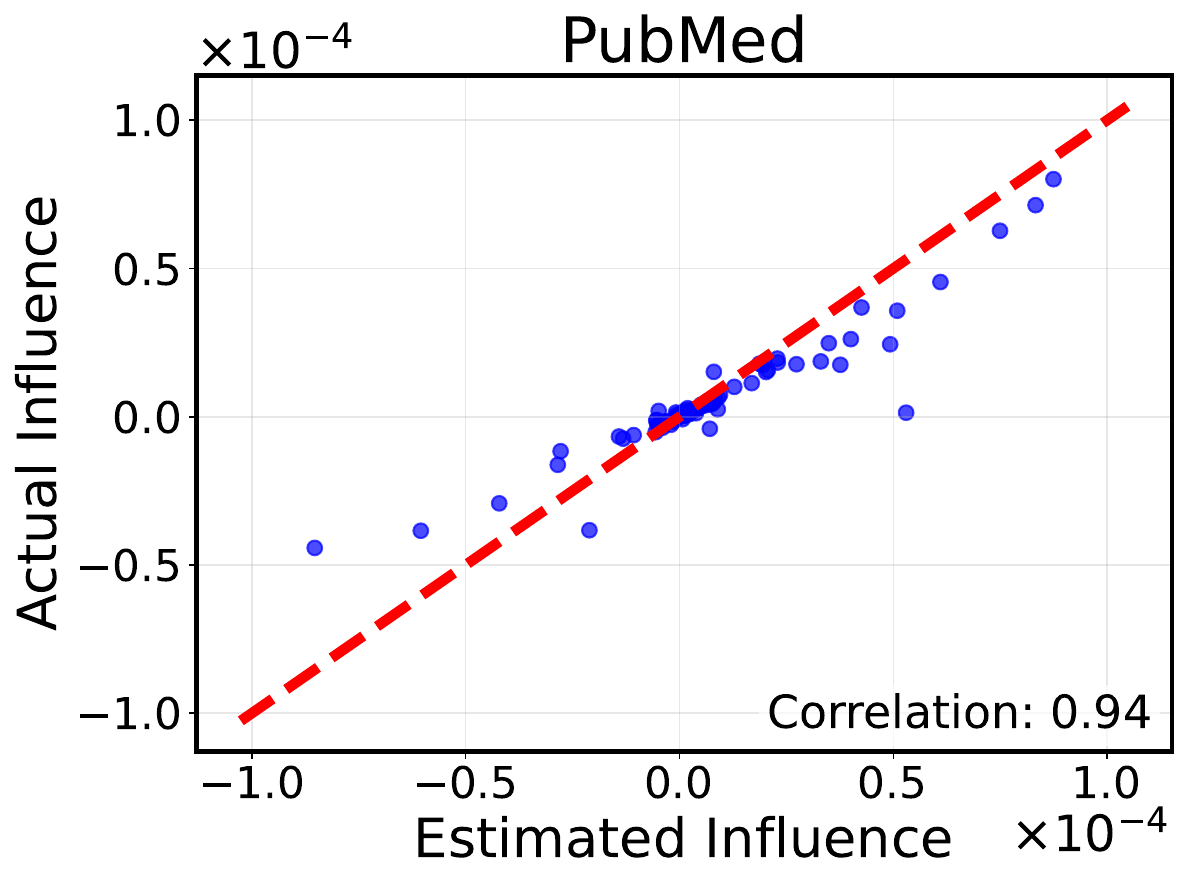}
        \caption{Insertion}
    \end{subfigure}
    
    \caption{The scatter plot of predicted influence and actual influence on four-layer GCN. The x-axis represents the predicted influence and y-axis represents the actual influence, and the red-dotted line represents the perfect alignment.}
    \label{fig:other_datasets}
\end{figure}

%% file: fig/figure/chebnet.tex
\begin{figure}[ht]
    \centering
    \begin{subfigure}[b]{\textwidth}
        \centering
        \includegraphics[width=0.32\linewidth]{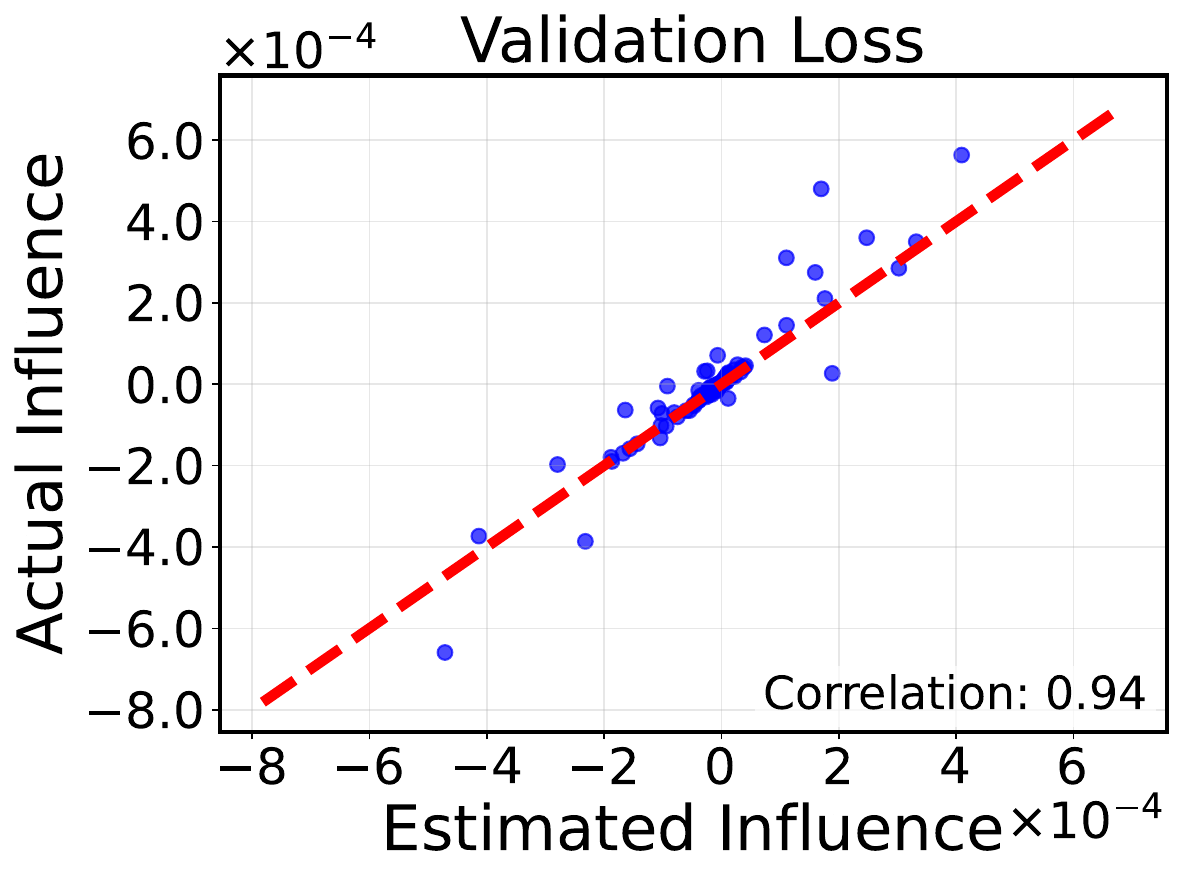}
        \includegraphics[width=0.32\linewidth]{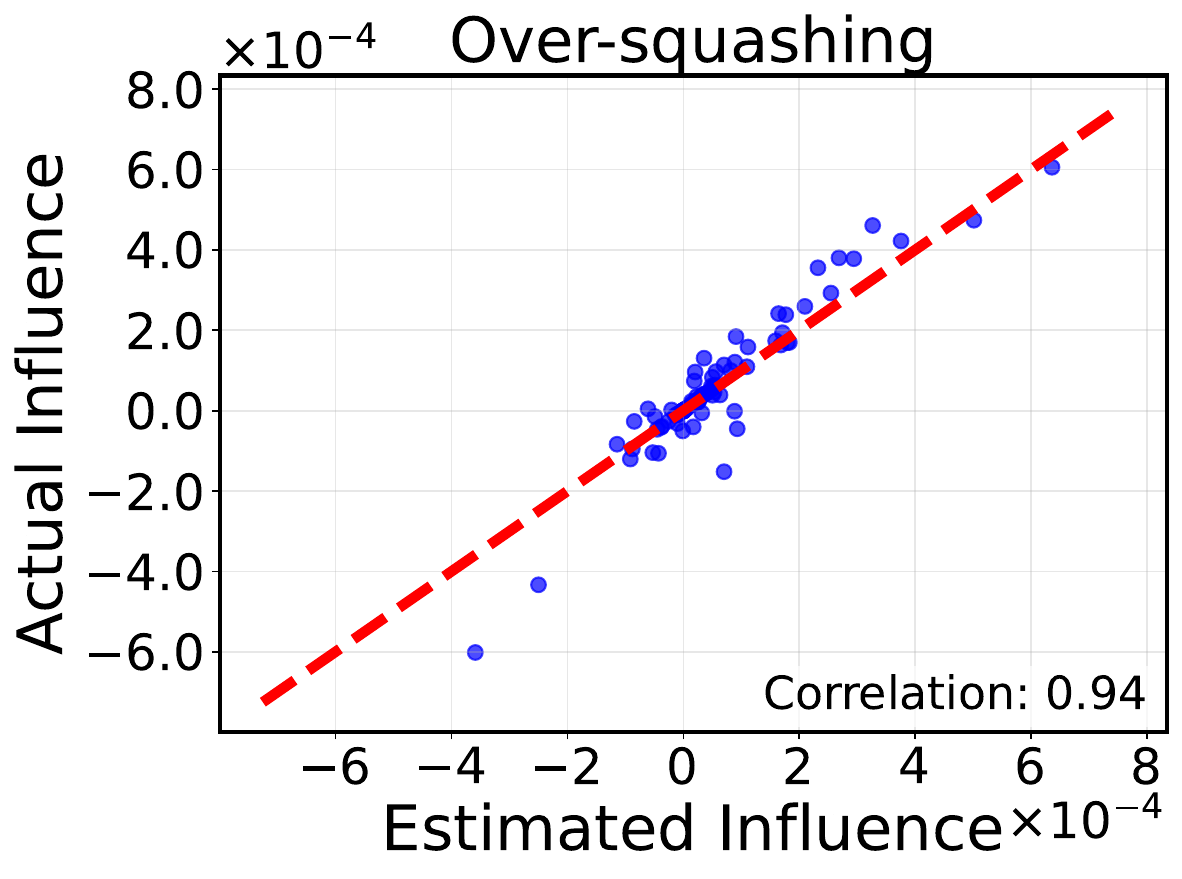}
        \includegraphics[width=0.32\linewidth]{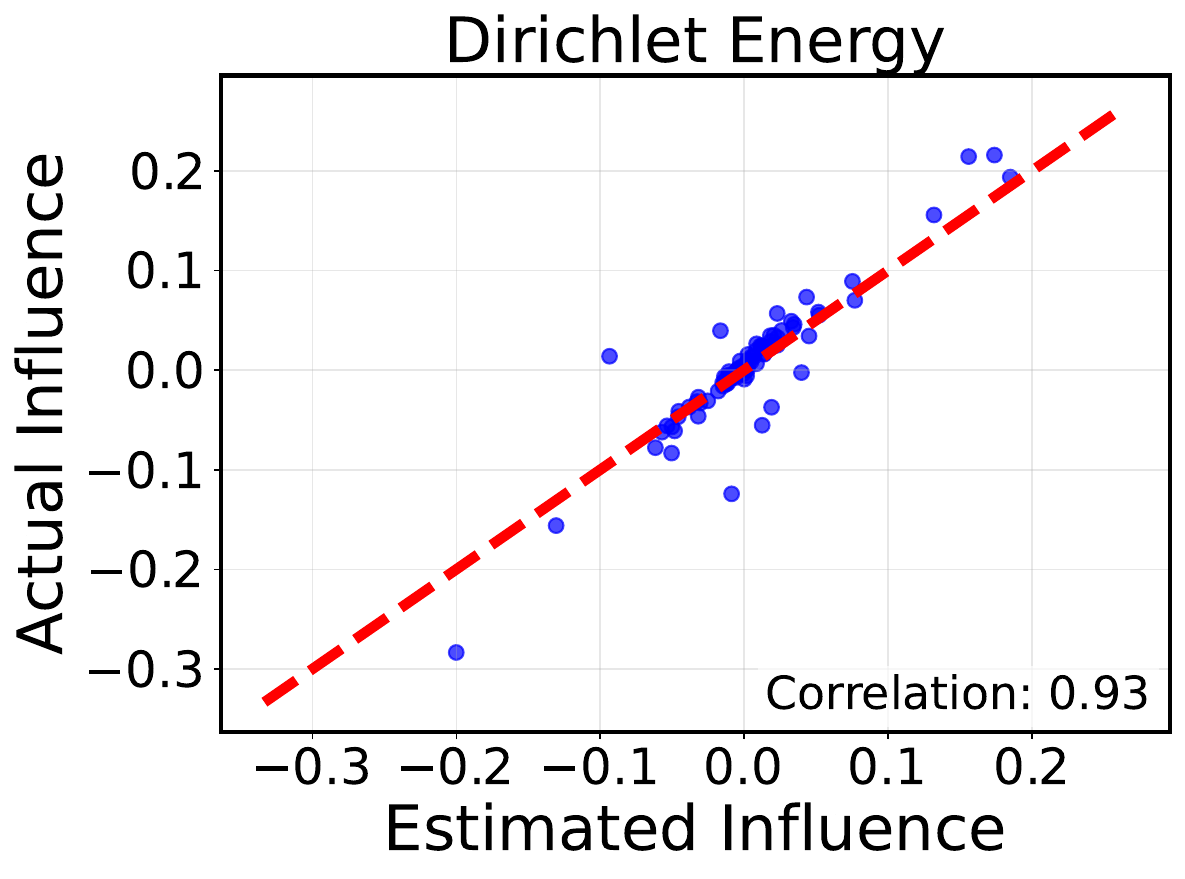}
        \caption{Deletion}
    \end{subfigure}

    \begin{subfigure}[b]{\textwidth}
        \centering
        \includegraphics[width=0.32\linewidth]{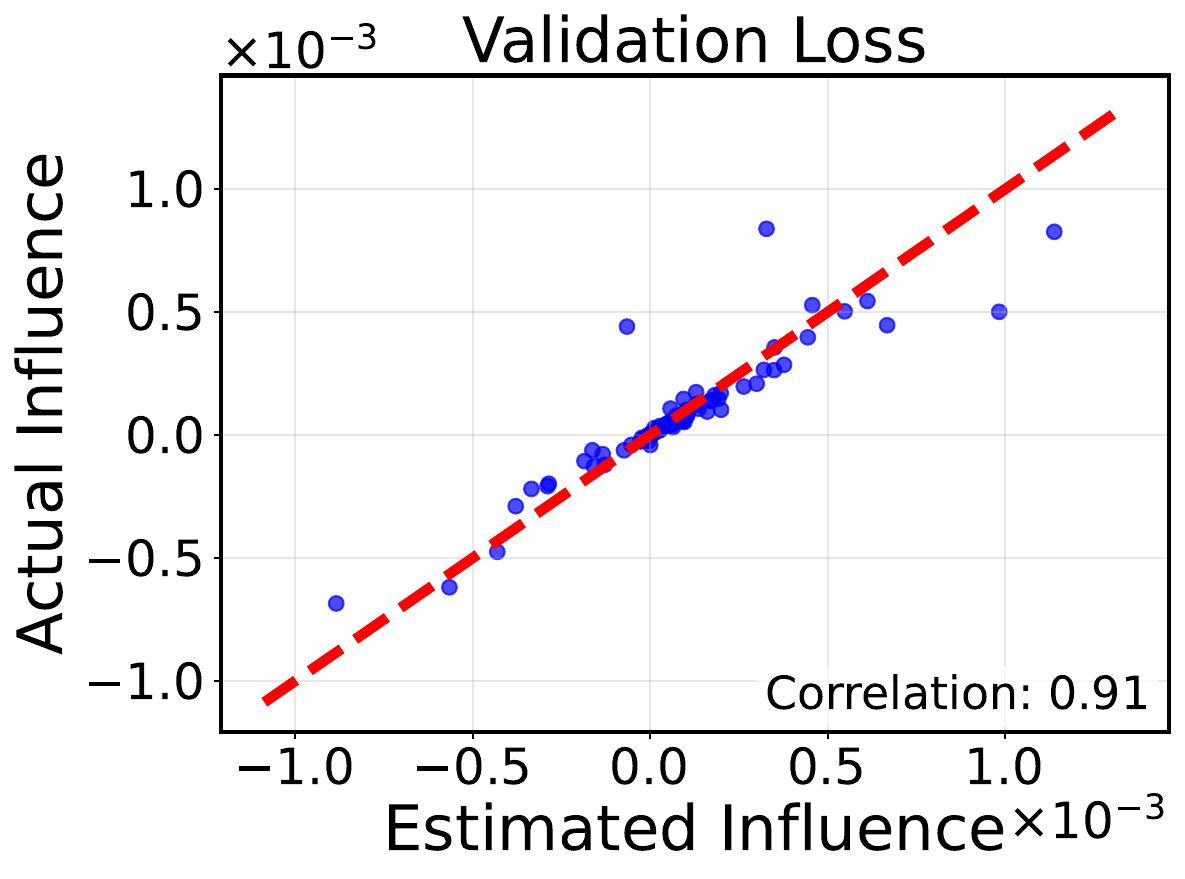}
        \includegraphics[width=0.32\linewidth]{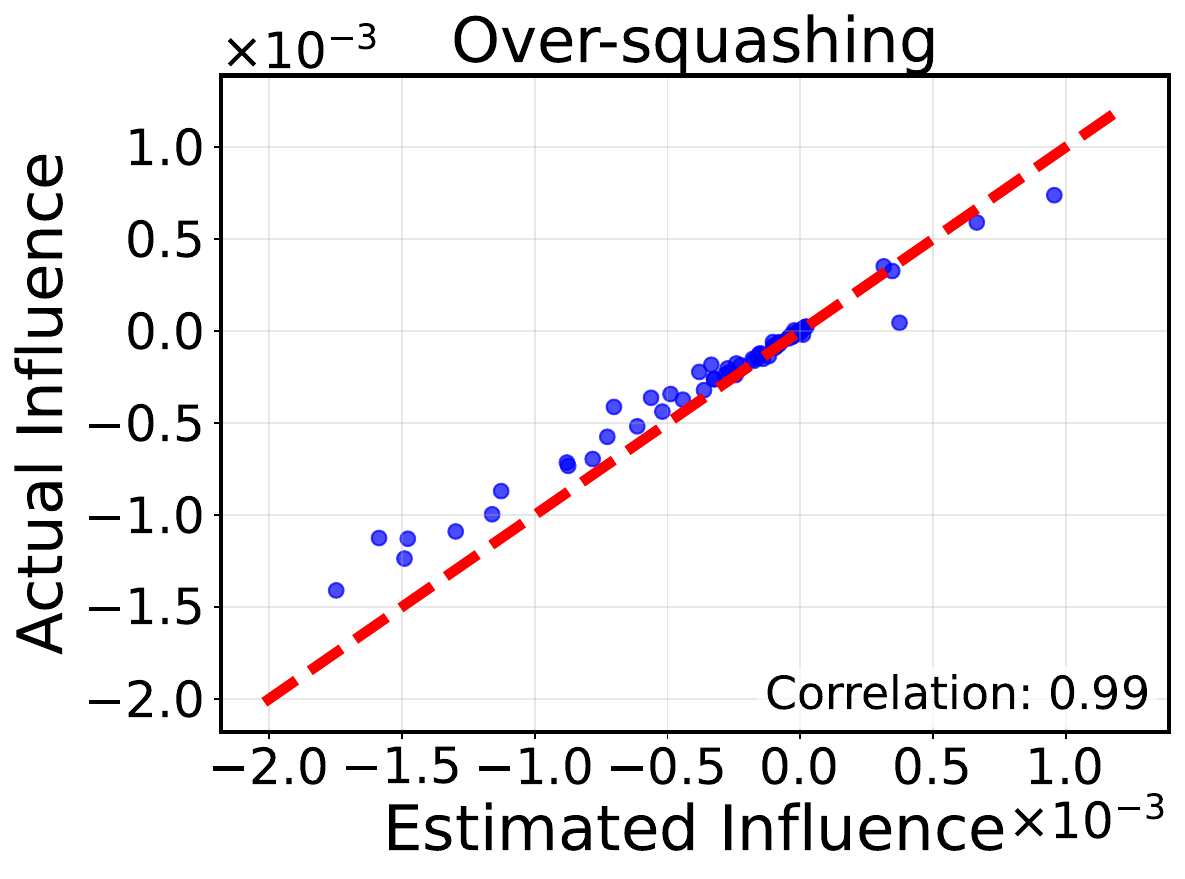}
        \includegraphics[width=0.32\linewidth]{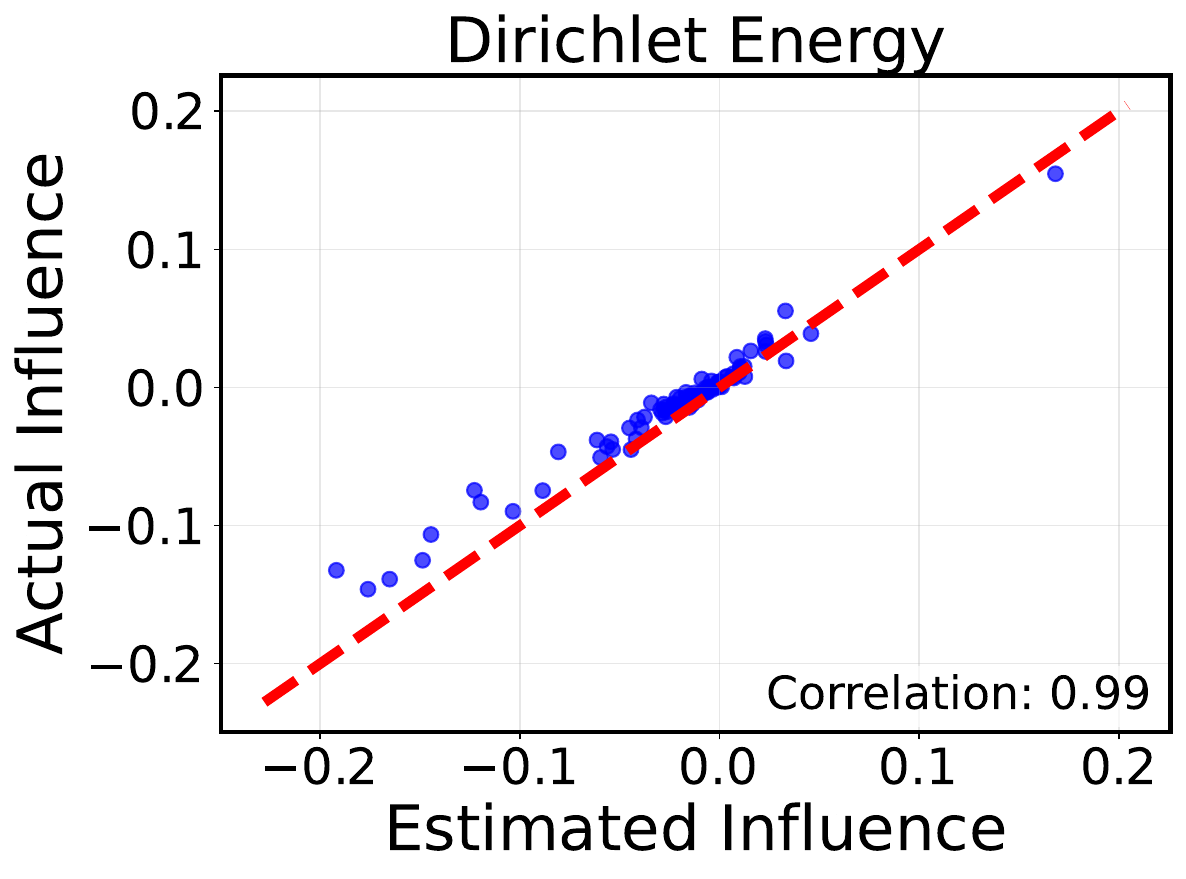}
        \caption{Insertion}
    \end{subfigure}
    
    \caption{The scatter plot of predicted influence and actual influence on two-layer ChebNet. The x-axis represents the predicted influence and y-axis represents the actual influence, and the red-dotted line represents the perfect alignment.}
    \label{fig:chebnet}
\end{figure}

%% file: fig/figure/gat.tex
\begin{figure}[ht]
    \centering
    \begin{subfigure}[b]{\textwidth}
        \centering
        \includegraphics[width=0.32\linewidth]{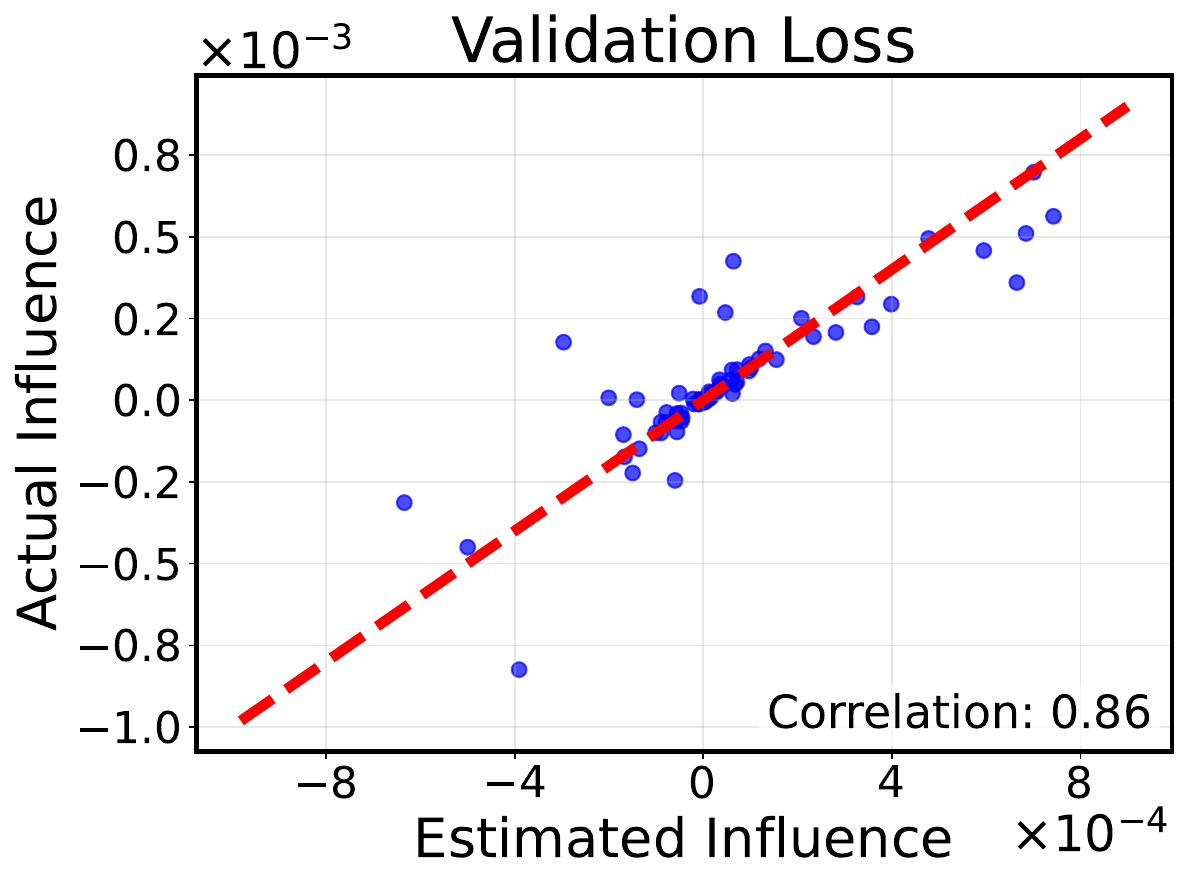}
        \includegraphics[width=0.32\linewidth]{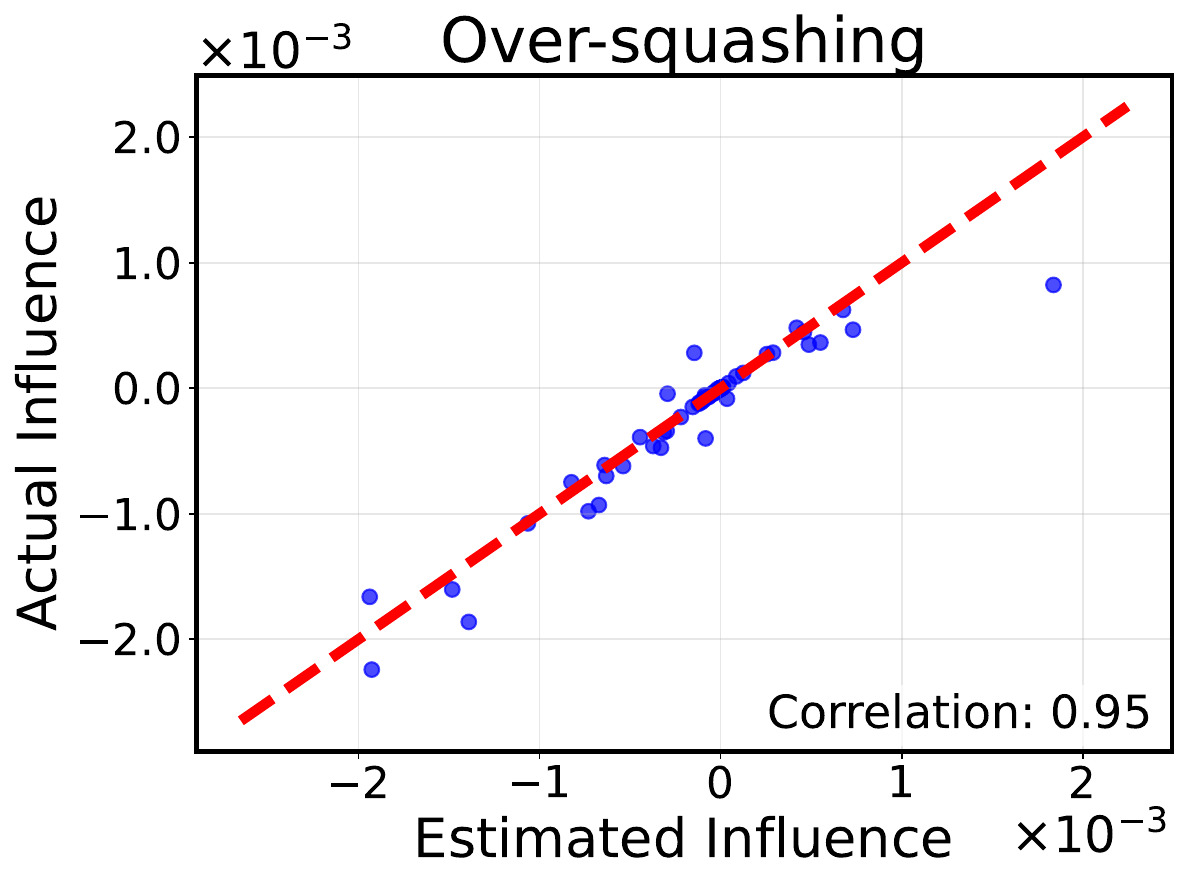}
        \includegraphics[width=0.32\linewidth]{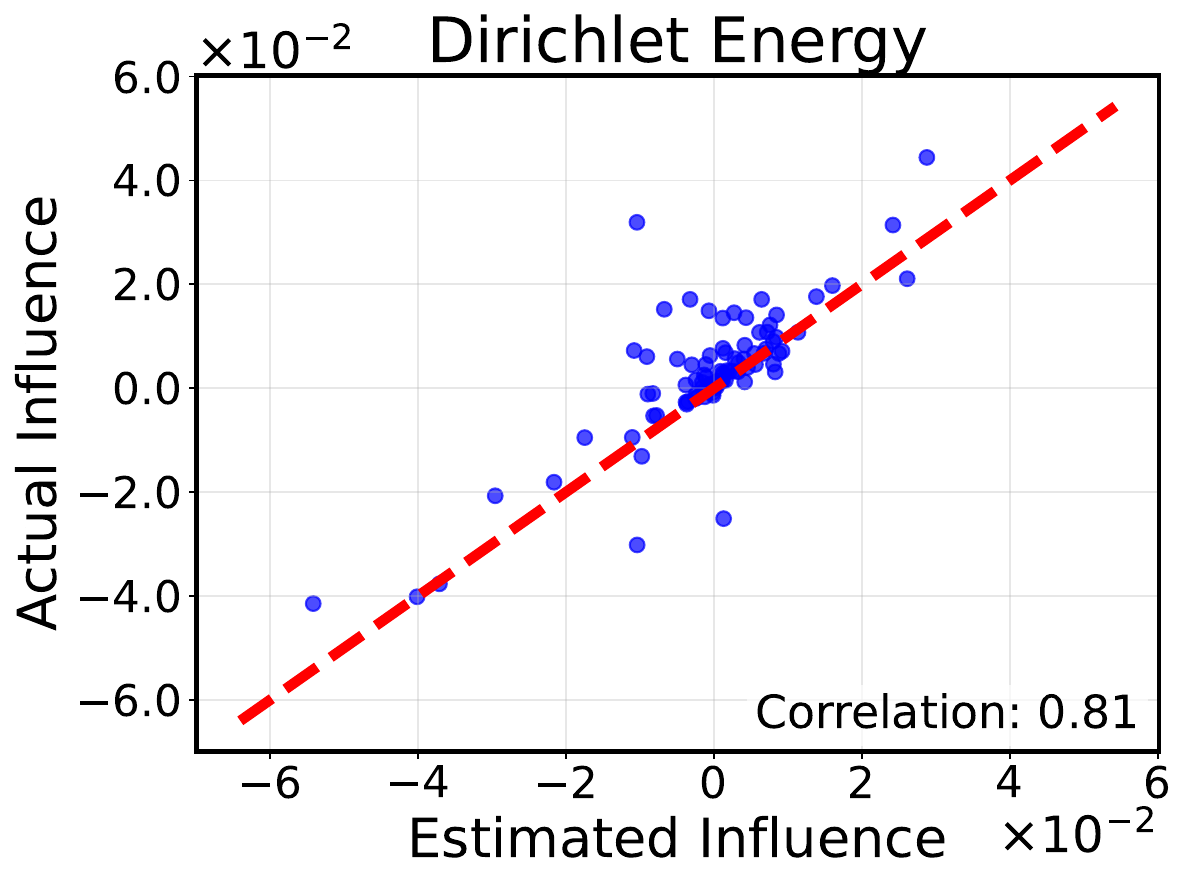}
        \caption{Deletion}
    \end{subfigure}

    \begin{subfigure}[b]{\textwidth}
        \centering
        \includegraphics[width=0.32\linewidth]{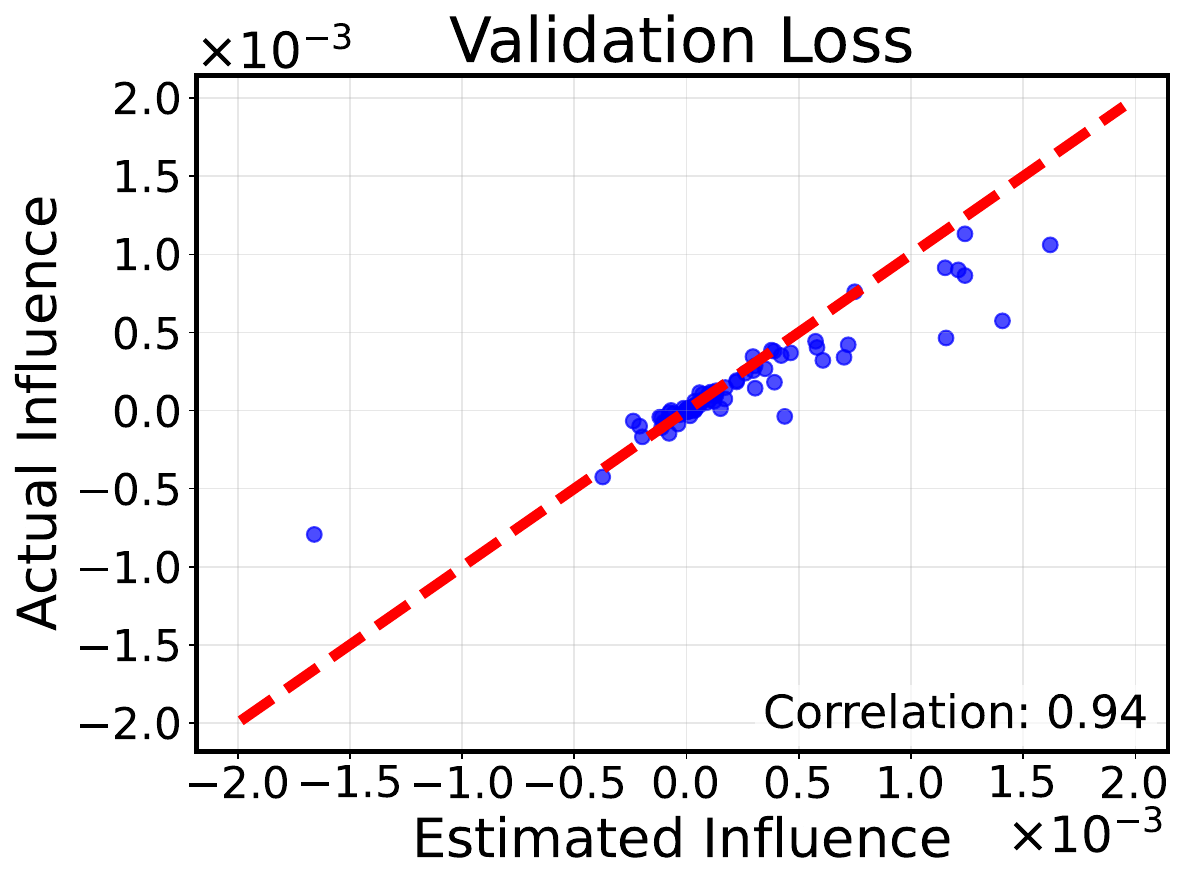}
        \includegraphics[width=0.32\linewidth]{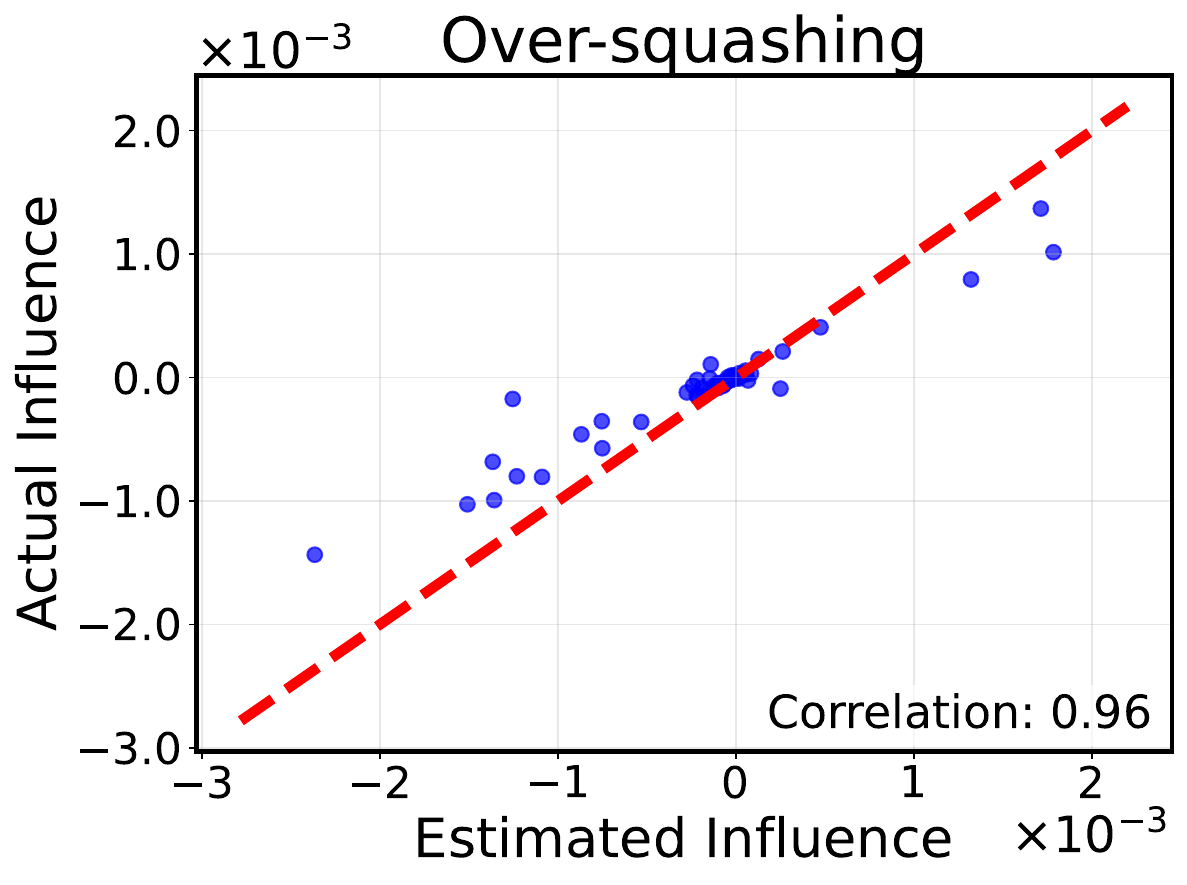}
        \includegraphics[width=0.32\linewidth]{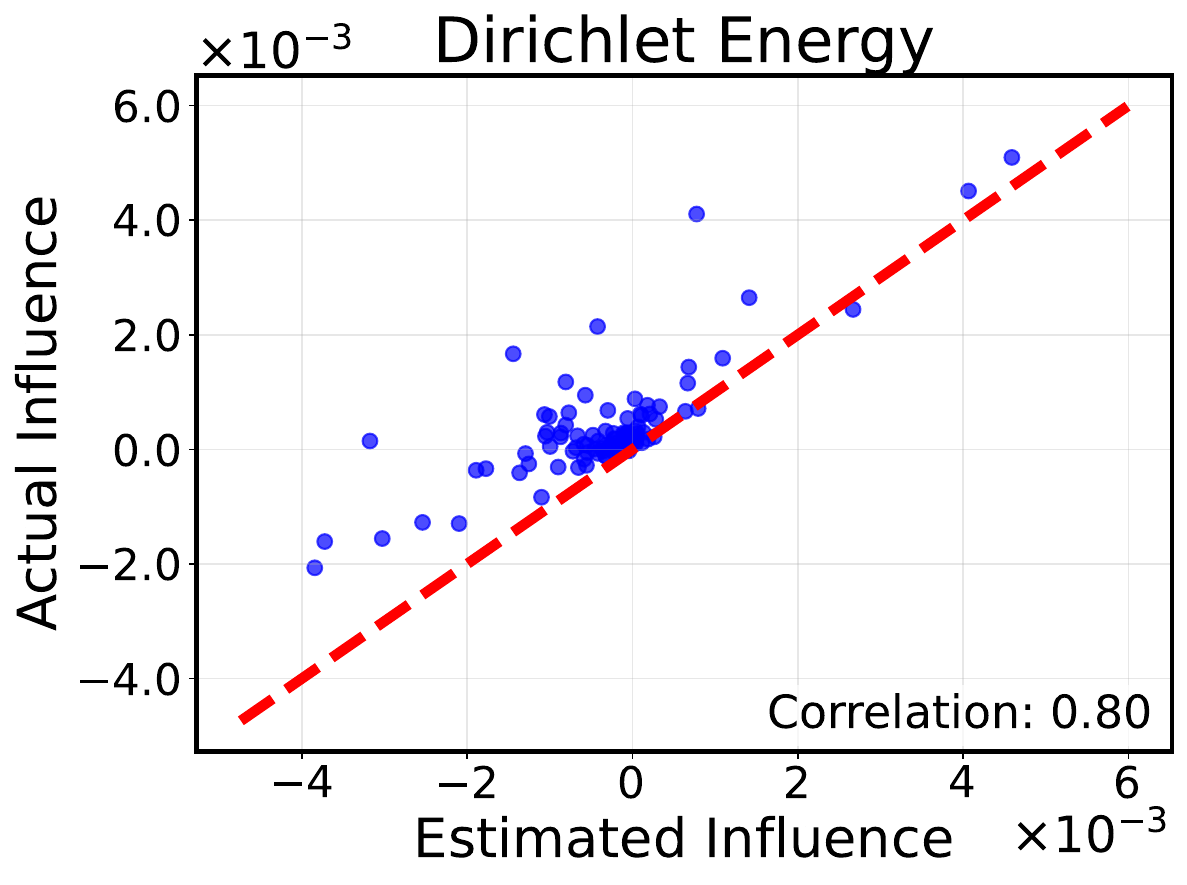}
        \caption{Insertion}
    \end{subfigure}
    
    \caption{The scatter plot of predicted influence and actual influence on two-layer GAT. The x-axis represents the predicted influence and y-axis represents the actual influence, and the red-dotted line represents the perfect alignment.}
    \label{fig:gat}
\end{figure}

%% file: tex/appendix/6.experimental_settings.tex
\section{Experimental configuration}
\label{apx:exp_config}

All experiments are conducted using NVIDIA GeForce RTX 3090, NVIDIA RTX A5000, and NVIDIA RTX A6000 GPUs. The experiments presented in the main text employ a 4-layer GCN model as a representative non-convex GNN.  
To produce the results in \Cref{tab:test_acc} and \Cref{tab:attack}, we tune the model and training hyperparameters of a vanilla GCN over the following search space: learning rates \(\{0.1, 0.03, 0.01\}\), hidden dimensions \(\{32, 64\}\), and weight decays \(\{10^{-3}, 10^{-4}, 10^{-5}, 10^{-6}, 10^{-7}\}\). Training is performed for 2000 epochs using the SGD optimizer. For influence function computation, we run the LiSSA algorithm for 10{,}000 iterations. The damping parameter \(\lambda\) is selected from \(\{0.1, 0.01, 0.001, 0.0001\}\). We randomly sample 10{,}000 candidate edges for both deletion and insertion, and estimate their influence. For \Cref{tab:test_acc}, the number of edges to edit is determined based on validation accuracy. The experiments are repeated for 10 independent runs using different random seeds provided by the BernNet implementation.

\textbf{BORF}~\citep{nguyen2023revisiting} and \textbf{FoSR}~\citep{karhadkar2023fosr}, used in the analysis of edge rewiring methods, are applied with the default settings from the original paper. The number of rewired edges reported in \cref{fig:histogram_rewiring} is aggregated over 10 runs, with a total of 200 edges inserted. Unlike FoSR, which performs only edge insertions, BORF also considers edge deletions.  
To improve visualization, we reduce the number of edge deletions, as the influence scores for insertions exhibit a long-tailed distribution, making it difficult to display them on the same scale as deletions.  
Accordingly, edge deletions are performed with 120, 100, and 100 edges, respectively.

For the adversarial attack experiments in \cref{tab:attack}, \textbf{DICE}~\citep{waniek2018hiding} and \textbf{PRBCD}~\citep{NEURIPS2021_3ea2db50} are implemented by modifying the PyTorch-based DeepRobust library~\citep{li2020deeprobust}, while maintaining its default settings. For our method, we consider both edge insertions and deletions, and select the operations with the highest influence scores for validation loss, and the lowest influence scores for over-squashing and Dirichlet energy.